\documentclass{article}

\usepackage{arxiv}
\makeatletter
\renewcommand{\@bottomtitlebar}{%
  \vskip 0.15in
  \hrule height 2pt
  \vskip -0.1in
}
\makeatother
\usepackage[utf8]{inputenc} 
\usepackage[T1]{fontenc}    
\usepackage{hyperref}       
\usepackage{url}            
\usepackage{booktabs}       
\usepackage{amsfonts}       
\usepackage{amssymb}
\usepackage{amsmath}
\usepackage{amsthm}
\usepackage{nicefrac}       
\usepackage{microtype}      
\usepackage{graphicx}
\usepackage{xcolor}
\usepackage{cite}
\usepackage{placeins}       
\usepackage[numbers]{natbib}
\usepackage{hyperref}
\usepackage{doi}  

\newtheorem{proposition}{Proposition}
\theoremstyle{remark}
\newtheorem{remark}{Remark}

\newcommand{\R}{\mathbb{R}}
\newcommand{\ncalls}{n_{\mathrm{calls}}}

\newcommand{\graphicalabstract}{
    figs/graphical_abstract.pdf
}

\title{Twincher: Towards Robust and Precise Physical AI \\
via Bijective Representation Learning}

\author{
  Arkady Gonoskov \\
  \small Department of Physics, University of Gothenburg,
  SE-41296 Gothenburg, Sweden\\
  \footnotesize arkady.gonoskov@physics.gu.se
}

\date{} 

\begin{document}
\begingroup
\renewcommand{\today}{}
\date{}
\maketitle
\endgroup

\vspace{-4.5em}


\begin{center}
    \includegraphics[
        width=\textwidth,
        keepaspectratio
    ]{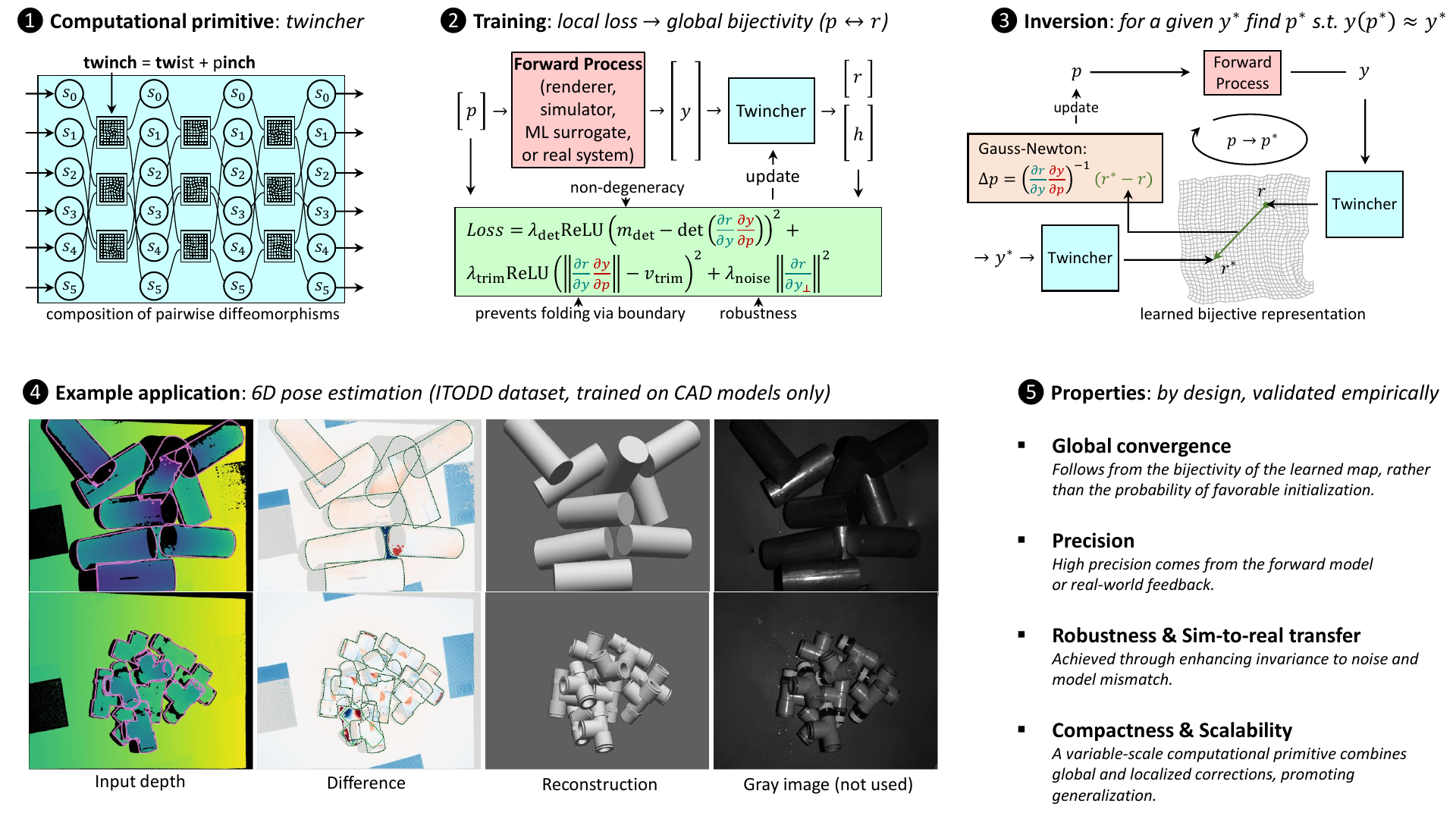}
\end{center}

\vspace{-0.5em}

\begin{abstract}
Perception and planning in physical systems often require inverting a continuous forward map $p \mapsto y = f(p)$, e.g., a renderer or a simulator, that can be queried but not inverted in closed form. Learned inverses are limited by their approximation error, whereas iterative solvers in observation space can stall at spurious stationary points. We instead learn a representation $r(y)$, insensitive to perturbations of $y$ normal to the data manifold, in which $p \mapsto r(f(p))$ is a well-conditioned bijection. In these coordinates, the least-squares inverse problem has no spurious interior stationary points, so that Gauss--Newton iteration with the exact forward model converges to the accuracy the data permit; the learned model need only be qualitatively correct. We introduce Twincher, an exactly invertible architecture of localized area-preserving rotations, trained with Jacobian-level objectives and a domain-growing curriculum. On synthetic maps with tunable nonconvexity, the worst-case residual reaches machine precision within five iterations once the query budget suffices, whereas an MLP-based baseline improves only gradually. In 6D localization of industrial objects in real depth images of the MVTec ITODD benchmark, given instance masks, twinchers trained only on CAD renderings make the iteration converge from a fixed initial guess within a few steps, without feature matching, pose regression, or pose hypotheses; in observation space, the same iteration fails for up to half of the views. All validation poses pass the strictest BOP thresholds and fit the sensor data about as closely as the ground-truth poses; 80\% of all 239 instances (67\% with visible masks only) are solved in agreement with the sensor data, and the number of training iterations grew only 1.5-fold from two to four pose parameters. Code is open source.
\end{abstract}


\section{Introduction}

Many tasks in perception and control can be cast as inverse problems for a continuous forward process $y = f(p)$ that maps controllable or latent parameters $p$ to observable quantities $y$~\cite{tarantola.siam.2005, arridge.actanumerica.2019}. Examples include estimating the pose or shape of an object from an image or a depth map by means of a renderer, i.e., analysis by synthesis~\cite{yuille.tics.2006, kulkarni.neurips.2015, loper.eccv.2014}; identifying the parameters of a physical system from measurements by means of a simulator~\cite{cranmer.pnas.2020}; and finding joint configurations that realize a desired end-effector state~\cite{dsouza.iros.2001, ardizzone.iclr.2019}. In such settings the forward process---a renderer, a physics-based simulator, a learned surrogate, or the physical system itself together with its digital twin~\cite{rasheed.access.2020}---can be evaluated, and often differentiated, but it cannot be inverted in closed form.

Two families of approaches are commonly used. \emph{Amortized} (direct) inversion trains a regressor $y \mapsto p$ on simulated pairs~\cite{arridge.actanumerica.2019, ongie.jsait.2020, cranmer.pnas.2020, radev.tnnls.2022}. It is fast at test time, but its accuracy is limited by the approximation error of the regressor, which typically decreases only as a power law in the amount of data~\cite{hestness.arxiv.2017, sharma.jmlr.2022}. \emph{Iterative} (optimization-based) inversion minimizes a data misfit such as $\|f(p) - y^\star\|^2$ by Gauss--Newton or related methods~\cite{nocedal.springer.2006}. It uses the exact forward model and can therefore reach high accuracy, but the misfit is generally nonconvex, and the iteration may stall at spurious stationary points---a difficulty known, for instance, as cycle skipping in seismic full-waveform inversion~\cite{virieux.geophysics.2009} and manifested as the limited convergence basin of render-and-compare pose estimation~\cite{li.deepim.eccv.2018}. Hybrid schemes that initialize an iterative solver with an amortized estimate~\cite{marino.icml.2018, adler.ip.2017} alleviate but do not remove this difficulty, because the residual error of the initial estimate may still place it outside the basin of attraction of the solution.

In this work we pursue a different division of labor. Rather than approximating the inverse map itself, we learn a change of coordinates in observation space, $y \mapsto r(y) \in \R^{n_p}$, such that the composite map $g = r\circ f$ is a well-conditioned bijection of the parameter domain onto its image. In these coordinates, the inverse problem $g(p) = r(y^\star)$ is a square nonlinear system with a nonsingular Jacobian. Its least-squares formulation has no spurious interior stationary points (Proposition~\ref{prop:stationary}), and Newton-type iterations, which query the exact forward model, converge quadratically near the solution. The learned component thus plays the role of a learned nonlinear (left) preconditioner of the inverse problem~\cite{cai.sisc.2002} or, equivalently, of a learned misfit function~\cite{engquist.cms.2014, sun.frontiers.2022}. Crucially, it needs to be \emph{qualitatively} correct---in terms of topology and conditioning---rather than accurate, because accuracy is supplied by the forward model during refinement. We additionally require $r$ to be insensitive to perturbations of $y$ normal to the data manifold $\mathcal{M} = f(\mathcal{P})$; to first order, this makes the inversion as stable as least-squares projection onto $\mathcal{M}$ (Proposition~\ref{prop:sensitivity}). We refer to the problem of learning such representations as \emph{bijective representation learning}. It is related to approximate bijective correspondence~\cite{murphy.arxiv.2021} and to identifiable representation learning~\cite{hyvarinen.nn.1999, khemakhem.aistats.2020, locatello.arxiv.2018}, with the difference that here the generative factors are known and can be queried, so that the emphasis is on the conditioning of inference rather than on identifiability.

To learn such representations, we introduce \emph{Twincher}, a parameterized diffeomorphism of observation space whose leading $n_p$ output coordinates define $r$. A twincher is a stack of elementary pairwise transformations---\emph{twinches}---each of which rotates a pair of coordinates by an angle that decays with the distance from a learnable center. A twinch is an area-preserving twist map with a closed-form inverse. Consequently, every twincher is volume-preserving and exactly invertible, in common with additive coupling flows~\cite{dinh.arxiv.2014}, while the locality and rotational character of its building block provide an inductive bias toward redistributing, rather than amplifying or suppressing, sensitivity among coordinates. The training objectives act on Jacobians rather than on output values and are optimized with a domain-growing curriculum and dedicated routines that propagate derivative tensors with memory independent of depth.

Our contributions are as follows.
\begin{itemize}
    \item \textbf{Formulation.} We formulate bijective representation learning for the inversion of queryable continuous forward models and state two elementary properties that motivate it: in representation coordinates the least-squares formulation of the inverse problem has no spurious interior stationary points, and insensitivity to normal perturbations makes the inversion, to first order, as stable as least-squares projection onto the data manifold (Sec.~\ref{sec:formulation}).
    \item \textbf{Architecture.} We introduce the twinch, a localized and area-preserving rotation with a closed-form inverse, and the Twincher architecture built from it, together with forward- and reverse-mode routines for derivative tensors whose cost is $O(n_p)$ forward passes and whose memory is independent of depth (Sec.~\ref{sec:architecture} and Appendix~\ref{app:derivatives}).
    \item \textbf{Training.} We propose a training procedure based on Jacobian-level objectives that enforce nondegeneracy, bounded distortion, and insensitivity to normal perturbations, combined with a domain-growing curriculum that counteracts folding of the learned representation (Sec.~\ref{sec:training}).
    \item \textbf{Empirical evaluation.} We introduce \emph{harmonic entanglers}, a family of synthetic forward maps with tunable nonconvexity, and show that once the query budget suffices to form a bijective representation, the worst-case residual reaches machine precision within a few iterations, in contrast to the gradual improvement of a parameter-matched baseline. A depth-map example demonstrates an inference error that scales linearly with the observation noise (Sec.~\ref{sec:experiments}).
    \item \textbf{Application to real sensor data.} We apply twinchers to 6D localization of industrial objects in real depth images of the MVTec ITODD benchmark~\cite{drost.iccvw.2017}. Trained only on renderings of CAD models, with atlases of up to six charts of $n_p = 4$ parameters, they make Gauss--Newton iteration with the exact renderer converge from a fixed initial guess for all tested views, whereas the same iteration in observation space fails for up to half of the views from the same initial guess and for about two thirds of them from random initial guesses. On all validation instances, the poses are correct at the strictest thresholds of the BOP benchmark. The number of training iterations grew only 1.5-fold from two to four parameters, far less than the number of grid nodes needed to resolve the forward map. We also report what made training reliable at this scale: a diagnostic of global injectivity by winding numbers, constraints beyond the current curriculum domain, function-preserving growth of depth, and a momentum reset against loss spikes (Sec.~\ref{sec:itodd} and Appendix~\ref{app:itodd}).
    \item \textbf{Software.} We release an open-source Python package for CPU and GPU that implements the method and reproduces the experiments~\cite{twincher-code}.
\end{itemize}

Our synthetic experiments are deliberately small-scale: they are designed to isolate the mechanism. The application in Sec.~\ref{sec:itodd} then tests it on a well-studied real-world task, in which a mature field offers a clear reference for what is achievable; its purpose is to show what the principle enables at an early stage of its development, not to propose a competing pose estimator. Limitations, including scaling to higher-dimensional parameter spaces and to non-injective forward maps, are discussed in Secs.~\ref{sec:itodd-discussion} and~\ref{sec:discussion}. More broadly, we view robust bijective representations as a building block for perception--planning--control pipelines in robotics and physical AI. Such pipelines increasingly combine learned models with simulators and world models~\cite{ha.arxiv.2018, moerland.arxiv.2020, plaat.arxiv.2021, sakagami.frai.2023, ding.arxiv.2024, guan.arxiv.2024, hou.arxiv.2026}, vision-language(-action) foundation models~\cite{han.if.2026, kawaharazuka.arxiv.2025, shao.arxiv.2025}, and direct policy learning~\cite{celemin.arxiv.2022, wang.arxiv.2024, wolf.arxiv.2025, li.arxiv.2026}. Physical AI places two demands on their inference modules that are hard to meet simultaneously: \emph{robustness}, i.e., reliable inference for every input in the operating range, including inputs corrupted by noise, occlusion, or model mismatch; and \emph{precision}, i.e., an accuracy limited by the measurements rather than by the learned model. In the approach studied here, the two are supplied by different components: the learned representation provides the global structure that makes the iteration converge, and the physics-based forward model provides the accuracy. The present work also illustrates how the choice of computational primitive can supply an inductive bias~\cite{Lecun1998-xc, vaswani.arxiv.2017, bietti.arxiv.2019, liu.arxiv.2024} that is complementary to scaling~\cite{kaplan.arxiv.2020}.

\section{Related work}
\label{sec:related}

\paragraph{Learning-based inversion.}
Data-driven methods for inverse problems are reviewed in~\cite{arridge.actanumerica.2019, ongie.jsait.2020}. Simulation-based inference targets posterior distributions over parameters using simulators~\cite{cranmer.pnas.2020}, often with amortized neural estimators built from invertible networks~\cite{radev.tnnls.2022}. Invertible neural networks have been used to analyze non-unique inverse problems by mapping parameters to observations augmented with latent variables~\cite{ardizzone.iclr.2019}, and conditional normalizing flows have been used as visuomotor policies~\cite{li.arxiv.2026}. Learned iterative schemes learn or unroll the update rule of an optimization method~\cite{adler.ip.2017, marino.icml.2018}; in variational inference, the gap between amortized and per-instance optimized solutions is known as the amortization gap~\cite{gershman.cogsci.2014, cremer.icml.2018}. In vision, analysis-by-synthesis and render-and-compare approaches iteratively compare rendered and observed images~\cite{yuille.tics.2006, kulkarni.neurips.2015, loper.eccv.2014, li.deepim.eccv.2018}. Perception as inference over a generative model is also central to active inference~\cite{Friston2018-do}. Our approach differs from these lines of work in that it neither regresses $p$ nor learns the update rule; it learns coordinates in which a standard Newton-type iteration on the exact forward model becomes reliable, and a learned inverse estimate, if any, is used only for initialization.

\paragraph{Object pose estimation.}
Model-based 6D pose estimation of rigid objects is benchmarked by BOP~\cite{hodan.eccv.2018, hodan.eccvw.2020, sundermeyer.cvprw.2023, hodan.cvprw.2024}. Classical methods match templates~\cite{hinterstoisser.accv.2012} or point-pair features of the depth map to the object model~\cite{drost.cvpr.2010, vidal.sensors.2018} and refine the pose by the iterative closest point algorithm~\cite{besl.tpami.1992}. Learned methods regress the pose or keypoints~\cite{xiang.rss.2018, peng.cvpr.2019}, predict dense 2D--3D correspondences for PnP with RANSAC~\cite{hodan.cvpr.2020, wang.cvpr.2021, haugaard.cvpr.2022, su.cvpr.2022}, or refine pose hypotheses by learned render-and-compare~\cite{li.deepim.eccv.2018, labbe.eccv.2020, labbe.corl.2022, lipson.cvpr.2022, wen.cvpr.2024}. Analysis by synthesis with differentiable renderers~\cite{kato.cvpr.2018, liu.iccv.2019, yen-chen.iros.2021} minimizes an image-space misfit and therefore inherits its nonconvexity. Closest in spirit are methods that learn, from synthetic renderings, an embedding of views in which distances reflect pose similarity, and retrieve the nearest view from a codebook~\cite{wohlhart.cvpr.2015, sundermeyer.eccv.2018}. A twincher, too, learns a representation from renderings alone, but this representation is an invertible coordinate system designed for Newton-type iteration with the exact renderer rather than for retrieval, so that its accuracy is not limited by the resolution of a codebook.

\paragraph{Reformulating the misfit and nonlinear preconditioning.}
The nonconvexity of data misfits is a long-standing difficulty in inverse problems~\cite{tarantola.siam.2005, engl.kluwer.1996}. In full-waveform inversion, cycle skipping~\cite{virieux.geophysics.2009} has motivated alternative misfits based on optimal transport~\cite{engquist.cms.2014} as well as learned misfit functions~\cite{sun.frontiers.2022}. In numerical analysis, nonlinear preconditioning transforms a nonlinear system into an equivalent one on which Newton's method converges more reliably~\cite{cai.sisc.2002}, and neural operators have recently been used for this purpose~\cite{lee.arxiv.2025}; continuation (homotopy) methods address the same difficulty by deforming an easy problem into the target one~\cite{allgower.springer.1990}. A twincher can be viewed as a learned left preconditioner, or learned misfit, for which the properties that matter for Newton-type iterations---bijectivity and conditioning---are targeted explicitly.

\paragraph{Invertible architectures and diffeomorphic transformations.}
Normalizing flows~\cite{papamakarios.jmlr.2021, kobyzev.tpami.2021} construct expressive invertible maps from coupling layers~\cite{dinh.arxiv.2014, dinh.arxiv.2016, kingma.arxiv.2018}, from localized planar and radial transformations~\cite{rezende.icml.2015}, or from residual blocks with Lipschitz constraints~\cite{behrmann.icml.2019}; the numerical stability of their inverses has been analyzed in~\cite{behrmann.aistats.2021}. Coupling-based flows are universal approximators of diffeomorphisms~\cite{teshima.neurips.2020}, and augmenting the dimension of the state enlarges the class of maps realizable by continuous flows~\cite{dupont.neurips.2019}. Exact invertibility can be exploited to recompute activations during backpropagation instead of storing them~\cite{gomez.neurips.2017, jacobsen.iclr.2018}, and fully invertible networks show that discarding information is not necessary for learning useful representations~\cite{jacobsen.iclr.2018}. In computer vision, parameterized diffeomorphisms are used for alignment and registration, e.g., continuous piecewise-affine-based (CPAB) transformations~\cite{freifeld.iccv.2015, freifeld.tpami.2017, shapiraweber.neurips.2019} and large-deformation diffeomorphic metric mapping~\cite{beg.ijcv.2005}. The twinch is closely related to the radial flow of~\cite{rezende.icml.2015}, which dilates space around a learnable center; a twinch instead rotates it, which keeps the transformation volume-preserving. Twinchers can also be viewed as networks of Givens rotations~\cite{golub.jhu.2013}, such as those used to parameterize orthogonal and unitary layers~\cite{jing.icml.2017}, in which the rotation angles are localized functions of the state rather than constants. Unlike normalizing flows, twinchers are not trained by maximum likelihood; their objectives constrain the geometry of a sub-block of the Jacobian.

\paragraph{Alternative computational primitives.}
Our work belongs to a line of research that explores alternatives to MLPs built from affine maps and pointwise nonlinearities by changing the functional form of the elementary building block, as in Kolmogorov--Arnold networks~\cite{liu.arxiv.2024}, spline-based networks~\cite{fakhoury.arxiv.2022}, and tensor-product networks~\cite{park.arxiv.2025}. These works show that such changes can improve interpretability, compositionality, or generalization. Whereas these approaches primarily target expressivity and structural decomposition, twinchers target invertibility and well-conditioned geometry of the learned map.

\paragraph{Geometry and invariance of learned representations.}
Penalizing derivatives of a representation is a classical way of inducing invariance: contractive autoencoders penalize the full encoder Jacobian~\cite{rifai.icml.2011}, tangent propagation penalizes derivatives along known transformation directions~\cite{simard.neurips.1992}, and Jacobian regularization uses random projections to control input sensitivity~\cite{hoffman.arxiv.2019}; losses on derivatives are also used in double backpropagation~\cite{drucker.tnn.1992} and Sobolev training~\cite{czarnecki.neurips.2017}. Learning invariant mappings has also been approached via contrastive objectives~\cite{hadsell.cvpr.2006}. A substantial body of work structures latent spaces geometrically~\cite{arvanitidis.arxiv.2017, hadjeres.arxiv.2017, miani.neurlips.2022, gallego-posada.2021}, including approaches that relate encoders to the nearest-point projection onto the decoder manifold and use its reach to certify uniqueness~\cite{hauschultz.arxiv.2022}, or that make the encoder the decoder's pseudo-inverse~\cite{gropp.arxiv.2020}; equivariance-regularized latent spaces improve latent generative models~\cite{kouzelis.2025, rombach.arxiv.2021}. Topological structure has been incorporated via persistent homology and simplicial constructions~\cite{hofer.2019, gabrielsson.2020, medbouhi.mdpi.2023, polianskii.thesis.2018}. Disentanglement methods~\cite{higgins.arxiv.2017, chen.arxiv.2018, kim.arxiv.2018, rotman.arxiv.2022, higgins.arxiv.2022} and information-bottleneck objectives~\cite{alemi.arxiv.2017} aim at representations aligned with generative factors, which is impossible without inductive biases or supervision~\cite{locatello.arxiv.2018, hyvarinen.nn.1999}. In our setting, the factors are queryable, and the normal-space penalty of Sec.~\ref{sec:training} can be viewed as a contractive penalty restricted to the directions that carry no information about $p$; Proposition~\ref{prop:sensitivity} relates it to the pseudo-inverse property considered in~\cite{gropp.arxiv.2020}.

\paragraph{Robust training.}
Adversarial training constructs worst-case input perturbations by gradient-based methods~\cite{goodfellow.arxiv.2014, madry.arxiv.2017} and can be interpreted within distributionally robust optimization~\cite{rahimian.arxiv.2019}. Our robustness objective is different in nature: it targets the sensitivity of the representation in directions that are irrelevant to $p$, rather than the output of a predictor; adversarial variants are discussed in Sec.~\ref{sec:variants}.

\section{Problem formulation}
\label{sec:formulation}

\paragraph{Forward process and inverse problem.}
Let $p \in \mathcal{P} = [-1,1]^{n_p}$ denote the parameters of interest (e.g., pose, shape, or control variables) and $\nu \in \mathcal{N}$ nuisance variables that capture noise and model mismatch. Observations are generated by a continuous forward process
\begin{equation}
y = f(p,\nu) \in \mathcal{Y} \subseteq \R^{n_y}, \qquad n_y \ge n_p,
\label{eq:forward}
\end{equation}
which is treated as a black box that can be evaluated at arbitrary $p$ (and possibly differentiated) but not inverted in closed form. Hereafter, for a vector $x$, $n_x$ denotes its dimension, and vector components are indexed from zero. Unless stated otherwise, we write $f(p) = f(p,\nu_0)$ for a nominal nuisance configuration $\nu_0$, assume that $f$ is continuously differentiable with $\partial f/\partial p$ of full column rank on $\mathcal{P}$, and denote by $\mathcal{M} = f(\mathcal{P})$ the data manifold. Given an observation $y^\star$, the inverse problem is to find
\begin{equation}
p^\star = \operatorname*{argmin}_{p \in \mathcal{P}} \|f(p) - y^\star\|,
\label{eq:inverse}
\end{equation}
which coincides with the exact solution of $f(p^\star) = y^\star$ when $y^\star \in \mathcal{M}$, and otherwise accounts for observations that lie off the data manifold because of noise or model mismatch. In this work we restrict attention to forward maps that are injective on $\mathcal{P}$, so that the inverse problem has a unique exact solution; non-injective problems, in which several parameter vectors explain the same observation, are left for future work.

\paragraph{Bijective representations.}
We seek a parameterized diffeomorphism of the (zero-padded) observation space,
\begin{equation}
T_a: \R^{n_s} \to \R^{n_s}, \qquad (r, h) = T_a(y), \qquad n_r = n_p, \qquad n_r + n_h = n_s \ge n_y,
\label{eq:T}
\end{equation}
with trainable parameters $a \in \R^{n_a}$. Its leading $n_p$ output coordinates, $r \in \mathcal{R} \subset \R^{n_p}$, form the representation used for inference, while the remaining coordinates, $h \in \mathcal{H}$, collect the variability of $y$ that is not needed for recovering $p$ and may be used, e.g., for generative modeling or uncertainty characterization. The central object is the composite map
\begin{equation}
g = r \circ f : \mathcal{P} \to \R^{n_p}, \qquad J(p) = \frac{\partial g}{\partial p} = \frac{\partial r}{\partial y}\,\frac{\partial f}{\partial p}.
\label{eq:g}
\end{equation}
We require the representation to satisfy two conditions:
\begin{enumerate}
\item[(i)] \emph{Bijectivity with bounded distortion}: $g$ is injective on $\mathcal{P}$, and its Jacobian is uniformly nondegenerate and bounded, $\det J \ge m > 0$ and $\|\partial g/\partial p_l\| \le v$ for all $l$ and all $p \in \mathcal{P}$, so that $g$ and its inverse are Lipschitz continuous;
\item[(ii)] \emph{Insensitivity to irrelevant perturbations}: $r$ varies little under perturbations of $y$ that do not correspond to changes of $p$, i.e., $r(f(p,\nu)) \approx r(f(p,\nu + \delta\nu))$ for admissible $\delta\nu$. When the structure of the nuisance is unknown, we use the generic surrogate $(\partial r/\partial y)\, P_N \approx 0$ on $\mathcal{M}$, where $P_N = I - \frac{\partial f}{\partial p}\big(\frac{\partial f}{\partial p}\big)^{+}$ is the orthogonal projector onto the normal space of $\mathcal{M}$ and $(\cdot)^+$ denotes the Moore--Penrose pseudoinverse.
\end{enumerate}
We refer to the problem of learning representations that satisfy (i) and (ii) as \emph{bijective representation learning}. Condition (i) is targeted explicitly by the training objectives of Sec.~\ref{sec:training}. It is not guaranteed by the architecture, and whether training attains it depends on model capacity, data coverage, and optimization.

\paragraph{Inference in representation space.}
Given $y^\star$, we compute $r^\star = r(y^\star)$ and solve the square system $g(p) = r^\star$ by the damped iteration
\begin{equation}
p^{(t+1)} = \Pi_{\mathcal{P}}\!\left(p^{(t)} + \gamma_t\, J^{-1}\big(r^\star - g(p^{(t)})\big)\right),
\label{eq:newton}
\end{equation}
where $J$ is evaluated at $p^{(t)}$, the step length $\gamma_t \in (0,1]$ restricts the update to the region in which the linearization is trustworthy, and $\Pi_{\mathcal{P}}$ clips the iterate onto $\mathcal{P}$. Because $J$ is square, the Gauss--Newton step $J^{+}(r^\star - g)$ with $J^{+} = (J^\top J)^{-1} J^\top$ coincides with the Newton step for $g(p) = r^\star$. Each iteration requires $f(p^{(t)})$ and $\partial f/\partial p$ at $p^{(t)}$---that is, $n_p+1$ forward evaluations when finite differences are used---together with one evaluation of $T_a$ and of $\partial r/\partial y$. The initial estimate $p^{(0)}$ may be produced by any approximate inverse; in our experiments, we use a small regressor $r \mapsto p$ trained on the exploration data.

\paragraph{Elementary properties.}
The following elementary facts motivate conditions (i) and (ii).

\begin{proposition}[No spurious stationary points]
\label{prop:stationary}
Let $g$ be continuously differentiable on an open set containing $\mathcal{P}$, with $\det J(p) \ne 0$ for all $p \in \mathcal{P}$. Then, for every $r^\star \in \R^{n_p}$, every stationary point of $\Phi(p) = \tfrac{1}{2}\|g(p) - r^\star\|^2$ in the interior of $\mathcal{P}$ is an exact solution of $g(p) = r^\star$. If, in addition, $g$ is injective on $\mathcal{P}$ and $r^\star \in g(\mathcal{P})$, this solution is unique.
\end{proposition}
\begin{proof}
$\nabla\Phi = J^\top (g(p) - r^\star)$ vanishes if and only if $g(p) = r^\star$, because $J$ is nonsingular. Uniqueness follows from injectivity.
\end{proof}

\begin{remark}[Contrast with observation space]
\label{rem:dataspace}
For the data-space misfit $\Phi_{\mathcal{Y}}(p) = \tfrac{1}{2}\|f(p) - y^\star\|^2$, the gradient $(\partial f/\partial p)^\top (f(p) - y^\star)$ vanishes whenever the residual is orthogonal to the tangent space of $\mathcal{M}$ at $f(p)$. For $n_y > n_p$ and curved $\mathcal{M}$ this generically happens at points other than the solution---for a spiral, for instance, at points of a neighboring winding (Sec.~\ref{sec:spiral}). Condition (i) removes such points by reducing the problem to a square system with a nonsingular Jacobian. Note, however, that a nonvanishing Jacobian determinant guarantees only \emph{local} injectivity; on a bounded domain, global injectivity requires additional control, e.g., injectivity on the boundary~\cite{meisters.duke.1963} or stronger conditions on the Jacobian~\cite{gale.mathann.1965, parthasarathy.lnm.1983}. This motivates the distortion bound in (i) and the curriculum of Sec.~\ref{sec:training}.
\end{remark}

\begin{remark}[Newton flow]
\label{rem:newtonflow}
In the limit of small steps, Eq.~\eqref{eq:newton} follows the continuous Newton flow $\dot p = J^{-1}\big(r^\star - g(p)\big)$, along which $g(p(t)) - r^\star = e^{-t}\big(g(p(0)) - r^\star\big)$; that is, in representation space the iterate moves along a straight segment toward $r^\star$~\cite{deuflhard.springer.2011}. Hence, if this segment lies in $g(\mathcal{P})$---for instance, if $g(\mathcal{P})$ is convex---the flow reaches $p^\star$ without encountering the boundary of $\mathcal{P}$, and near $p^\star$ the undamped iteration converges quadratically~\cite{nocedal.springer.2006}. Convexity of $g(\mathcal{P})$ is not enforced in our training; contact with the boundary is handled by clipping.
\end{remark}

\begin{proposition}[Sensitivity to observation errors]
\label{prop:sensitivity}
Let $p_0$ be an interior point of $\mathcal{P}$, $y_0 = f(p_0)$, and $J_f = \frac{\partial f}{\partial p}(p_0)$ of full column rank. Let $r$ be differentiable at $y_0$ with $J = \frac{\partial r}{\partial y} J_f$ nonsingular, and let $P_N = I - J_f J_f^{+}$. Then the representation-space inverse $\hat p(y) = g^{-1}(r(y))$, defined in a neighborhood of $y_0$, has the derivative $A = J^{-1}\frac{\partial r}{\partial y}$ at $y_0$, which satisfies $A J_f = I$ and
\begin{equation}
A A^\top = J_f^{+}\big(J_f^{+}\big)^\top + Z Z^\top, \qquad Z = J^{-1}\,\frac{\partial r}{\partial y}\,P_N.
\label{eq:sensitivity}
\end{equation}
Consequently, $\|A\|_2 \ge \|J_f^{+}\|_2 = 1/\sigma_{\min}(J_f)$, and $A = J_f^{+}$ if and only if $\frac{\partial r}{\partial y} P_N = 0$.
\end{proposition}
\begin{proof}
Differentiating $g(\hat p(y)) = r(y)$ gives $J A = \partial r/\partial y$, hence $A = J^{-1}\partial r/\partial y$ and $A J_f = J^{-1} J = I$. With $J_f J_f^{+} = I - P_N$, we obtain $A(I - P_N) = A J_f J_f^{+} = J_f^{+}$ and therefore $A = J_f^{+} + Z$. Since $J_f^{+} P_N = 0$, the cross terms in $A A^\top$ vanish, which gives Eq.~\eqref{eq:sensitivity}. The norm inequality follows from $A A^\top \succeq J_f^{+}(J_f^{+})^\top$, and $Z = 0$ if and only if $(\partial r/\partial y) P_N = 0$ because $J$ is nonsingular.
\end{proof}

Proposition~\ref{prop:sensitivity} separates the propagation of an observation error $\delta y$ into two parts. The tangential part is propagated exactly as by the linearized least-squares estimator, $J_f^{+}\delta y$, independently of $r$: it reflects the intrinsic conditioning of the forward problem and cannot be reduced by any choice of representation. The normal part is propagated through $Z$, which vanishes precisely when condition (ii) holds. Condition (ii) therefore makes the representation-space inversion, to first order, as stable as orthogonal projection onto $\mathcal{M}$---the optimal encoder in the sense of~\cite{hauschultz.arxiv.2022, gropp.arxiv.2020}. An empirical bound of the form $\|\Delta p\| \lesssim \eta\, \|\Delta y\|$ can thus approach, but not fall below, $\sup_{p\in\mathcal{P}} 1/\sigma_{\min}\big(\partial f/\partial p\big)$. Beyond first order, the uniqueness of the least-squares projection itself is guaranteed only within the reach of $\mathcal{M}$~\cite{federer.tams.1959, hauschultz.arxiv.2022}, which bounds the range of perturbations over which such a comparison is meaningful.

\section{Twincher architecture}
\label{sec:architecture}

\subsection{Layered structure}

\begin{figure}[htbp]
  \centering
  \includegraphics[width=\linewidth]{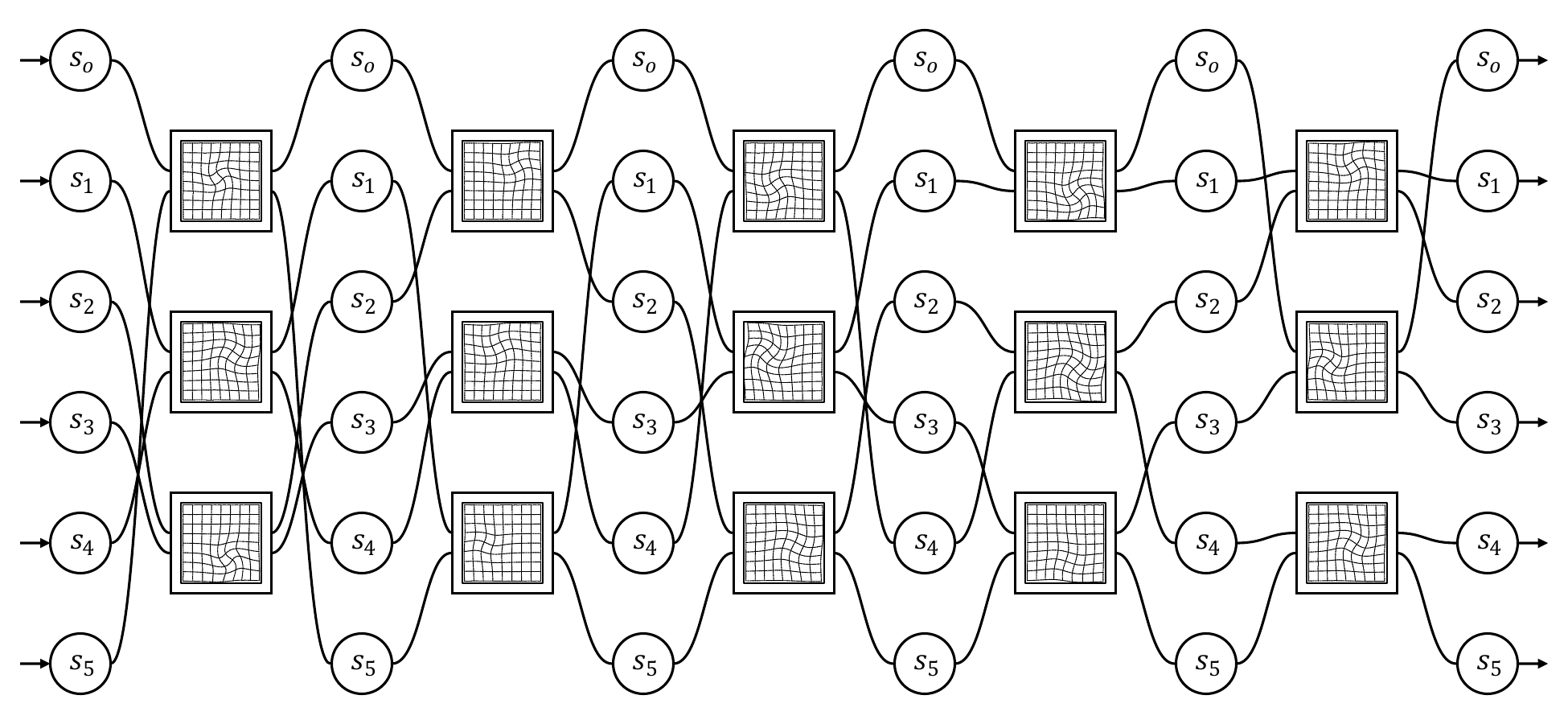}
  \caption{Layered structure of a twincher, shown for $n_s = 6$ and $n_l = 5$ layers. Circles denote the components of $s$ at each layer; boxes denote twinches acting on pairs of components, with the pairing following a round-robin schedule.}
  \label{fig:structure}
\end{figure}

A twincher is a parameterized diffeomorphism $T_a: s^{\mathrm{in}} \mapsto s^{\mathrm{out}}$ of $\R^{n_s}$ (Eq.~\eqref{eq:T}) composed of $n_l$ layers, $T_a = \phi_{n_l} \circ \cdots \circ \phi_1$, with states $s^{j+1} = \phi_{j+1}(s^{j})$, $s^0 = s^{\mathrm{in}}$, and $s^{n_l} = s^{\mathrm{out}}$. Each layer partitions the $n_s$ coordinates into disjoint pairs and applies an elementary two-dimensional diffeomorphism, a twinch (Sec.~\ref{sec:twinch}), to each pair (Fig.~\ref{fig:structure}). The pairing changes from layer to layer according to a round-robin schedule, as used for pairing indices in parallel Jacobi methods~\cite{brent.sisc.1985}; for even $n_s$, every pair of coordinates then interacts directly once in every $n_s - 1$ consecutive layers. More general wiring patterns and heterogeneous primitives are possible but are not considered here. Observations are zero-padded to dimension $n_s \ge n_y$ at the input, which provides additional room for the transformation, similarly in spirit to augmented neural ODEs~\cite{dupont.neurips.2019}.

\subsection{The twinch primitive}
\label{sec:twinch}

\begin{figure}[htbp]
  \centering
  \includegraphics[width=0.5\linewidth]{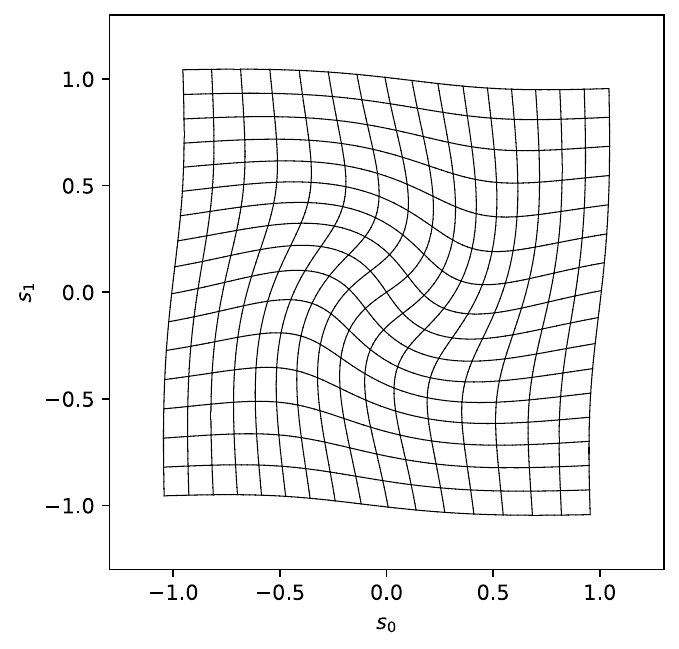}
  \caption{A single twinch: image of a uniform grid on $s^{\mathrm{in}} \in [-1,1]^2$ in the $s^{\mathrm{out}}$ plane.}
  \label{fig:twinch}
\end{figure}

A twinch acts on a pair of coordinates $(s_0, s_1)$ (local indices) as a rotation about a center $c = (c_0, c_1)$ by an angle that decays with the distance $\rho$ from the center:
\begin{equation}
\begin{gathered}
\begin{pmatrix} s^{\mathrm{out}}_0 \\ s^{\mathrm{out}}_1 \end{pmatrix}
= R(\alpha) \begin{pmatrix} s^{\mathrm{in}}_0 - c_0 \\ s^{\mathrm{in}}_1 - c_1 \end{pmatrix}
+ \begin{pmatrix} c_0 \\ c_1 \end{pmatrix}, \qquad
R(\alpha) = \begin{pmatrix} \cos\alpha & \sin\alpha \\ -\sin\alpha & \cos\alpha \end{pmatrix},\\
\alpha = t \exp\!\left(-\frac{\rho^2}{w^2}\right), \qquad \rho^2 = (s^{\mathrm{in}}_0 - c_0)^2 + (s^{\mathrm{in}}_1 - c_1)^2,
\end{gathered}
\label{eq:twinch}
\end{equation}
where $t$ and $w$ control the amplitude and the width of the twist, respectively. The effect resembles twisting a pinched piece of fabric; the conflation of the two words gives the primitive its name. For efficiency on GPUs, we replace the transcendental functions by rational and polynomial counterparts,
\begin{equation}
\sin\alpha = \frac{\kappa(3 - \kappa^2)}{2}, \qquad \cos\alpha = \sqrt{1 - \sin^2\alpha}, \qquad
\kappa = \frac{t\, w^4}{\left(\rho^2 + w^2\right)^2},
\label{eq:twinch-fast}
\end{equation}
where $|t| < 1$ ensures $|\kappa| < 1$ and hence $|\sin\alpha| < 1$. This variant is illustrated in Fig.~\ref{fig:twinch}.

\paragraph{Exact invertibility and volume preservation.}
Because $R(\alpha)$ is orthogonal, a twinch preserves the distance $\rho$ to its center, so that the angle $\alpha(\rho)$ can be recomputed from the output. The inverse is therefore available in closed form at the cost of a forward evaluation, $s^{\mathrm{in}} - c = R(\alpha(\rho))^\top (s^{\mathrm{out}} - c)$. In polar coordinates centered at $c$, a twinch is the twist map $(\rho, \varphi) \mapsto (\rho, \varphi - \alpha(\rho))$, whose Jacobian determinant is identically one. Explicitly, with $x = s^{\mathrm{in}} - c$ and the antisymmetric generator $K = \left(\begin{smallmatrix} 0 & 1 \\ -1 & 0 \end{smallmatrix}\right)$, for which $\mathrm{d}R/\mathrm{d}\alpha = R(\alpha) K$, the Jacobian of a twinch is $R(\alpha)\big(I + \rho^{-1}\alpha'(\rho)\, K x x^\top\big)$, and $\det\big(I + \rho^{-1}\alpha'(\rho)\, K x x^\top\big) = 1 + \rho^{-1}\alpha'(\rho)\, x^\top K x = 1$ because $K$ is antisymmetric. Consequently, every twincher is a volume-preserving diffeomorphism, $\det\big(\partial s^{\mathrm{out}}/\partial s^{\mathrm{in}}\big) \equiv 1$, with an exact inverse. Locally, a twinch acts as a rotation composed with an area-preserving shear: it can redistribute sensitivity between coordinates but never contracts or expands area. The design intent is an inductive bias toward realizing a required local Jacobian by localized rotation and shear rather than by stretching or compressing, so that the information about $p$ contained in $s^{\mathrm{in}}$ can be gathered into the leading $n_p$ coordinates of $s^{\mathrm{out}}$ without the associated Jacobian block collapsing. Volume preservation of $T_a$ does not, by itself, constrain this block, which is why the objectives of Sec.~\ref{sec:training} are required.

\paragraph{Parameterization.}
Each twinch has four parameters: two for its center, one for its amplitude, and one for its width. We have found empirically that this suffices to construct a broad range of well-behaved diffeomorphisms. We optimize unconstrained parameters and map them smoothly onto the admissible ranges of the twinch parameters, a standard reparameterization that removes box constraints from the optimization. With $(a_0, a_1, a_2, a_3)$ denoting the four consecutive components of $a$ that belong to one twinch,
\begin{equation}
\begin{aligned}
&c_0 = \frac{2 a_0}{\sqrt{1 + a_0^2}} \in (-2, 2), \qquad
 c_1 = \frac{2 a_1}{\sqrt{1 + a_1^2}} \in (-2, 2), \qquad
 t = \frac{0.9\, a_3}{\sqrt{1 + a_3^2}} \in (-0.9, 0.9),\\
&w = \frac{2 w_s + a_2^2(w_u + w_l) + a_2 \sqrt{a_2^2(w_u - w_l)^2 + 4(w_u - w_s)(w_s - w_l)}}{2(1 + a_2^2)} \in (w_l, w_u),
\end{aligned}
\label{eq:param-map}
\end{equation}
where the hyperparameters $w_l < w_s < w_u$ set the lower bound, the initial value ($w = w_s$ at $a_2 = 0$), and the upper bound of $w$; the map $a_2 \mapsto w$ is strictly increasing. The components of $s$ are kept within a window in which the twinches remain effective by a penalty term (Appendix~\ref{app:derivatives}).

\subsection{Derivative tensors}
\label{sec:derivatives-summary}

The objectives of Sec.~\ref{sec:training} depend on Jacobians of $r$ rather than on its values: no target values are prescribed for $r$, which leaves the representation free to adopt any values compatible with conditions (i) and (ii). Optimizing such objectives requires derivatives of functions of a Jacobian with respect to the parameters $a$, as in double backpropagation and Sobolev training~\cite{drucker.tnn.1992, czarnecki.neurips.2017}. We implement three passes that exploit the block sparsity of twinch layers and their exact invertibility (details in Appendix~\ref{app:derivatives}):
\begin{itemize}
\item \emph{forward-query} propagates $n_p$ tangent vectors $Q = \partial s/\partial p$ through the layers, i.e., forward-mode differentiation~\cite{griewank.siam.2008, baydin.jmlr.2018}, and yields $J = \partial r/\partial p$;
\item \emph{backward-variance} propagates $n_p$ cotangent vectors $V = \partial r/\partial s$ backward, i.e., reverse-mode differentiation, and yields $\partial r/\partial y$ for inference;
\item \emph{backward-query} differentiates a loss $L(J)$ with respect to $a$ by reverse-mode differentiation of the forward-query computation.
\end{itemize}
Because layer inputs are reconstructed from layer outputs on the fly with the closed-form inverse, as in reversible networks~\cite{gomez.neurips.2017, jacobsen.iclr.2018}, the memory footprint for a batch of $n_c$ cases is $O(n_c n_s n_p)$, independently of the depth $n_l$, while the cost of each pass is $O(n_c n_s n_l n_p)$, i.e., $O(n_p)$ times that of a plain forward evaluation, in line with the cost of computing $n_p$ directional derivatives by automatic differentiation~\cite{griewank.siam.2008}. Generic automatic-differentiation frameworks would, by default, store the intermediate states of all layers.

\section{Training: the hyper-sail procedure}
\label{sec:training}

We now describe the training procedure used in our experiments, available as \texttt{arch='hs'} (hyper-sail) in the package~\cite{twincher-code}. Training data consist of parameter vectors $p_i$, the corresponding observations $y_i = f(p_i)$, and Jacobians $(\partial f/\partial p)_i$, obtained by differentiating the forward model when possible and otherwise by finite differences or by interpolation of the collected data. We use Jacobians estimated through finite differences unless stated otherwise.

\subsection{Objectives}

\paragraph{Nondegeneracy and bounded distortion.}
Condition (i) is promoted by
\begin{equation}
\begin{aligned}
& L_{\mathrm{hs}} = L_{\det} + L_{\mathrm{trim}},\\
& L_{\det} = \lambda_{\det}\, \mathrm{ReLU}\big(m_{\det} - \det J\big)^2, \\
& L_{\mathrm{trim}} = \lambda_{\mathrm{trim}} \sum_{l < n_p} \mathrm{ReLU}\left(\left\| \frac{\partial r}{\partial p_l}\right\| - v_{\mathrm{trim}}\right)^2,
\end{aligned}
\label{eq:loss-hs}
\end{equation}
where $\mathrm{ReLU}(\xi) = \max(0, \xi)$, $m_{\det}$ is a margin for $\det J$, $v_{\mathrm{trim}}$ is a threshold on the column norms of $J$, and $\lambda_{\det}$ and $\lambda_{\mathrm{trim}}$ are weights. The determinant term keeps $J$ away from singularity and fixes its orientation; the trimming term bounds the distortion of $g$ and, together with the curriculum below, counteracts folding of $g(\mathcal{P})$ onto itself, which nondegeneracy alone does not prevent on a bounded domain (Remark~\ref{rem:dataspace}).

\paragraph{Insensitivity to normal perturbations.}
Condition (ii) is promoted by penalizing directional derivatives of $r$ along random unit vectors in the normal space of $\mathcal{M}$:
\begin{equation}
\begin{aligned}
& L_{\mathrm{noise}} = \lambda_{\mathrm{noise}} \sum_{k < n_p} \left\|\frac{\partial r}{\partial y}\, \hat{y}_k\right\|^2, \\
& \hat{y}_k = \mathrm{normalize}\big(P_N\, \tilde{y}_k\big), \qquad \tilde{y}_k \sim U\big([-1,1]^{n_y}\big),
\end{aligned}
\label{eq:loss-noise}
\end{equation}
where $\mathrm{normalize}(\xi) = \xi/\|\xi\|$, $P_N$ is defined in Sec.~\ref{sec:formulation}, and new directions are drawn for each batch element and iteration. This provides a stochastic surrogate for $\|(\partial r/\partial y) P_N\|_F^2$, in the spirit of Hutchinson-type estimators~\cite{hutchinson.1989, hoffman.arxiv.2019}, without forming an $n_y \times (n_y - n_p)$ basis of the normal space. The term can be viewed as a contractive penalty~\cite{rifai.icml.2011} restricted to the directions that carry no information about $p$, or as a tangent-propagation-type constraint~\cite{simard.neurips.1992} imposed on the normal rather than the tangent directions. By Proposition~\ref{prop:sensitivity}, driving it to zero makes the inversion first-order equivalent to least-squares projection onto $\mathcal{M}$.

\paragraph{Domain-growing curriculum.}
Empirically, $\det J$ varies smoothly under the inductive bias of twinchers, so that $\det J > 0$ can be enforced from a finite sample of points with a sufficiently large margin $m_{\det}$; that is, $\det J$ is unlikely to change sign between samples. A more serious difficulty is that optimization frequently gets trapped in configurations in which $g(\mathcal{P})$ folds over itself. We therefore train on a growing hypercube $[-R, R]^{n_p}$: starting from $R = R_{\mathrm{step}}$, we increase $R$ by $R_{\mathrm{step}}$ whenever $\det J \ge m_{\mathrm{clearance}}$ holds for all samples of the current hypercube (optionally over several consecutive iterations), until $R = 1$. For sufficiently small $R$, nondegeneracy at the center implies injectivity on $[-R,R]^{n_p}$ by the inverse function theorem. Growing the domain while maintaining nondegeneracy can be pictured as inflating a bubble of $\det J > 0$ that pushes regions with $\det J \le 0$ out of $\mathcal{P}$ and progressively flattens the manifold $g(\mathcal{P})$---the picture that inspired the name \emph{hyper-sail}. The procedure is a form of curriculum learning~\cite{bengio.icml.2009} and is closely related to continuation methods~\cite{allgower.springer.1990}.

\subsection{Training iteration}

We initialize a twincher with $n_s \ge n_y$, allocate batch buffers for $n_p$ tangent directions (the \texttt{Shuttle} object in the package), set $R = R_{\mathrm{step}}$, and repeat the following steps. Tensor components with indices beyond $n_y$ are zero-padded, and the batch index is omitted.
\begin{enumerate}
  \item Sample $p \sim U\big([-R, R]^{n_p}\big)$ and obtain $y = f(p)$ and $\partial f/\partial p$.

  \item Set $s^0 = y$, $Q^0 = \partial f/\partial p$, and perform a forward-query pass. It yields $s^{n_l}$ and $Q^{n_l}$, whose leading $n_p \times n_p$ block is $J = \partial r/\partial p$.

  \item \label{step:hs-backward} Set $b^{n_l} = 0$ and $B^{n_l} = \partial L_{\mathrm{hs}}/\partial J = B_{\det} + B_{\mathrm{trim}}$ (zero-padded to $n_s \times n_p$), with
  \begin{equation}
  \begin{aligned}
  & B_{\det} = -2\lambda_{\det}\, \mathrm{ReLU}\big(m_{\det} - \det J\big)\, \det(J)\, J^{-\top},\\
  & \big(B_{\mathrm{trim}}\big)_{k,l} = 2\lambda_{\mathrm{trim}}\, \mathrm{ReLU}\big(\|J_{:,l}\| - v_{\mathrm{trim}}\big)\, \frac{J_{k,l}}{\|J_{:,l}\|},
  \end{aligned}
  \end{equation}
  and perform a backward-query pass to obtain $\partial L_{\mathrm{hs}}/\partial a$.

  \item Draw $\tilde{y}_k \sim U\big([-1,1]^{n_y}\big)$ and form $\hat{y}_k = \mathrm{normalize}\Big(\tilde{y}_k - \frac{\partial f}{\partial p}\Big(\big(\frac{\partial f}{\partial p}\big)^{+} \tilde{y}_k\Big)\Big)$ for $k = 0, \ldots, n_p - 1$.

  \item Set $s^0 = y$, $Q^0 = [\hat{y}_0, \ldots, \hat{y}_{n_p-1}]$, and perform a forward-query pass; the leading block of $Q^{n_l}$ is now $(\partial r/\partial y)\hat{Y}$ with $\hat Y = [\hat{y}_0, \ldots, \hat{y}_{n_p-1}]$.

  \item Set $b^{n_l} = 0$ and $B^{n_l} = 2\lambda_{\mathrm{noise}}(\partial r/\partial y)\hat{Y}$ (zero-padded), and perform a backward-query pass to obtain $\partial L_{\mathrm{noise}}/\partial a$.

  \item Update the parameters with an optimizer of choice, $a \leftarrow \mathrm{Optimizer}\big(a,\, \partial L_{\mathrm{hs}}/\partial a + \partial L_{\mathrm{noise}}/\partial a\big)$.

  \item If $\det J \ge m_{\mathrm{clearance}}$ for all samples in step~\ref{step:hs-backward} (optionally over several consecutive iterations), set $R \leftarrow \min(1, R + R_{\mathrm{step}})$.
\end{enumerate}
This is a minimal version of the procedure; the package contains several straightforward extensions that are omitted here for clarity.

\subsection{Variants and extensions}
\label{sec:variants}

The formulation of Sec.~\ref{sec:formulation} does not depend on the specific primitive or objectives used above. In principle, other invertible families---coupling flows~\cite{dinh.arxiv.2016, kingma.arxiv.2018}, CPAB transformations~\cite{freifeld.iccv.2015}, or invertible residual networks~\cite{behrmann.icml.2019}---could serve as building blocks, provided that (a) invertibility is preserved robustly for all parameter values, e.g., $|\det(\partial T_a/\partial y)| \ge \epsilon > 0$, which twinchers satisfy with $\epsilon = 1$; (b) $T_a$ is differentiable with respect to $a$; and (c) the family is expressive enough to realize nontrivial rearrangements of coordinates. We explored several architectural and training variants on synthetic problems. In our experience, the choice of primitive strongly affects how often training gets trapped in folded configurations, which suggests that the twinch provides a beneficial inductive bias for this objective; a systematic comparison is left for future work.

The following objectives may replace or complement those above; they are \emph{not} used in the experiments of Sec.~\ref{sec:experiments}. Bijectivity can be promoted by a co-Lipschitz (lower Lipschitz) condition,
\begin{equation}
\|g(p) - g(p')\| \ge M\, \|p - p'\|, \qquad \forall\, p, p' \in \mathcal{P},
\label{eq:colipschitz}
\end{equation}
with $M > 0$, where pairs $(p, p')$ are selected or adversarially optimized to expose near-violations while $a$ is updated to increase the margin $M$; by minimizing the probability of such violations, $\mathbb{P}\big(\|g(p) - g(p')\| < M\|p - p'\|\big)$; or by geometric regularizers that promote unique projections, such as reach-based penalties~\cite{hauschultz.arxiv.2022}. Robustness can be targeted at a known nuisance structure by selecting perturbations $\delta\nu$ that maximize $\|\delta r\| = \|(\partial r/\partial y)(\partial f/\partial \nu)\,\delta\nu\|$, in the spirit of adversarial training~\cite{goodfellow.arxiv.2014, madry.arxiv.2017} and distributionally robust optimization~\cite{rahimian.arxiv.2019}, while updating $a$ to minimize it; exploring perturbations jointly in $\nu$ and $p$ yields a worst-case sensitivity analysis over the joint space. Finally, because training reveals where $\det J$ approaches its margin or where the sensitivity of $r$ is largest, these quantities could guide the acquisition of new forward evaluations, i.e., active learning~\cite{settles.tr.2009}. We refer to this as the \emph{dynamic} regime, in contrast to the \emph{static} regime used below, in which the training set is sampled once; its evaluation is left for future work.

\section{Experiments}
\label{sec:experiments}

The open-source Python package~\cite{twincher-code} implements the hyper-sail procedure on CPU and GPU, together with the computational primitives (forward- and backward-query passes, batch buffers) needed to construct other training strategies, and reproduces the experiments below. All experiments use the static regime: the training set is sampled uniformly at random at initialization, without feedback-driven data acquisition.

\subsection{Illustrative example: a spiral}
\label{sec:spiral}

\begin{figure}[htbp]
\centering
\includegraphics[width=1.0\columnwidth]{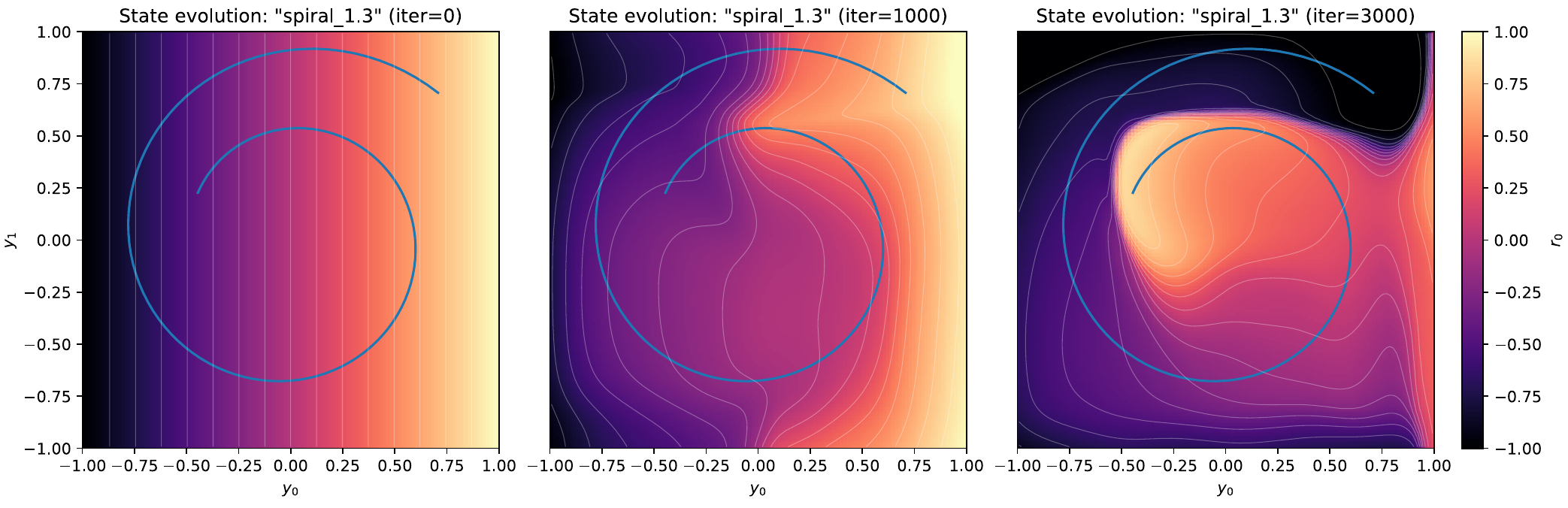}
\caption{Learned representation for the spiral example ($n_p = 1$, $n_y = 2$). Color shows the scalar representation $r_0$ as a function of $(y_0, y_1)$ before, during and after training, and thin lines show its iso-contours. The blue curve is the data manifold $\mathcal{M} = f(\mathcal{P})$. After training each iso-contour intersects it exactly once.}
\label{fig:spiral}
\end{figure}

We first consider a scalar parameter $p$ that traces a spiral in the plane (blue curve in Fig.~\ref{fig:spiral}), so that $n_p = n_r = 1$ and $n_y = 2$. In observation space, the least-squares misfit has spurious local minima: for a target on one winding, points on the neighboring windings that are locally closest to the target are stationary points of the misfit (Remark~\ref{rem:dataspace}). Figure~\ref{fig:spiral} shows the scalar representation $r_0(y_0, y_1)$ learned by a twincher, with thin lines indicating its iso-contours. After training, every iso-contour crosses the spiral exactly once, so that $r_0$ restricted to the spiral is strictly monotone in $p$, i.e., $g = r \circ f$ is a bijection. Moreover, the iso-contours cross the spiral nearly perpendicularly, which corresponds to insensitivity of $r_0$ to normal perturbations (condition (ii)).

\subsection{Synthetic benchmark: harmonic entanglers}
\label{sec:entangler}

\paragraph{Benchmark family.}
To study inversion under controlled nonlinearity, we introduce \emph{harmonic entanglers}, a family of random smooth embeddings $E: [-1,1]^{n_p} \to (-1,1)^{n_y}$ of tunable complexity. The parameter vector is first embedded into an $n_y$-dimensional state,
\begin{equation}
x^{(0)} = \big(p_0, \ldots, p_{n_p-1}, \xi_{n_p}, \ldots, \xi_{n_y-1}\big),
\end{equation}
where the padding components $\xi_i \sim U(-1,1)$ are drawn once at initialization. The state is then transformed by $n_e$ invertible layers, $x^{(k)} \mapsto x^{(k+1)}$, and the observation is $y = x^{(n_e)}$. Each layer splits the state (of even dimension) into halves, $x^{(k)} = (x_1, x_2)$, and applies two affine coupling transformations in the style of RealNVP~\cite{dinh.arxiv.2016},
\begin{equation}
\begin{aligned}
v_1 &= x_1, & \qquad v_2 &= x_2 \odot \exp\!\big(\sigma_1(x_1)\big) + \tau_1(x_1),\\
z_2 &= v_2, & \qquad z_1 &= v_1 \odot \exp\!\big(\sigma_2(v_2)\big) + \tau_2(v_2),
\end{aligned}
\end{equation}
where $\odot$ denotes elementwise multiplication. Each component of $(z_1, z_2)$ is then mapped into $(-1,1)$ by the strictly monotone bijection $z \mapsto z/\sqrt{1+z^2}$, which prevents uncontrolled growth across layers, and the components are reordered by a fixed random permutation to form $x^{(k+1)}$. The functions $\sigma_1, \tau_1, \sigma_2, \tau_2: \R^{n_y/2} \to \R^{n_y/2}$ are independent random harmonic maps whose $j$-th components have the form
\begin{equation}
\frac{2}{n_y}\sum_{i=1}^{n_y/2} \sin\!\big(w_{ji} x_i + b_{ji}\big), \qquad
w_{ji} \sim U\big(-\pi w_{\mathrm{amp}},\, \pi w_{\mathrm{amp}}\big), \qquad
b_{ji} \sim U(-\pi, \pi),
\end{equation}
with coefficients drawn once at initialization. Since every constituent map is a diffeomorphism, $E$ is a smooth embedding of the parameter domain, and the amplitude $w_{\mathrm{amp}}$ controls the frequency of the modulation and hence the degree of nonlinear entanglement. Unless stated otherwise, $n_p = 2$, $n_y = 4$, $n_e = 3$, and $w_{\mathrm{amp}} \in [0.5, 1.5]$. Although entanglers are built from coupling layers, a learner is not required to invert this construction: it only needs to find coordinates in which $E$ becomes a well-conditioned bijection onto its image.

\begin{figure}[htbp]
\centering
\includegraphics[width=1.0\columnwidth]{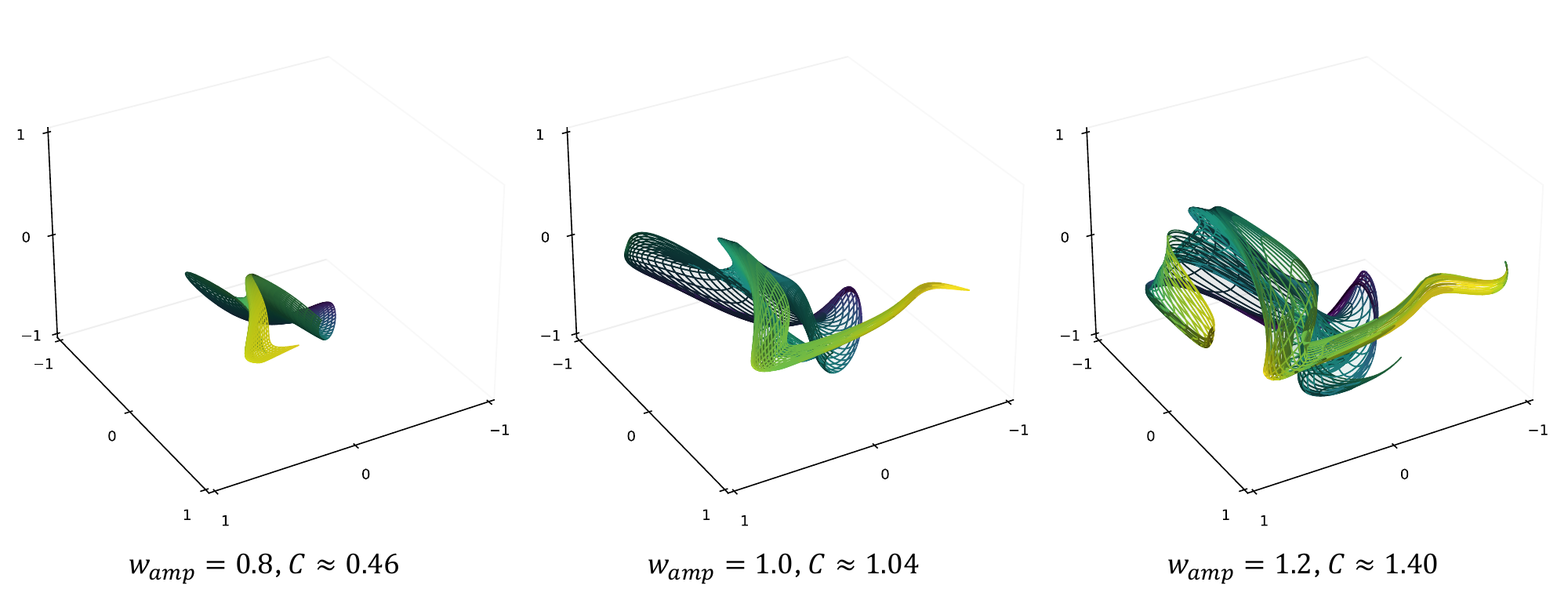}
\caption{A harmonic entangler for three values of $w_{\mathrm{amp}}$, with the corresponding estimated complexity $C$. The image of a grid on $p \in [-1,1]^2$ is shown in the $(y_0, y_1, y_2)$ space, and the fourth coordinate is represented by color.}
\label{fig:entangler-examples}
\end{figure}

\paragraph{Complexity measure.}
As a measure of problem complexity, we use
\begin{equation}
C := -\ln p_{\mathrm{rand}}^{\mathrm{success}},
\end{equation}
where $p_{\mathrm{rand}}^{\mathrm{success}}$ is the probability that a random initial point $p_{\mathrm{in}} \sim U\big([-1,1]^{n_p}\big)$ lies in the basin of attraction of a random target $p^\star \sim U\big([-1,1]^{n_p}\big)$ for iterative minimization of the observation-space misfit. We estimate it over a large set of trials with a clipped-step Gauss--Newton iteration in $\mathcal{Y}$. Thus, $C$ quantifies the nonconvexity of the observation-space inverse problem, with $C = 0$ corresponding to a globally convergent iteration. Figure~\ref{fig:entangler-examples} shows an example entangler for three values of $w_{\mathrm{amp}}$.

\paragraph{Protocol.}
Each trial draws a random entangler $E$. A learner first explores $E$ by querying it at points of its choice, subject to a budget of $\ncalls$ forward evaluations. It is then given a test batch of $10^3$ targets $y_i^{\mathrm{true}} = E(p_i^{\mathrm{true}})$ with undisclosed $p_i^{\mathrm{true}} \sim U\big([-1,1]^{n_p}\big)$ and must produce estimates $p_i^{(t)}$ such that $E(p_i^{(t)})$ matches $y_i^{\mathrm{true}}$. The initial estimate $p_i^{(0)}$ must be computed from $y_i^{\mathrm{true}}$ alone; each subsequent refinement step may use $n_p + 1$ forward evaluations per target to obtain $E(p_i^{(t)})$ and a finite-difference Jacobian. A trial is declared a \emph{success} if the worst-case residual after five refinement steps satisfies
\begin{equation}
\max_i \big\|E\big(p_i^{(5)}\big) - y_i^{\mathrm{true}}\big\| < 10^{-2},
\label{eq:success}
\end{equation}
and a \emph{failure} otherwise. We use the worst case rather than the mean because the goal is reliable inversion for every target.

\paragraph{Baseline: MLP with Gauss--Newton refinement.}
The baseline combines an amortized inverse with data-space refinement. During exploration, it evaluates $E$ at $\ncalls$ points $p_i \sim U\big([-1,1]^{n_p}\big)$, exhausting the budget in a single pass without feedback-driven selection, and trains an MLP $\R^{n_y} \to \R^{n_p}$ on the pairs $(y_i, p_i)$. The MLP has four hidden layers of width 16 with $\tanh$ activations and the output activation $1.5\tanh(\cdot)$, which makes the target range $[-1,1]^{n_p}$ attainable without saturating the output; for $n_p = 2$ and $n_y = 4$ this amounts to approximately $10^3$ trainable parameters. It is trained by minimizing the mean squared error with Adam~\cite{kingma.iclr.2015} (learning rate $10^{-3}$, up to $10^3$ epochs), with early stopping (patience of 10 epochs, 10\% validation split) when $\ncalls \ge 20$. At test time, $p^{(0)} = \mathrm{MLP}(y^{\mathrm{true}})$ is refined in observation space by a damped Gauss--Newton (Levenberg--Marquardt-type) iteration~\cite{levenberg.qam.1944, marquardt.siam.1963},
\begin{equation}
\big(J_E^\top J_E + \lambda I\big)\,\Delta p = J_E^\top\big(y^{\mathrm{true}} - E(p^{(t)})\big), \qquad
\big(J_E\big)_{ij} = \frac{\big[E(p^{(t)} + \delta e_j)\big]_i - \big[E(p^{(t)})\big]_i}{\delta},
\end{equation}
with $\lambda = 10^{-3}$, $\delta = 10^{-7}$, and $e_j$ the $j$-th canonical basis vector. The step is clipped, $\Delta p \leftarrow \Delta p\,\min\big(1, \Delta_{\max}/\|\Delta p\|\big)$ with $\Delta_{\max} = 0.1$, to remain within the region where the linearization is valid, and the iterate is projected onto the domain, $p^{(t+1)} = \mathrm{clip}\big(p^{(t)} + \Delta p, -1, 1\big)$.

\paragraph{Twincher learner.}
The Twincher learner uses a twincher with 64 layers and approximately $10^3$ trainable parameters, matching the capacity of the baseline. We used $n_s=4$, $m_{\det}=0.4$, and $v_{\mathrm{trim}}=1.3$. For optimization, we used a custom optimizer that converged faster than PyTorch's Adam ($lr=10^{-3}$, $\beta_1=0.99$, all other parameters at their default values) while yielding no significant difference in the final results. During exploration, twincher evaluates $E$ at points drawn uniformly at random within the same budget and is trained with the procedure of Sec.~\ref{sec:training}. Once a bijective representation has been formed, an auxiliary MLP with an architecture similar to that of the baseline is trained to approximate $r \mapsto p$ and provides $p^{(0)}$. If reaching a bijective representation fails (i.e. $R=1$ is not reached), the achieved representation is still used. Refinement uses the same damped iteration as the baseline, but in representation space, i.e., with $E$ replaced by $r \circ E$ and $y^{\mathrm{true}}$ by $r(y^{\mathrm{true}})$, and with $\Delta_{\max} = 1$.

\begin{figure}[htbp]
\centering
\includegraphics[width=0.75\columnwidth]{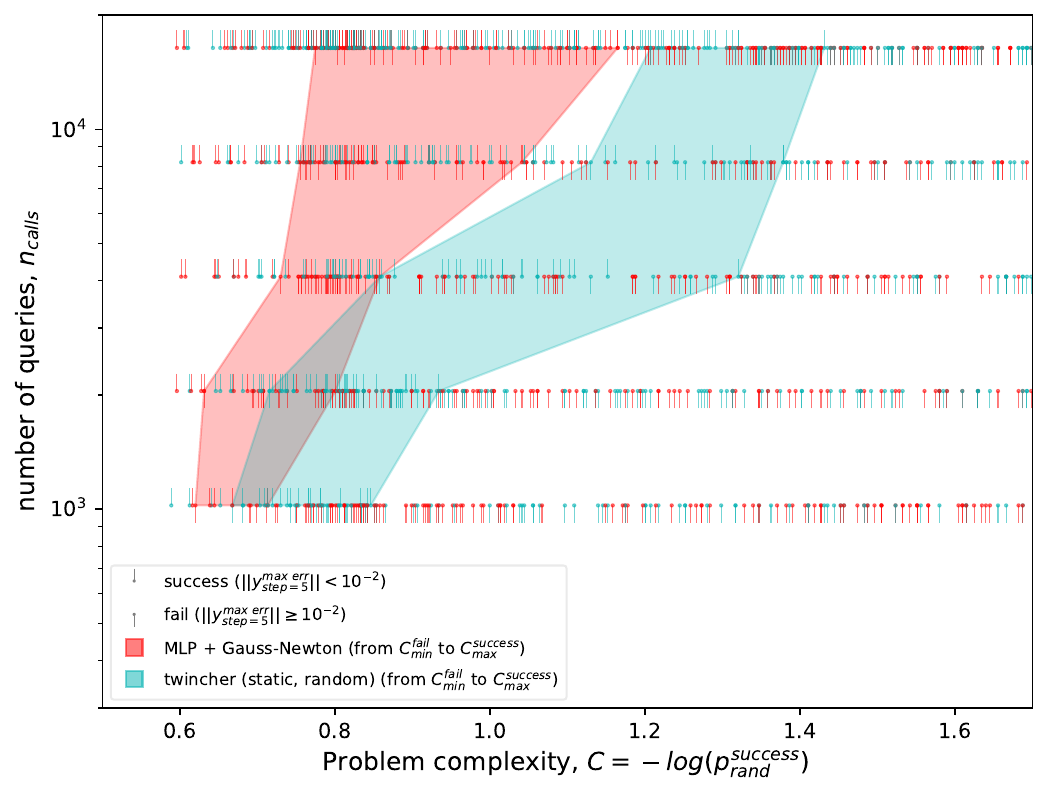}
\caption{Learning outcome as a function of problem complexity $C$ and query budget $\ncalls$ for the MLP\,+\,Gauss--Newton baseline and the Twincher learner. Each mark is one trial on a randomly drawn harmonic entangler; upward ticks denote success ($\max_i\|E(p_i^{(5)}) - y_i^{\mathrm{true}}\| < 10^{-2}$) and downward ticks denote failure. For each learner, the shaded band spans the transition region at each budget, from the smallest $C$ with an observed failure (left edge) to the largest $C$ with an observed success (right edge).}
\label{fig:entangler-phase}
\end{figure}

\paragraph{Results.}
Figure~\ref{fig:entangler-phase} summarizes the outcome across problem complexities $C$ and query budgets $\ncalls$. Each mark corresponds to one trial, i.e., a randomly drawn entangler with independently estimated $C$; upward and downward ticks denote success and failure according to Eq.~\eqref{eq:success}. For each learner and budget, the shaded band spans the transition region, from the smallest $C$ at which a failure was observed to the largest $C$ at which a success was observed. Problems to the left of a band were thus consistently solved at that budget, and problems to its right consistently failed. For $\ncalls \ge 4096$, the transition band of the Twincher learner lies at systematically higher $C$ than that of the baseline, which we attribute to the formation of a bijective representation, and it continues to shift with the budget. The slower shift between $\ncalls = 8192$ and $16384$ is consistent with the onset of an underparameterized regime, in which problems with $C \gtrsim 1.4$ require more trainable parameters to form a bijective representation.

\begin{figure}[htbp]
\centering
\includegraphics[width=0.65\columnwidth]{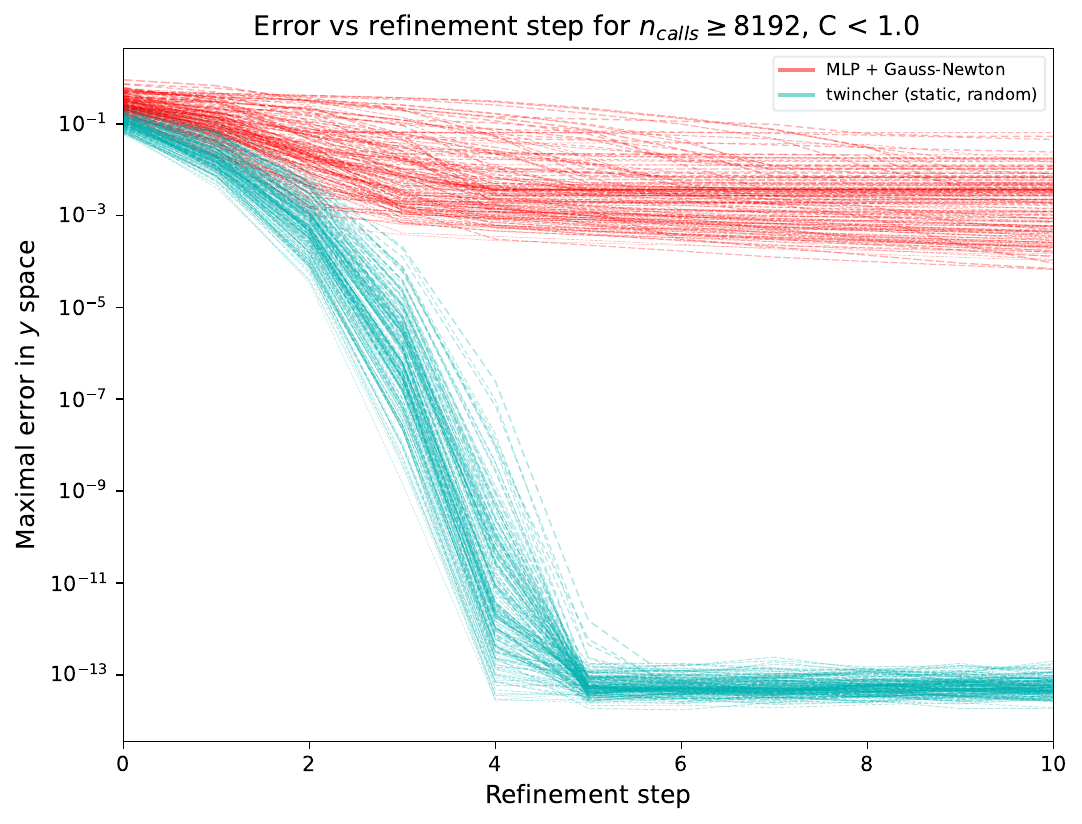}
\caption{Worst-case residual $\max_i \|E(p_i^{(t)}) - y_i^{\mathrm{true}}\|$ as a function of the refinement step $t$ for all trials with $C < 1$ and $\ncalls \ge 8192$. Each curve is one trial; line width increases with $C$, and dash density increases with $\log_2(\ncalls/1024)$.}
\label{fig:entangler-steps}
\end{figure}

Figure~\ref{fig:entangler-steps} shows the worst-case residual as a function of the refinement step for all trials with $C < 1$ and $\ncalls \ge 8192$, a regime in which the budget suffices to form a bijective representation (Fig.~\ref{fig:entangler-phase}). For the Twincher learner, the worst-case residual decreases to the level of machine precision within four to five steps in every trial, with the accelerating decrease characteristic of the quadratic convergence of Newton's method (Remark~\ref{rem:newtonflow}). The baseline levels off at worst-case residuals between approximately $10^{-4}$ and $10^{-2}$, indicating that for a fraction of the targets the data-space iteration is trapped or progresses slowly.

\begin{figure}[htbp]
\centering
\includegraphics[width=0.65\columnwidth]{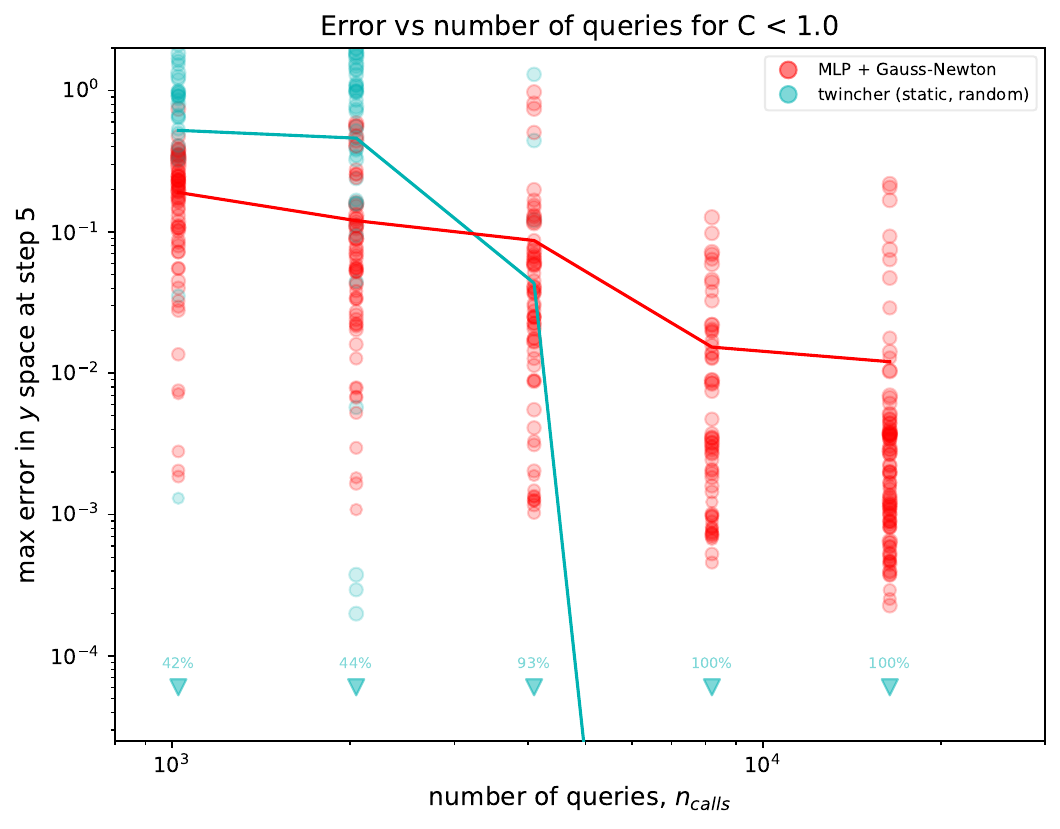}
\caption{Worst-case residual $\max_i \|E(p_i^{(5)}) - y_i^{\mathrm{true}}\|$ after five refinement steps as a function of the query budget $\ncalls$ for all trials with $C < 1$. Solid lines show the per-learner mean. Triangles at the lower boundary indicate budgets at which a fraction of the trials (given as a percentage) fell below the plotted range.}
\label{fig:entangler-budget}
\end{figure}

Figure~\ref{fig:entangler-budget} shows the worst-case residual after five refinement steps as a function of the budget for all trials with $C < 1$. For the baseline, the residual decreases gradually with the budget, qualitatively consistent with the power-law data scaling typical of statistical function approximators~\cite{hestness.arxiv.2017, kaplan.arxiv.2020, sharma.jmlr.2022}. The Twincher learner behaves differently: its outcome is bimodal. At $\ncalls = 1024$ and $2048$, the residual falls below the plotted range in about 40\% of the trials, while the remaining trials fail with residuals that exceed those of the baseline, so that the mean is dominated by failures. At $\ncalls = 4096$, the fraction of trials below the plotted range rises to 93\%, and from $\ncalls = 8192$ on, all trials reach machine precision. Such a sharp transition, rather than a gradual improvement, is what one expects if the accuracy is limited not by the approximation error of a learned model but by whether the learned representation is qualitatively correct: once $g$ is a well-conditioned bijection, the attainable accuracy is set by the forward model and by floating-point precision.

\subsection{Pose and shape parameters from noisy depth maps}
\label{sec:hinge}

\begin{figure}[htbp]
\centering
\includegraphics[width=1.0\columnwidth]{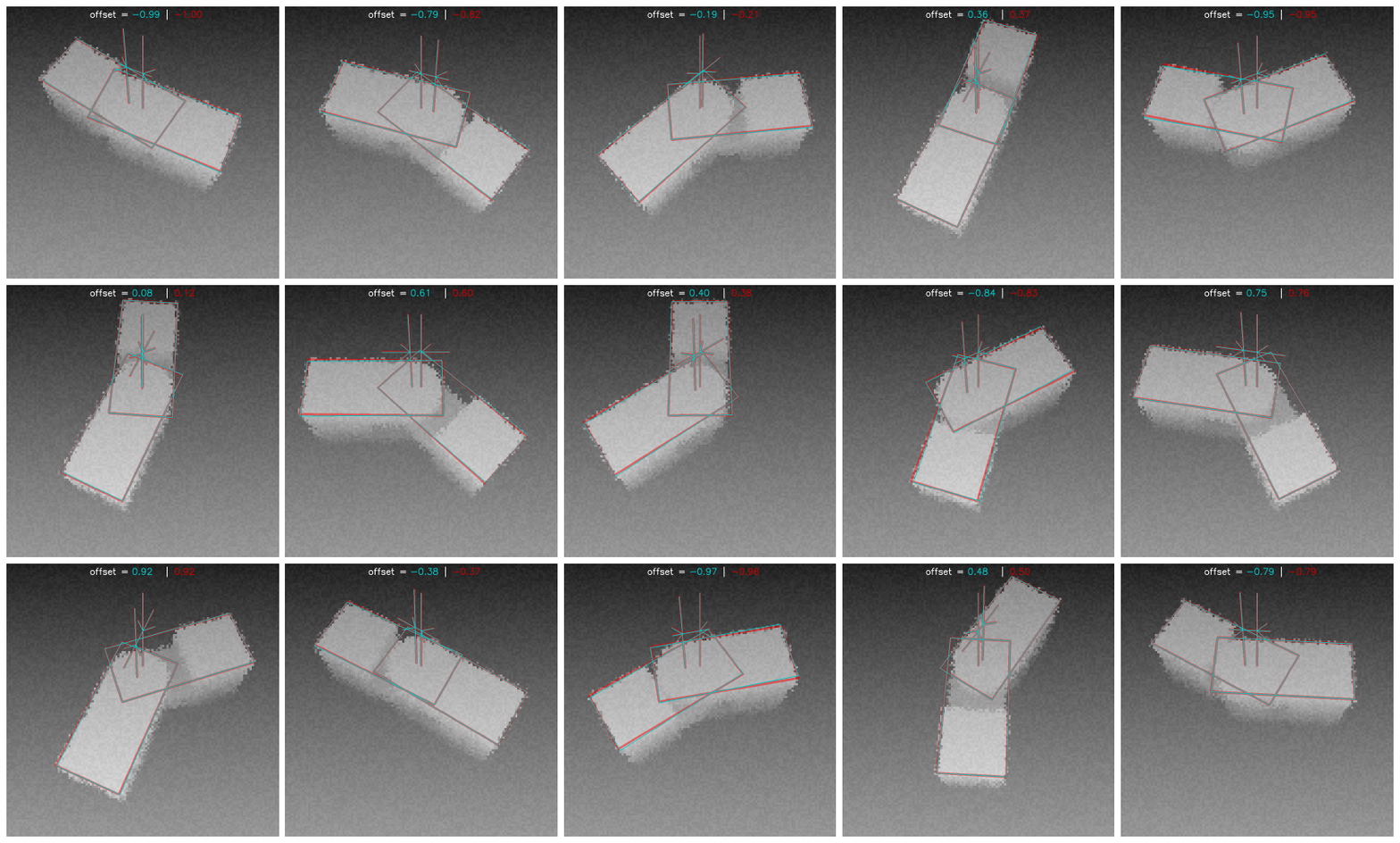}
\caption{Fifteen randomly generated test configurations of the hinge task, inferred from noisy depth maps with the trained twincher (not selected). True (red) and inferred (cyan) bounding rectangles of both parts, connection axes, and offset values are shown.}
\label{fig:hinge-tests}
\end{figure}

We next consider inferring the configuration of a 3D object from a noisy depth map. This is a prototype of render-and-compare pose estimation~\cite{li.deepim.eccv.2018} and of virtual metrology for quality inspection, e.g., of objects on a conveyor belt observed by a depth sensor, and it is relevant to robotic manipulation. For rigid objects, classical model-based registration such as the iterative closest point algorithm~\cite{besl.tpami.1992} is a strong alternative; here, the task serves as a controlled testbed for a generic learned approach that requires only query access to the renderer. The forward process is a renderer that produces depth maps of a hinge composed of two parts, controlled by $n_p = 3$ parameters: the relative angle between the parts, restricted to $[-\pi/3, \pi/3]$ to avoid self-intersection; the orientation of the first part, restricted to $[-\pi/3, 4\pi/3]$; and the offset between the connection axes of the two parts (see Fig. \ref{fig:hinge-tests}). The orientation range deliberately excludes the full circle, which would make the forward map non-injective because of the symmetry of the object; the full range could be covered by several twinchers trained on overlapping intervals, i.e., by an atlas of charts. Each parameter is mapped affinely to $[-1, 1]$. The hinge is assumed to be centered in the field of view, and the renderer outputs a $128 \times 128$ depth image with values scaled to $[-1, 1]$.

To emulate sensing artifacts, we add independent uniform noise $U(-0.5, 0.5)$ to each pixel and, independently for each image axis, displace each pixel by $\pm 1$ with probability $0.2$, so that its value is taken from a neighboring pixel. The image is then average-pooled over non-overlapping $16 \times 16$ patches, which yields an $8 \times 8$ map, i.e., $n_y = 64$.

\begin{figure}[htbp]
\centering
\includegraphics[width=0.7\columnwidth]{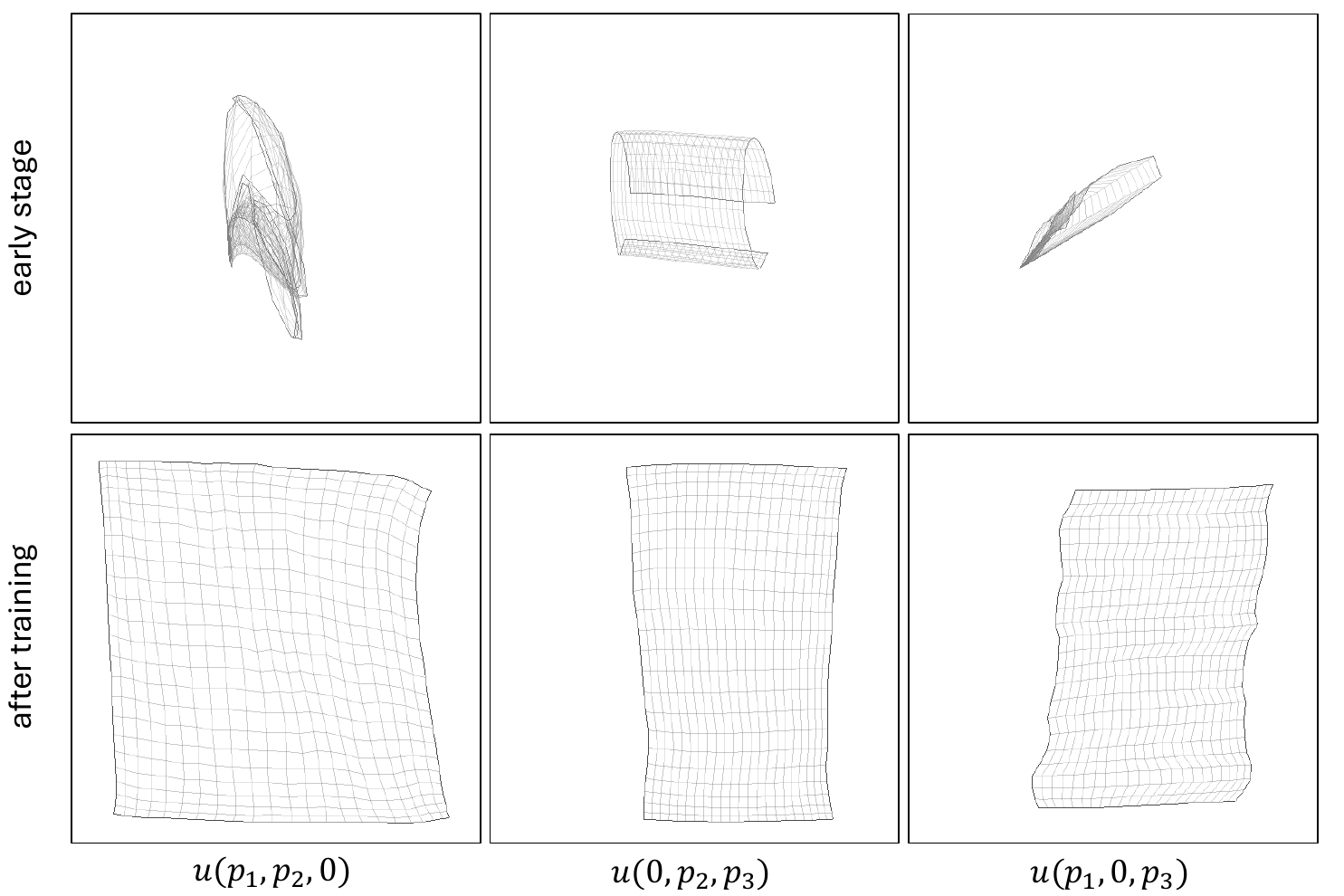}
\caption{Two-dimensional slices of the learned representation $r(f(p))$ for the hinge task, obtained by fixing one component of $p$ to zero and varying the other two over $[-1, 1]$ (label $u(\cdot)$ indicates component fixed to zero), at an early stage of training (top) and after training (bottom). The slices are projected onto the tangent plane at $p = 0$ spanned by $J e_i$ and $J e_k$, where $e_i, e_k$ are canonical basis vectors in $\mathcal{P}$ and $J = \partial r/\partial p\,|_{p=0}$.}
\label{fig:hinge-learning}
\end{figure}

We train a twincher with approximately $2.5 \times 10^4$ parameters ($n_s=64$, $n_l=196$) in the static regime, using approximately $6.4 \times 10^4$ forward evaluations without noise. Figure~\ref{fig:hinge-learning} visualizes the representation at an early stage and after training by projecting two-dimensional slices of $r(f(p))$, obtained by varying two parameters with the third fixed at zero, onto the tangent plane at $p = 0$. Training yields smooth and regular slices without folds, consistent with a bijective representation over the evaluated range.

Figure~\ref{fig:hinge-tests} shows 15 random test configurations with true (red) and inferred (cyan) bounding rectangles, connection axes, and offset values, inferred from noisy depth maps. The model recovers the configuration in all cases, with small deviations attributable to the observation noise. Given the small size of the model, this experiment should be regarded as a proof of concept; Sec.~\ref{sec:itodd} applies the approach to real depth images.

\begin{figure}[htbp]
\centering
\includegraphics[width=0.45\columnwidth]{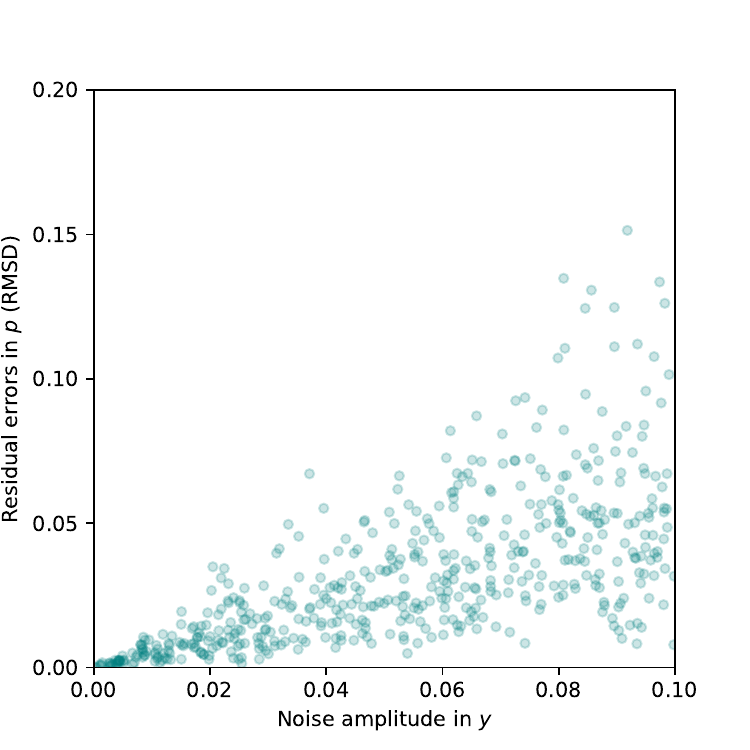}
\caption{RMS deviation of the inferred parameters from the true ones as a function of the RMS perturbation in observation space $\mathcal{Y}$, for the twincher trained on the hinge task.}
\label{fig:hinge-err}
\end{figure}

To quantify stability, we generated random configurations and applied observation perturbations of varying magnitude. Figure~\ref{fig:hinge-err} shows the RMS deviation $\Delta p$ of the inferred parameters as a function of the RMS perturbation $\Delta y$. The deviation vanishes together with the perturbation and remains below $\eta\,\Delta y$ with $\eta \approx 2$ over the evaluated range. Two features are noteworthy. First, there is no visible error floor as $\Delta y \to 0$, whereas a direct regressor would be expected to retain its approximation error even for noise-free observations; this is consistent with the errors being caused by the observation perturbations rather than by approximation errors of the learned model. Second, by Proposition~\ref{prop:sensitivity}, $\eta$ is bounded from below by the intrinsic conditioning of the forward map, and reducing it toward this limit corresponds to reducing $L_{\mathrm{noise}}$. We expect larger models and longer training to tighten the bound but have not quantified this here.

\FloatBarrier
\section{Application: 6D localization of industrial objects in real depth images}
\label{sec:itodd}

The experiments of Sec.~\ref{sec:experiments} isolate the mechanism on synthetic problems. We now test it on a well-studied perception task with real sensor data: estimating the 6D pose---rotation and translation---of rigid industrial objects from depth images, a core capability for robotic manipulation (Sec.~\ref{sec:related}). We do not propose a competing pose estimator. Instead, we use the task as a demanding test of the principle of Sec.~\ref{sec:formulation}: twinchers trained only on renderings of a CAD model should provide coordinates in which the iteration of Eq.~\eqref{eq:newton}, driven by the exact renderer, converges from a fixed initial guess on real observations. The pipeline therefore deliberately contains no feature extraction or matching, no learned regression of the pose or of its initial estimate, no pose hypotheses, voting, or RANSAC, and no ICP. The twinchers are its only learned components, and the pose is determined by comparing renderings with the observation. A research log in the repository documents all steps, including approaches that did not work~\cite{twincher-itodd}.

\begin{figure}[t]
\centering
\includegraphics[width=\linewidth]{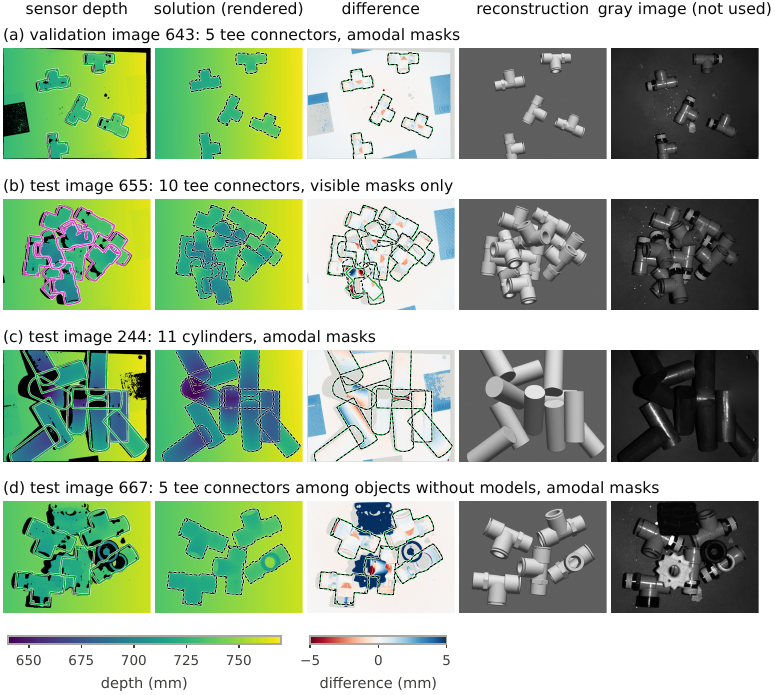}
\caption{Whole ITODD scenes solved instance by instance, cropped to the objects. From left to right: sensor depth with the masks used (green: amodal; magenta: visible part); depth rendered at the solutions in front of a fitted table plane; difference between rendered and sensor depth (gray: no sensor depth); shaded rendering of the solutions; and the gray image of the scene, which the method does not use. Dashed black outlines show the full silhouettes of the solutions, including parts hidden by other instances. (a)~Validation image; (b)~test image with a pile of connectors, solved without the amodal masks; (c)~test image with stacked cylinders, in which all eleven solutions match their masks (IoU $\ge 0.986$); (d)~test image in which the connectors lie among objects of other types, for which the method has neither models nor masks and which therefore appear in the difference; all five solutions agree with the observation.}
\label{fig:itodd-scenes}
\end{figure}

\subsection{Setting}
\label{sec:itodd-setting}

\paragraph{Data and task.}
We use the MVTec ITODD dataset~\cite{drost.iccvw.2017} in the format of the BOP benchmark. It shows 28 texture-less industrial objects in realistic arrangements, including piles as in bin picking, recorded by a high-quality multi-shot structured-light sensor with $1280 \times 960$ depth maps. No real training images are provided, so that methods must be trained from the CAD models, and in the BOP Challenge 2023, ITODD had the lowest scores of the seven core datasets for the leading methods~\cite{hodan.cvprw.2024}. On flat surfaces, the depth noise is 70--90\,$\mu$m RMS, whereas 10--24\% of the pixels of an image carry no depth because of stereo shadows and specular reflections. We use depth only and consider two objects (Fig.~\ref{fig:itodd-scenes}): a cylinder (object~12, diameter 107.7\,mm), which is symmetric under rotations about its axis and under a flip of the axis, and a tee connector (object~26, diameter 66.8\,mm), which has no exact symmetry but is nearly symmetric under a half-turn. We process all annotated instances of both objects: 4 cylinders and 6 connectors in the validation images, which have ground truth, and 70 cylinders and 159 connectors in the test images, whose ground truth is withheld by BOP. As in the BOP task of 6D localization of seen objects, the identities of the objects are given. In addition, each instance is located by a mask provided by BOP: either the \emph{amodal} mask, i.e., the full silhouette including occluded parts, or only the mask of its visible part. Because these masks are derived from the ground truth, the two versions emulate pipelines with an ideal amodal segmentation and with an ideal segmentation of visible parts, respectively. This separates the inverse problem addressed here---finding the pose of an object whose location in the image is known---from detection and segmentation, which we do not address.

\paragraph{Forward model.}
After centering on the object, the pose has $n_p = 4$ parameters: the distance of a virtual camera that looks at the model origin, the elevation and azimuth of its viewing direction, and its roll about the optical axis. For the cylinder, the symmetry removes the azimuth, so that $n_p = 3$. The two remaining degrees of freedom, the direction of the model origin in the camera frame, are determined during inference (see below). Each parameter is mapped affinely onto $[-1,1]$. No single box of parameters covers all orientations without singularities, as for any three-parameter representation of rotations~\cite{zhou.cvpr.2019}, and views related by symmetries have to be kept apart, because the method requires an injective forward map (Sec.~\ref{sec:formulation}). We therefore use an atlas of overlapping charts, each with its own twincher: two charts with different ranges of the roll for the cylinder, and six for the connector, which combine two families of elevation axes with three roll ranges centered $120^\circ$ apart (Appendix~\ref{app:itodd}). For the connector, every pose lies in some chart with $\|p\|_\infty \le 0.857$, and the twinchers are trained on $\|p\|_\infty \le 0.86$; the charts of the cylinder cover all views except those within $17^\circ$ of its axis. The forward map $f$ renders a $512 \times 512$ depth map of the CAD model by ray casting on the GPU (about 0.1\,ms per view in batches) and condenses it into $y$ by a fixed, non-learned encoder centered at the center of mass of the silhouette, which makes $y$ invariant to shifts in the image: for the cylinder, local averages of the depth with tent-shaped weights centered at the nodes of a $16 \times 16$ grid ($n_y = 256$); for the connector, 63 low-order Fourier--Bessel components~\cite{xiao.pr.2010} of the height above a plane 20\,mm behind the mean depth of the object, together with one component that encodes the distance ($n_y = 64$). All components are squashed into $(-1, 1)$, where the twinches act. Jacobians $\partial f/\partial p$ are computed by central finite differences, and during training, $y$ and $\partial f/\partial p$ are interpolated multilinearly from a grid in $p$. For the connector, the roll enters exactly through the phases of the Fourier--Bessel coefficients, which reduces the grid to the other three parameters ($8 \times 32 \times 96$ nodes, rendered in two minutes per chart).

\paragraph{Training.}
For the cylinder, each chart has a twincher with $n_s = n_l = 256$ ($1.3 \times 10^5$ parameters), trained with the hyper-sail procedure of Sec.~\ref{sec:training} in less than an hour on a laptop GPU (NVIDIA GeForce RTX 5050). For the connector, each chart has a twincher with $n_s = 128$ and $n_l = 384$ ($9.8 \times 10^4$ parameters). Its more complex geometry required four modifications, described in Appendix~\ref{app:itodd}: a finer curriculum, in which $R$ starts at 0.05 and grows in steps of 0.0025; evaluation of the trimming term also on samples drawn from the whole domain $[-1,1]^{n_p}$, which keeps $r$ well behaved beyond the current hypercube; insertion of identity layers to add capacity to a trained twincher; and protection against rare loss spikes by checkpoint rollback and, in later runs, by a reset of the optimizer's momentum. Training took 1--2.7\,h per chart. Afterwards, every connector twincher was an embedding of its chart on all evaluated samples: $\det J > 0$ at 8192 random points, and the iteration of Eq.~\eqref{eq:newton} from $p^{(0)} = 0$ solved all of 512--1024 random synthetic views to $\|\Delta p\|_\infty < 10^{-2}$, and 98--99\% of them to $10^{-3}$. All twinchers were then trained for robustness with a single-mode variant of $L_{\mathrm{noise}}$, included in the package~\cite{twincher-code}. For each sample, it penalizes the largest directional derivative of $r$ normal to $\mathcal{M}$, i.e., along the leading right singular vector of $(\partial r/\partial y)P_N$, instead of derivatives along random normal directions, also at points displaced from $\mathcal{M}$ by up to 0.05 in norm. This post-training (2--3\,h per connector chart) reduced the sensitivity to normal perturbations by a factor of 1.6--3.1 for the connector and 5.4--5.6 for the cylinder (Table~\ref{tab:itodd-charts}), and it also reduced the median error on noise-free synthetic views by a factor of 1.5--3.

\paragraph{Inference with incomplete observations.}
For each instance, the depth map is resampled, with perspective correction, into the view of a virtual camera rotated toward the object. Before each step of the iteration, the direction of this camera is adjusted so that the centers of mass of the observed and the rendered silhouettes coincide, which determines the two remaining degrees of freedom. In each chart, the iteration of Eq.~\eqref{eq:newton} starts at the center of the chart, $p^{(0)} = 0$, and runs for 20 steps with $p$ clamped to the trained box; of the solutions of all charts, the one with the smallest misfit between rendering and observation is kept. Real observations are incomplete: a median of 28\% (cylinder) and 30\% (connector) of the amodal silhouette of an object is unavailable, mostly because of sensor holes and, in piles, because of occlusions, and parts of an object may lie outside the image. We handle unknown pixels with a \emph{fused target}. At each step, they are filled with the rendering at the current estimate, which yields an observation $y_{\mathrm{fused}}(p)$, and Newton's method is applied to $F(p) = r(y_{\mathrm{fused}}(p)) - r(f(p)) = 0$, i.e., $\Delta p = (J - J_{\mathrm{fused}})^{-1} F(p)$, where $J_{\mathrm{fused}} = \partial r(y_{\mathrm{fused}}(p))/\partial p$ is obtained from the same renderings as $J$. Because $J - J_{\mathrm{fused}}$ can be nearly singular far from the solution, the iteration is also run with the simpler step $\Delta p = J^{-1} F(p)$, and the better result is kept. At a fixed point, the rendering agrees with the observation on all available pixels, and without unknown pixels the scheme reduces to Eq.~\eqref{eq:newton}; it resembles the EM algorithm, which imputes missing data from the current model~\cite{dempster.jrssb.1977}. Pixels of the visible mask without depth count as object pixels, so that a solution cannot slide off the visible silhouette. Without the amodal mask, pixels outside the visible mask at which the rendering lies more than 5\,mm in front of the observed surface count as free space, in the spirit of space carving in volumetric reconstruction~\cite{curless.siggraph.1996}. With two or six charts and two step variants, an instance takes 3.4\,s (cylinder) or 8.8\,s (connector) on the laptop GPU (medians), dominated by rendering in an unoptimized implementation.

\paragraph{Evaluation.}
On the validation images, we compute the pose errors of BOP~\cite{hodan.eccvw.2016, hodan.eccv.2018}---the visible surface discrepancy $e_{\mathrm{VSD}}$, the maximum symmetry-aware surface distance (MSSD), and the maximum symmetry-aware projection distance (MSPD)---and the average recall (AR) over the thresholds of BOP: from 0.05 to 0.5 for $e_{\mathrm{VSD}}$, from 5\% to 50\% of the object diameter for MSSD, and from 10 to 100 pixels for MSPD at this image width; our implementation follows the BOP toolkit. Because the ground truth of the test images is withheld, we evaluate test instances by their agreement with the observation: a solution \emph{agrees} if its silhouette overlaps the amodal mask with an intersection over union (IoU) of at least 0.95, and its rendered depth deviates from the sensor depth by at most 2\,mm on average over the pixels where both show the object. Agreement is necessary for a correct pose but not sufficient in principle, e.g., for nearly symmetric views; on the validation images, agreement and correctness coincided in all cases. We relate the results to the \emph{unavailable fraction} of an instance, the fraction of its amodal mask that is occluded or has no depth, and report the 21 instances cut by the image border separately, because the masks do not cover the parts outside the image.

\subsection{Results}
\label{sec:itodd-results}

\begin{figure}[t]
\centering
\includegraphics[width=\linewidth]{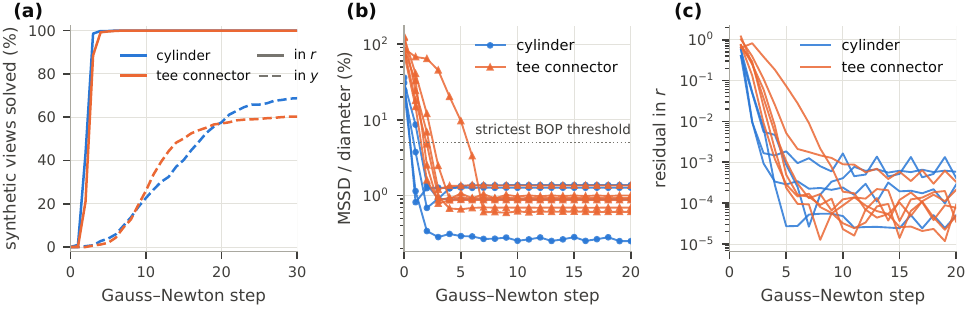}
\caption{Convergence of the iteration from the center of a chart, $p^{(0)} = 0$. (a)~Fraction of random synthetic views (256 per chart, averaged over the charts of each object) solved to $\|\Delta p\|_\infty < 10^{-2}$ by Gauss--Newton iteration in the learned representation $r$ and in observation space $y$, with the same forward model, initial guess, and clamp; continuing the iteration in $y$ to 60 steps raises these fractions by at most 0.013. (b)~MSSD to the ground truth, relative to the object diameter, for the 10 validation instances with real depth data; the dotted line marks the strictest threshold of BOP. (c)~Norm of the residual $F(p)$ of the fused-target iteration in representation space for the same instances.}
\label{fig:itodd-convergence}
\end{figure}

\paragraph{The learned coordinates make the iteration converge.}
To isolate the role of the twinchers, we solve random synthetic views of every chart (256 per chart, rendered without noise) with the same forward model, initial guess $p^{(0)} = 0$, and clamp in two ways: in representation coordinates for 30 steps, as above, and in observation space for 60 steps by the damped Gauss--Newton iteration of the baseline of Sec.~\ref{sec:entangler} ($\lambda = 10^{-3}$, $\Delta_{\max} = 0.1$). In representation coordinates, all views of all charts are solved to $\|\Delta p\|_\infty < 10^{-2}$ (99.5\% to $10^{-3}$), more than 99\% of them within five steps. In observation space, the iteration solves 70\% of the cylinder views and 61\% of the connector views, and only 50\% in the charts that contain views along the bores (Fig.~\ref{fig:itodd-convergence}a). Without step clipping, it solves 99\% of the cylinder views but no more of the connector views (60\%). From random initial points in the chart, the observation-space iteration succeeds for 68\% of the cylinder views and for 29--38\% of the connector views, which corresponds to complexities $C \approx 0.4$ and $C \approx 1.0$--$1.25$ in the sense of Sec.~\ref{sec:entangler}, comparable to the harder harmonic entanglers of Fig.~\ref{fig:entangler-phase}; from the same initial points, the iteration in representation coordinates succeeds for 99.6--100\% of the views. The forward model and its Jacobians are identical in both cases; only the coordinates differ, and these are what the twincher contributes.

\paragraph{Accuracy on the validation images.}
Table~\ref{tab:itodd-val} summarizes the validation instances. All 10 poses are correct at every threshold of BOP (AR~=~1), with and without the amodal mask. The MSSD is at most 1.4\% of the object diameter (1.48\,mm for the cylinder, 0.91\,mm for the connector), a factor of 3.6 below the strictest threshold of 5\%; translation errors are 0.1--0.7\,mm, and rotation errors are 0.1--1.6$^\circ$. Moreover, the depth rendered at the solutions deviates from the sensor depth by about as much as, or less than, the depth rendered at the annotated ground-truth poses. The remaining differences between the solutions and the ground truth are thus at the level at which the annotations themselves agree with the sensor data, so that the true accuracy of the solutions cannot be resolved further with these data.

\begin{table}[t]
\centering
\caption{Validation instances of ITODD: ranges over the instances of the pose errors of BOP, the average recall (AR), and the mean absolute difference $|\Delta Z|$ between rendered and sensor depth, at the solutions and at the ground-truth poses. The strictest thresholds of BOP are 5\% of the diameter for MSSD (5.4\,mm for the cylinder, 3.3\,mm for the connector), 10 pixels for MSPD, and 0.05 for $e_{\mathrm{VSD}}$.}
\label{tab:itodd-val}
\small
\begin{tabular}{llccccccc}
\toprule
 & & & \multicolumn{2}{c}{MSSD} & & & \multicolumn{2}{c}{mean $|\Delta Z|$ (mm)} \\
\cmidrule(lr){4-5} \cmidrule(lr){8-9}
Object (instances) & Mask & AR & (mm) & (\% of diameter) & MSPD (px) & $e_{\mathrm{VSD}}$ & solution & ground truth \\
\midrule
Cylinder (4) & amodal & 1 & 0.27--1.48 & 0.25--1.38 & 0.65--2.30 & 0.004--0.010 & 0.16--0.36 & 0.40--0.81 \\
 & visible & 1 & 0.40--1.20 & 0.38--1.12 & 0.68--2.34 & 0.005--0.010 & 0.24--0.52 & 0.40--0.81 \\
Connector (6) & amodal & 1 & 0.41--0.91 & 0.61--1.36 & 1.10--2.16 & 0.010--0.028 & 0.50--0.71 & 0.48--0.63 \\
 & visible & 1 & 0.47--0.90 & 0.70--1.34 & 1.23--2.19 & 0.014--0.028 & 0.51--0.72 & 0.48--0.63 \\
\bottomrule
\end{tabular}
\end{table}

\paragraph{Convergence on real data.}
Figure~\ref{fig:itodd-convergence}b,c shows the iteration on the validation instances. From the center of the chart, where the MSSD is 26--39\,mm for the cylinder and 49--84\,mm for the connector (24--126\% of the diameter), the error reaches its final level within 3--5 steps, and within 7 steps for an instance that is cut by the image border and has 62\% of its silhouette unavailable. The residual in representation space decreases by three to four orders of magnitude and then levels off at $10^{-5}$--$10^{-3}$, a floor set by finite-difference Jacobians, the pixel quantization of the renderer, and the mismatch between the CAD models and the real objects. Without ground truth, convergence can be measured against the final solution: for the 69 cylinder instances of the validation and test images whose solutions agree with the observation, the pose comes within 1\,mm (MSSD) of its final value after a median of 3 steps, and within 6 steps for 90\% of them.

\begin{figure}[t]
\centering
\includegraphics[width=\linewidth]{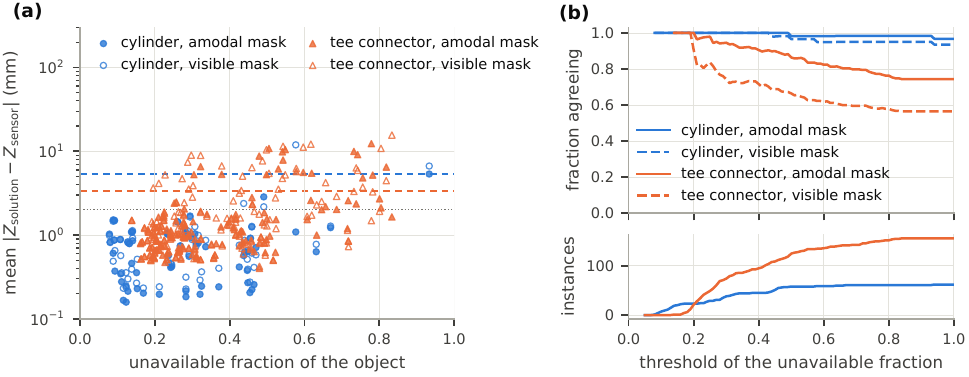}
\caption{Agreement with the observation on all instances of the validation and test images that are not cut by the image border (62 cylinders, 156 connectors). (a)~Mean absolute difference between rendered and sensor depth against the unavailable fraction, one marker per instance and version; dashed lines: 5\% of the object diameter, the strictest MSSD threshold of BOP, as a scale; dotted line: the 2\,mm criterion of agreement. (b)~Fraction of the solutions that agree with the observation (silhouette IoU $\ge 0.95$, mean depth difference $\le 2$\,mm) among the instances whose unavailable fraction lies below a threshold, and their number.}
\label{fig:itodd-agreement}
\end{figure}

\paragraph{Robustness to missing data.}
Figure~\ref{fig:itodd-agreement} and Table~\ref{tab:itodd-agreement} relate the agreement with the observation to the unavailable fraction for all instances. With the amodal mask, all 38 cylinders and 70 of 74 connectors with up to 30\% unavailable agree with the observation; overall, 97\% of the cylinders and 74\% of the connectors inside the image agree, as do 9 of 12 cylinders and 7 of 9 connectors cut by the image border. Over all 239 instances of both objects, 80\% of the solutions agree with the observation with amodal masks and 67\% with visible masks only. Agreement declines beyond about 45\% unavailable and almost vanishes beyond 75\%, which delimits the present operating range. The agreeing solutions fit the sensor data closely: the mean depth difference has a median of 0.63\,mm (cylinder) and 0.85\,mm (connector). A plane fitted to the depth difference on the interior of each object has a median absolute offset of 0.09 and 0.18\,mm, and the robust scatter around it is 0.19 and 0.49\,mm, compared with a sensor noise of 0.08\,mm on flat surfaces. This remaining scatter reflects deviations of the real objects from their CAD models and systematic sensor errors, which no pose can remove. Without the amodal mask, the cylinder is hardly affected (94\% agree), whereas the connector drops to 56\%: the amodal mask also constrains the parts of the silhouette hidden behind other objects and marks background where no depth is measured behind the object, which free-space carving cannot do, and the silhouette of the connector is far more informative than that of the cylinder.

\begin{table}[t]
\centering
\caption{Solutions that agree with the observation, by unavailable fraction of the instance (instances inside the image), and for instances cut by the image border.}
\label{tab:itodd-agreement}
\small
\begin{tabular}{llccccccc}
\toprule
 & & \multicolumn{5}{c}{Unavailable fraction} & & Cut by \\
\cmidrule(lr){3-7}
Object & Mask & 0--30\% & 30--45\% & 45--60\% & 60--75\% & 75--100\% & All inside & the border \\
\midrule
Cylinder & amodal & 38/38 & 13/13 & 7/8 & 2/2 & 0/1 & 60/62 (97\%) & 9/12 \\
 & visible & 38/38 & 12/13 & 6/8 & 2/2 & 0/1 & 58/62 (94\%) & 8/12 \\
Connector & amodal & 70/74 & 28/35 & 14/27 & 4/13 & 0/7 & 116/156 (74\%) & 7/9 \\
 & visible & 55/74 & 20/35 & 9/27 & 3/13 & 1/7 & 88/156 (56\%) & 7/9 \\
\bottomrule
\end{tabular}
\end{table}

\paragraph{Insensitivity to normal perturbations matters on real data.}
The robustness post-training, which targets condition (ii), translated directly into performance on real data. For the connector with the amodal mask, the number of test instances whose silhouette matched the mask with IoU $\ge 0.95$ rose from 105 to 114 (of the 148 instances evaluated at that stage), and in synthetic tests, the fraction of correct poses rose from 90\% to 97\% for objects with 20--40\% occluded and from 57\% to 77\% for objects of which 10--40\% lay outside the image. Imperfectly filled unknown pixels and the differences between real and rendered depth maps both displace $y$ from $\mathcal{M}$, and by Proposition~\ref{prop:sensitivity}, condition (ii) suppresses the effect of such displacements on the solution.

\begin{figure}[t]
\centering
\includegraphics[width=\linewidth]{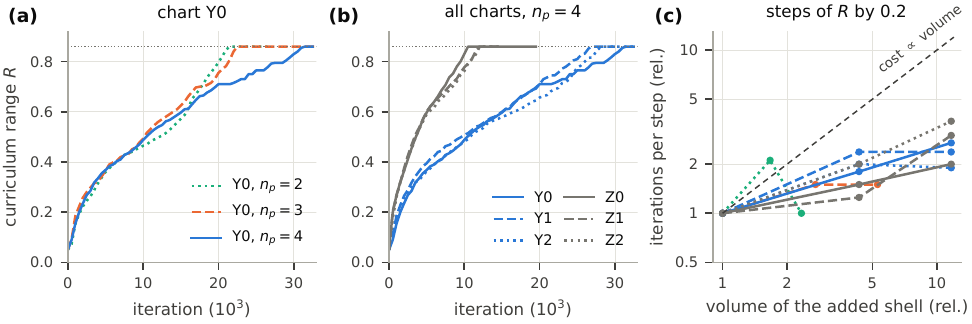}
\caption{Training effort and the number of parameters $n_p$. (a)~Growth of the curriculum range $R$ for chart Y0 of the connector with $n_p = 2$ (elevation and azimuth; 64 layers), 3 (and roll; 256 layers), and 4 (and distance; 256 layers, extended to 384 at $2.5 \times 10^4$ iterations), with the same encoder and training procedure; dotted line: the target $R = 0.86$. (b)~The six charts of the connector ($n_p = 4$, 384 layers). (c)~Number of iterations needed to grow $R$ from 0.2 to 0.4, 0.4 to 0.6, and 0.6 to 0.8, against the volume of the shell that each step adds to the training domain, both relative to the first step, for the runs of (a) and (b) (same line styles); dashed: a cost proportional to the added volume.}
\label{fig:itodd-scaling}
\end{figure}

\paragraph{Training iterations grow slowly with the number of parameters.}
A natural concern about any method that learns a map over a parameter domain is that the required capacity and training time might grow exponentially with $n_p$, as they would if the learned model had to represent the domain cell by cell. The connector provides initial evidence on this point (Fig.~\ref{fig:itodd-scaling}). With the same encoder and training procedure, chart Y0 reached $R = 0.86$ after $2.15$, $2.25$, and $3.15 \times 10^4$ iterations for $n_p = 2$, 3, and 4, whereas the number of grid nodes needed to resolve $f$ on the chart grew 192-fold, from 3072 to $5.9 \times 10^5$ (Appendix~\ref{app:itodd}); the number of parameters used grew from $1.6 \times 10^4$ to $9.8 \times 10^4$ (64 to 384 layers), which were sufficient rather than minimal capacities, and the total computation, whose cost per iteration grows with $n_l n_p$, grew 13-fold. Within a run, growing $R$ by a fixed step took 1.9--3.7 times as many iterations for the outermost step considered as for the innermost one, although the outermost step added 11.7 times the volume to the training domain in four dimensions (Fig.~\ref{fig:itodd-scaling}c). At equal $n_p = 4$, the Y charts, which contain views along the bores, needed about 2.5 times as many iterations as the Z charts, and the two cylinder charts ($n_p = 3$) covered their whole domain after $3 \times 10^3$ iterations. In these examples, the number of training iterations was thus governed by the geometric complexity of the forward map rather than by $n_p$, and new regions of the domain were learned at a cost far below proportionality to their volume. We interpret this as initial evidence that twinchers generalize across the domain rather than memorize it: when the curriculum extends the domain, the transformation learned so far extends continuously into the new region and needs to be adapted rather than built anew. In our view, the primitive supports this through its variable scale: the learned widths of the twinches with a noticeable twist span an order of magnitude (5--95\% quantiles 0.065--0.5 for the connector charts, median 0.24, compared with 0.1 at initialization), so that the transformation combines broad rotations acting on large parts of the state space with localized corrections.

\subsection{Discussion}
\label{sec:itodd-discussion}

\paragraph{What the example shows.}
The example combines four properties that are rarely found together. \emph{Global convergence without a learned initial estimate}: one deterministic iteration per chart, started at the center of the chart, followed by the choice of the solution with the smallest misfit among a few charts, replaces the detection of features and the generation and verification of many pose hypotheses. \emph{Precision}: the accuracy is limited by the sensor and the CAD models, not by the learned model; on noise-free synthetic depth maps of the connector, the median MSSD is 0.03\,mm. \emph{Learning from simulation only}: no real images, domain randomization~\cite{tobin.iros.2017}, or photorealistic synthetic data are used. The gap between simulation and reality is bridged not by making a learned model invariant to the appearance of real data but by the structure of the inference: the learned model only defines coordinates, the renderer is compared with the observation, and real-world discrepancies act as perturbations to which the coordinates are trained to be insensitive. \emph{Compactness}: about $10^5$ parameters per chart, roughly two orders of magnitude fewer than the convolutional or transformer backbones of learned pose estimators~\cite{wang.cvpr.2021, labbe.corl.2022, wen.cvpr.2024}, acting on 64--256 numbers condensed from a $512 \times 512$ rendering and trained in hours on a laptop GPU. That these properties result from a generic primitive, trained with the minimal procedure of Sec.~\ref{sec:training} and a few generic safeguards, in a domain whose established pipelines combine several task-specific components, is in our view the main result of this section.

\paragraph{Current limitations and outlook.}
As a first application, the example has limitations, many of which reflect its deliberately minimal configuration rather than the principle. First, instance masks are taken from the annotations. In a deployed system, they would come from an instance segmentation, e.g., based on foundation models~\cite{kirillov.iccv.2023, nguyen.iccvw.2023}, possibly with amodal completion~\cite{li.eccv.2016, zhu.cvpr.2017}; the results with visible masks only indicate what is at stake for the connector. Second, agreement decreases when more than about 45\% of an object is unavailable. The Fourier--Bessel components are global, so that every gap affects all of them; encoders built from localized kernels, as used for the cylinder, and hybrid pipelines in which a coarse estimator places the initial guess inside the basin of convergence of a twincher are natural remedies. Third, onboarding an object still involves choices made by hand---charts, encoder, training safeguards---and two of the 28 objects of ITODD have been processed so far. The difficulties we encountered were, however, generic and were resolved by generic means (Appendix~\ref{app:itodd}), which makes automation plausible. Fourth, an instance takes several seconds in the present implementation, which renders every chart and runs a fixed number of steps; since the iteration converges within about five steps and the relevant chart can often be anticipated, a substantial reduction appears straightforward. Finally, the present results are not directly comparable with the BOP leaderboard, on which the best average recall for ITODD is 73.4\% with RGB-D data and 57.0\% with depth only~\cite{bop.leaderboard.itodd}. These values cover all 28 objects and the test targets of the BOP protocol, and they include the detection of the objects, whereas our results concern two objects with masks taken from the annotations and rely on agreement with the observation as a proxy for correctness on the test images. A direct comparison requires replacing the annotation masks by a detection and segmentation stage, e.g., built on the default detections provided by BOP~\cite{hodan.cvprw.2024}, and a submission to the evaluation server, which we leave for future work. We regard these as steps of engineering and scaling rather than of principle. The example uses a single type of primitive and the minimal training procedure of this paper; the integration of the same primitive into more advanced architectures for robotics is the subject of ongoing work.

\FloatBarrier
\section{Discussion and limitations}
\label{sec:discussion}

\paragraph{Why accuracy need not scale with data.}
The results indicate that, once a bijective representation has been formed, iterative inference converges to the solution for all tested targets in the absence of noise and yields errors bounded linearly by the observation perturbation in its presence. This may appear atypical for trained models, whose errors are commonly expected to decrease gradually with the amount of training data~\cite{hestness.arxiv.2017, kaplan.arxiv.2020}. The behavior is structural. The learned model is used only to define coordinates in which the inverse problem is well conditioned, while accuracy is supplied by the exact forward model during refinement. What must be learned is therefore a qualitative property---injectivity and bounded distortion of $g$ over $\mathcal{P}$. For smooth forward maps, this property can be attained from a finite sample once $\mathcal{P}$ is covered densely relative to the scale on which $f$ varies, a condition that can often be checked or estimated for practical forward processes; in the dynamic regime, training itself would indicate the regions of $\mathcal{P}$ that require denser sampling. In this sense, the twincher parameters encode a compact summary of the forward process that is sufficient for stable inversion, and robustness to noise or model mismatch reduces to the invariance of $r$ with respect to the corresponding perturbations of $y$. The application to real depth images (Sec.~\ref{sec:itodd}) shows the same structure outside synthetic settings: the precision of the solutions is set by the sensor and by the fidelity of the CAD models, and the discrepancies between real and rendered data act as the perturbations that condition (ii) targets. We emphasize that these statements are empirical: bijectivity is verified on finite samples and test sets, not certified.

\paragraph{Scalability with the number of parameters.}
Whether the approach scales to many parameters depends on how the required data, capacity, and training time grow with $n_p$. Three observations suggest that this growth need not be exponential. First, the cost of a training iteration grows only linearly with $n_p$ (Sec.~\ref{sec:derivatives-summary}), and in Sec.~\ref{sec:itodd}, the number of iterations was governed by the geometric complexity of the forward map rather than by $n_p$, with new regions of the domain learned at a cost far below proportionality to their volume (Fig.~\ref{fig:itodd-scaling}). Second, the data have to resolve the variation of $f$, which is typically anisotropic: in Sec.~\ref{sec:itodd}, 8 grid nodes sufficed for the distance and 96 were needed for the azimuth, and an exact symmetry made the roll free; in the dynamic regime of Sec.~\ref{sec:variants}, sampling would be concentrated where $\det J$ approaches its margin. Third, the formulation does not require $\mathcal{P}$ to be a box: conditions (i) and (ii) are needed only on the part of the parameter space in which inference or control is to be performed, which for high-dimensional systems is typically a small subset of $[-1,1]^{n_p}$, and the domain-growing curriculum applies equally to domains grown from such a subset. Our evidence is limited to $n_p \le 4$, and the capacity used grew with $n_p$; establishing the scaling in higher dimensions is the main open question, for which we regard the present results as an encouraging starting point.

\paragraph{Limitations.}
First, our experiments are limited to $n_p \le 4$, and the behavior of the determinant-based objective and of the curriculum in higher dimensions remains to be investigated. Second, the method assumes an injective forward map; problems with multiple solutions, such as symmetric objects or redundant kinematic chains, require extensions, such as the atlases of charts used in Sec.~\ref{sec:itodd} or a combination with multimodal models~\cite{ardizzone.iclr.2019}. Third, each refinement step requires $n_p + 1$ forward evaluations (or a differentiable forward model), which an amortized inverse avoids; the approach is therefore most attractive when forward evaluations are affordable relative to the cost of errors, or when accuracy and reliability are paramount. Fourth, training requires Jacobians of the forward model. Fifth, our comparisons use a single parameter-matched baseline on synthetic problems and an ablation of the learned coordinates on real data. Comparisons with globalized or multi-start data-space solvers, conditional invertible networks~\cite{ardizzone.iclr.2019}, simulation-based inference~\cite{cranmer.pnas.2020}, other invertible primitives trained with the same objectives, and, for the application of Sec.~\ref{sec:itodd}, established pose estimators evaluated under identical conditions are needed to delineate the advantages more precisely. Finally, the robustness constant $\eta$ is measured empirically; certified bounds would require verifying conditions (i) and (ii) on all of $\mathcal{P}$.

\section{Conclusions}
\label{sec:conclusions}

We have formulated bijective representation learning for the inversion of continuous forward processes: learning coordinates in observation space in which the forward map becomes a well-conditioned bijection of the parameter domain, and which are insensitive to perturbations normal to the data manifold. In such coordinates, Newton-type iterations on the exact forward model have no spurious interior stationary points, and the inversion is, to first order, as stable as least-squares projection. We introduced Twincher, a volume-preserving architecture built from localized pairwise rotations, together with Jacobian-level objectives, a domain-growing curriculum, and memory-efficient routines for derivative tensors. On synthetic benchmarks with controllable nonconvexity, Twincher exhibits a sharp transition in the query budget beyond which the worst-case residual reaches machine precision within a few iterations, whereas a parameter-matched baseline improves only gradually; a depth-map example illustrates errors that scale linearly with the observation noise. On real depth images of industrial objects, twinchers trained only on renderings of CAD models made the same iteration, driven by the exact renderer, converge from a fixed initial guess to 6D poses whose accuracy is limited by the sensor and the CAD models rather than by the learned model---without feature matching, learned pose regression, or pose hypotheses, and with about $10^5$ parameters per chart. The number of training iterations was governed by the geometric complexity of the forward map rather than by the number of parameters, which we regard as initial evidence that twinchers generalize across the parameter domain rather than memorize it. Although these results are limited to low-dimensional parameter spaces and, on real data, to two objects with given instance masks, they suggest that learning representations whose role is to make an inverse problem well conditioned---rather than to solve it directly---is a useful principle for robust and precise perception and action planning in robotics, vision, and physical AI, and they illustrate how the choice of computational primitive can provide an inductive bias complementary to scale. Future work includes the integration of the primitive into more advanced architectures for robotics, scaling to higher-dimensional parameter spaces, sensitivity-driven active learning, and extensions to non-injective inverse problems.

\bibliographystyle{unsrtnat} 
\bibliography{literature}

@misc{twincher-code,
  author       = {Gonoskov, Arkady},
  title        = {{Twincher}: A {Python} package for bijective representation learning on {CPU} and {GPU}},
  year         = {2026},
  howpublished = {\url{https://github.com/arkady-gonoskov/twincher/}},
  note         = {Version 0.1.0. Also available on PyPI}
}

@inproceedings{vaswani.arxiv.2017,
  author    = {Vaswani, Ashish and Shazeer, Noam and Parmar, Niki and Uszkoreit, Jakob and Jones, Llion and Gomez, Aidan N. and Kaiser, {\L}ukasz and Polosukhin, Illia},
  title     = {Attention is all you need},
  booktitle = {Advances in Neural Information Processing Systems},
  volume    = {30},
  year      = {2017},
  url       = {https://arxiv.org/abs/1706.03762},
  eprint    = {1706.03762},
  archivePrefix= {arXiv}
}

@article{Lecun1998-xc,
  author  = {LeCun, Yann and Bottou, L{\'e}on and Bengio, Yoshua and Haffner, Patrick},
  title   = {Gradient-based learning applied to document recognition},
  journal = {Proceedings of the IEEE},
  volume  = {86},
  number  = {11},
  pages   = {2278--2324},
  year    = {1998},
  doi     = {10.1109/5.726791}
}

@article{kaplan.arxiv.2020,
  author  = {Kaplan, Jared and McCandlish, Sam and Henighan, Tom and Brown, Tom B. and Chess, Benjamin and Child, Rewon and Gray, Scott and Radford, Alec and Wu, Jeffrey and Amodei, Dario},
  title   = {Scaling laws for neural language models},
  journal = {arXiv preprint arXiv:2001.08361},
  year    = {2020},
  doi     = {10.48550/arXiv.2001.08361},
  eprint  = {2001.08361},
  archivePrefix= {arXiv}
}

@article{hestness.arxiv.2017,
  author  = {Hestness, Joel and Narang, Sharan and Ardalani, Newsha and Diamos, Gregory and Jun, Heewoo and Kianinejad, Hassan and Patwary, Md. Mostofa Ali and Yang, Yang and Zhou, Yanqi},
  title   = {Deep learning scaling is predictable, empirically},
  journal = {arXiv preprint arXiv:1712.00409},
  year    = {2017},
  doi     = {10.48550/arXiv.1712.00409},
  eprint  = {1712.00409},
  archivePrefix= {arXiv}
}

@article{sharma.jmlr.2022,
  author  = {Sharma, Utkarsh and Kaplan, Jared},
  title   = {Scaling laws from the data manifold dimension},
  journal = {Journal of Machine Learning Research},
  volume  = {23},
  number  = {9},
  pages   = {1--34},
  year    = {2022},
  url     = {https://arxiv.org/abs/2004.10802},
  eprint  = {2004.10802},
  archivePrefix= {arXiv}
}

@inproceedings{bietti.arxiv.2019,
  author    = {Bietti, Alberto and Mairal, Julien},
  title     = {On the inductive bias of neural tangent kernels},
  booktitle = {Advances in Neural Information Processing Systems},
  volume    = {32},
  year      = {2019},
  url       = {https://arxiv.org/abs/1905.12173},
  eprint    = {1905.12173},
  archivePrefix= {arXiv}
}

@inproceedings{locatello.arxiv.2018,
  author    = {Locatello, Francesco and Bauer, Stefan and Lucic, Mario and R{\"a}tsch, Gunnar and Gelly, Sylvain and Sch{\"o}lkopf, Bernhard and Bachem, Olivier},
  title     = {Challenging common assumptions in the unsupervised learning of disentangled representations},
  booktitle = {Proceedings of the 36th International Conference on Machine Learning (ICML)},
  series    = {PMLR},
  volume    = {97},
  year      = {2019},
  url       = {https://arxiv.org/abs/1811.12359},
  eprint    = {1811.12359},
  archivePrefix= {arXiv}
}

@inproceedings{liu.arxiv.2024,
  author    = {Liu, Ziming and Wang, Yixuan and Vaidya, Sachin and Ruehle, Fabian and Halverson, James and Solja{\v{c}}i{\'c}, Marin and Hou, Thomas Y. and Tegmark, Max},
  title     = {{KAN}: {Kolmogorov--Arnold} networks},
  booktitle = {International Conference on Learning Representations (ICLR)},
  year      = {2025},
  url       = {https://arxiv.org/abs/2404.19756},
  eprint    = {2404.19756},
  archivePrefix= {arXiv}
}

@article{fakhoury.arxiv.2022,
  author  = {Fakhoury, Daniele and Fakhoury, Emanuele and Speleers, Hendrik},
  title   = {{ExSpliNet}: An interpretable and expressive spline-based neural network},
  journal = {arXiv preprint arXiv:2205.01510},
  year    = {2022},
  doi     = {10.48550/arXiv.2205.01510},
  eprint  = {2205.01510},
  archivePrefix= {arXiv}
}

@article{park.arxiv.2025,
  author  = {Park, Seokhun and Kong, Insung and Choi, Yongchan and Park, Chanmoo and Kim, Yongdai},
  title   = {Tensor product neural networks for functional {ANOVA} model},
  journal = {arXiv preprint arXiv:2502.15215},
  year    = {2025},
  doi     = {10.48550/arXiv.2502.15215},
  eprint  = {2502.15215},
  archivePrefix= {arXiv}
}

@inproceedings{kingma.iclr.2015,
  author    = {Kingma, Diederik P. and Ba, Jimmy},
  title     = {Adam: A method for stochastic optimization},
  booktitle = {International Conference on Learning Representations (ICLR)},
  year      = {2015},
  url       = {https://arxiv.org/abs/1412.6980},
  eprint    = {1412.6980},
  archivePrefix= {arXiv}
}

@article{ha.arxiv.2018,
  author  = {Ha, David and Schmidhuber, J{\"u}rgen},
  title   = {World models},
  journal = {arXiv preprint arXiv:1803.10122},
  year    = {2018},
  doi     = {10.48550/arXiv.1803.10122},
  eprint  = {1803.10122},
  archivePrefix= {arXiv}
}

@article{Friston2018-do,
  author  = {Friston, Karl J. and Rosch, Richard and Parr, Thomas and Price, Cathy and Bowman, Howard},
  title   = {Deep temporal models and active inference},
  journal = {Neuroscience \& Biobehavioral Reviews},
  volume  = {90},
  pages   = {486--501},
  year    = {2018},
  doi     = {10.1016/j.neubiorev.2018.04.004}
}

@article{sakagami.frai.2023,
  author  = {Sakagami, Ryo and Lay, Florian S. and D{\"o}mel, Andreas and Schuster, Martin J. and Albu-Sch{\"a}ffer, Alin and Stulp, Freek},
  title   = {Robotic world models---conceptualization, review, and engineering best practices},
  journal = {Frontiers in Robotics and AI},
  volume  = {10},
  pages   = {1253049},
  year    = {2023},
  doi     = {10.3389/frobt.2023.1253049}
}

@article{ding.arxiv.2024,
  author  = {Ding, Jingtao and Zhang, Yunke and Shang, Yu and Feng, Jie and Zhang, Yuheng and Zong, Zefang and Yuan, Yuan and Su, Hongyuan and Li, Nian and Piao, Jinghua and Deng, Yucheng and Sukiennik, Nicholas and Gao, Chen and Xu, Fengli and Li, Yong},
  title   = {Understanding world or predicting future? {A} comprehensive survey of world models},
  journal = {arXiv preprint arXiv:2411.14499},
  year    = {2024},
  doi     = {10.48550/arXiv.2411.14499},
  eprint  = {2411.14499},
  archivePrefix= {arXiv}
}

@article{guan.arxiv.2024,
  author  = {Guan, Yanchen and Liao, Haicheng and Li, Zhenning and Hu, Jia and Yuan, Runze and Li, Yunjian and Zhang, Guohui and Xu, Chengzhong},
  title   = {World models for autonomous driving: An initial survey},
  journal = {arXiv preprint arXiv:2403.02622},
  year    = {2024},
  doi     = {10.48550/arXiv.2403.02622},
  eprint  = {2403.02622},
  archivePrefix= {arXiv}
}

@article{hou.arxiv.2026,
  author  = {Hou, Bohan and Li, Gen and Jia, Jindou and An, Tuo and Guo, Xinying and Leng, Sicong and Geng, Haoran and Ze, Yanjie and Harada, Tatsuya and Torr, Philip and Mees, Oier and Pollefeys, Marc and Liu, Zhuang and Wu, Jiajun and Abbeel, Pieter and Malik, Jitendra and Du, Yilun and Yang, Jianfei},
  title   = {World model for robot learning: A comprehensive survey},
  journal = {arXiv preprint arXiv:2605.00080},
  year    = {2026},
  doi     = {10.48550/arXiv.2605.00080},
  eprint  = {2605.00080},
  archivePrefix= {arXiv}
}

@article{moerland.arxiv.2020,
  author  = {Moerland, Thomas M. and Broekens, Joost and Plaat, Aske and Jonker, Catholijn M.},
  title   = {Model-based reinforcement learning: A survey},
  journal = {arXiv preprint arXiv:2006.16712},
  year    = {2020},
  doi     = {10.48550/arXiv.2006.16712},
  eprint  = {2006.16712},
  archivePrefix= {arXiv}
}

@article{plaat.arxiv.2021,
  author  = {Plaat, Aske and Kosters, Walter and Preuss, Mike},
  title   = {High-accuracy model-based reinforcement learning, a survey},
  journal = {arXiv preprint arXiv:2107.08241},
  year    = {2021},
  doi     = {10.48550/arXiv.2107.08241},
  eprint  = {2107.08241},
  archivePrefix= {arXiv}
}

@article{han.if.2026,
  author  = {Han, Xiaofeng and Chen, Shunpeng and Fu, Zenghuang and Feng, Zhe and Fan, Lue and An, Dong and Wang, Changwei and Guo, Li and Meng, Weiliang and Zhang, Xiaopeng and Xu, Rongtao and Xu, Shibiao},
  title   = {Multimodal fusion and vision--language models: A survey for robot vision},
  journal = {Information Fusion},
  volume  = {126},
  pages   = {103652},
  year    = {2026},
  url     = {https://arxiv.org/abs/2504.02477},
  eprint  = {2504.02477},
  archivePrefix= {arXiv}
}

@article{kawaharazuka.arxiv.2025,
  author  = {Kawaharazuka, Kento and Oh, Jihoon and Yamada, Jun and Posner, Ingmar and Zhu, Yuke},
  title   = {Vision-language-action models for robotics: A review towards real-world applications},
  journal = {arXiv preprint arXiv:2510.07077},
  year    = {2025},
  doi     = {10.48550/arXiv.2510.07077},
  eprint  = {2510.07077},
  archivePrefix= {arXiv}
}

@article{shao.arxiv.2025,
  author  = {Shao, Rui and Li, Wei and Zhang, Lingsen and Zhang, Renshan and Liu, Zhiyang and Chen, Ran and Nie, Liqiang},
  title   = {Large {VLM}-based vision-language-action models for robotic manipulation: A survey},
  journal = {arXiv preprint arXiv:2508.13073},
  year    = {2025},
  doi     = {10.48550/arXiv.2508.13073},
  eprint  = {2508.13073},
  archivePrefix= {arXiv}
}

@article{celemin.arxiv.2022,
  author  = {Celemin, Carlos and P{\'e}rez-Dattari, Rodrigo and Chisari, Eugenio and Franzese, Giovanni and de Souza Rosa, Leandro and Prakash, Ravi and Ajanovi{\'c}, Zlatan and Ferraz, Marta and Valada, Abhinav and Kober, Jens},
  title   = {Interactive imitation learning in robotics: A survey},
  journal = {arXiv preprint arXiv:2211.00600},
  year    = {2022},
  doi     = {10.48550/arXiv.2211.00600},
  eprint  = {2211.00600},
  archivePrefix= {arXiv}
}

@article{wang.arxiv.2024,
  author  = {Wang, Yuchen and Hayashibe, Mitsuhiro and Owaki, Dai},
  title   = {Data-driven policy learning methods from biological behavior: A systematic review},
  journal = {Applied Sciences},
  volume  = {14},
  number  = {10},
  pages   = {4038},
  year    = {2024},
  doi     = {10.3390/app14104038}
}

@article{wolf.arxiv.2025,
  author  = {Wolf, Rosa and Shi, Yitian and Liu, Sheng and Rayyes, Rania},
  title   = {Diffusion models for robotic manipulation: A survey},
  journal = {arXiv preprint arXiv:2504.08438},
  year    = {2025},
  doi     = {10.48550/arXiv.2504.08438},
  eprint  = {2504.08438},
  archivePrefix= {arXiv}
}

@article{li.arxiv.2026,
  author  = {Li, Jialong and Lind, Simon Kristoffersson and Xie, Wenrui and Stenmark, Maj and Kr{\"u}ger, Volker},
  title   = {Normalizing flows are capable models for bi-manual visuomotor policy},
  journal = {arXiv preprint arXiv:2509.21073},
  year    = {2026},
  doi     = {10.48550/arXiv.2509.21073},
  eprint  = {2509.21073},
  archivePrefix= {arXiv}
}

@article{rasheed.access.2020,
  author  = {Rasheed, Adil and San, Omer and Kvamsdal, Trond},
  title   = {Digital twin: Values, challenges and enablers from a modeling perspective},
  journal = {IEEE Access},
  volume  = {8},
  pages   = {21980--22012},
  year    = {2020},
  doi     = {10.1109/ACCESS.2020.2970143}
}

@inproceedings{dsouza.iros.2001,
  author    = {D'Souza, Aaron and Vijayakumar, Sethu and Schaal, Stefan},
  title     = {Learning inverse kinematics},
  booktitle = {Proceedings of the IEEE/RSJ International Conference on Intelligent Robots and Systems (IROS)},
  pages     = {298--303},
  year      = {2001},
  url       = {https://is.mpg.de/publications/dsouza_iicirs_2001}
}

@book{tarantola.siam.2005,
  author    = {Tarantola, Albert},
  title     = {Inverse Problem Theory and Methods for Model Parameter Estimation},
  publisher = {SIAM},
  address   = {Philadelphia},
  year      = {2005},
  doi       = {10.1137/1.9780898717921}
}

@book{engl.kluwer.1996,
  author    = {Engl, Heinz W. and Hanke, Martin and Neubauer, Andreas},
  title     = {Regularization of Inverse Problems},
  publisher = {Kluwer Academic Publishers},
  address   = {Dordrecht},
  year      = {1996}
}

@article{arridge.actanumerica.2019,
  author  = {Arridge, Simon and Maass, Peter and {\"O}ktem, Ozan and Sch{\"o}nlieb, Carola-Bibiane},
  title   = {Solving inverse problems using data-driven models},
  journal = {Acta Numerica},
  volume  = {28},
  pages   = {1--174},
  year    = {2019},
  doi     = {10.1017/S0962492919000059}
}

@article{ongie.jsait.2020,
  author  = {Ongie, Gregory and Jalal, Ajil and Metzler, Christopher A. and Baraniuk, Richard G. and Dimakis, Alexandros G. and Willett, Rebecca},
  title   = {Deep learning techniques for inverse problems in imaging},
  journal = {IEEE Journal on Selected Areas in Information Theory},
  volume  = {1},
  number  = {1},
  pages   = {39--56},
  year    = {2020},
  url     = {https://arxiv.org/abs/2005.06001},
  eprint  = {2005.06001},
  archivePrefix= {arXiv}
}

@article{cranmer.pnas.2020,
  author  = {Cranmer, Kyle and Brehmer, Johann and Louppe, Gilles},
  title   = {The frontier of simulation-based inference},
  journal = {Proceedings of the National Academy of Sciences},
  volume  = {117},
  number  = {48},
  pages   = {30055--30062},
  year    = {2020},
  doi     = {10.1073/pnas.1912789117}
}

@article{radev.tnnls.2022,
  author  = {Radev, Stefan T. and Mertens, Ulf K. and Voss, Andreas and Ardizzone, Lynton and K{\"o}the, Ullrich},
  title   = {{BayesFlow}: Learning complex stochastic models with invertible neural networks},
  journal = {IEEE Transactions on Neural Networks and Learning Systems},
  volume  = {33},
  number  = {4},
  pages   = {1452--1466},
  year    = {2022},
  url     = {https://arxiv.org/abs/2003.06281},
  eprint  = {2003.06281},
  archivePrefix= {arXiv}
}

@inproceedings{ardizzone.iclr.2019,
  author    = {Ardizzone, Lynton and Kruse, Jakob and Wirkert, Sebastian and Rahner, Daniel and Pellegrini, Eric W. and Klessen, Ralf S. and Maier-Hein, Lena and Rother, Carsten and K{\"o}the, Ullrich},
  title     = {Analyzing inverse problems with invertible neural networks},
  booktitle = {International Conference on Learning Representations (ICLR)},
  year      = {2019},
  url       = {https://arxiv.org/abs/1808.04730},
  eprint    = {1808.04730},
  archivePrefix= {arXiv}
}

@article{adler.ip.2017,
  author  = {Adler, Jonas and {\"O}ktem, Ozan},
  title   = {Solving ill-posed inverse problems using iterative deep neural networks},
  journal = {Inverse Problems},
  volume  = {33},
  number  = {12},
  pages   = {124007},
  year    = {2017},
  doi     = {10.1088/1361-6420/aa9581}
}

@inproceedings{gershman.cogsci.2014,
  author    = {Gershman, Samuel J. and Goodman, Noah D.},
  title     = {Amortized inference in probabilistic reasoning},
  booktitle = {Proceedings of the 36th Annual Conference of the Cognitive Science Society},
  year      = {2014}
}

@inproceedings{cremer.icml.2018,
  author    = {Cremer, Chris and Li, Xuechen and Duvenaud, David},
  title     = {Inference suboptimality in variational autoencoders},
  booktitle = {Proceedings of the 35th International Conference on Machine Learning (ICML)},
  series    = {PMLR},
  volume    = {80},
  year      = {2018},
  url       = {https://arxiv.org/abs/1801.03558},
  eprint    = {1801.03558},
  archivePrefix= {arXiv}
}

@inproceedings{marino.icml.2018,
  author    = {Marino, Joseph and Yue, Yisong and Mandt, Stephan},
  title     = {Iterative amortized inference},
  booktitle = {Proceedings of the 35th International Conference on Machine Learning (ICML)},
  series    = {PMLR},
  volume    = {80},
  year      = {2018},
  url       = {https://arxiv.org/abs/1807.09356},
  eprint    = {1807.09356},
  archivePrefix= {arXiv}
}

@article{yuille.tics.2006,
  author  = {Yuille, Alan and Kersten, Daniel},
  title   = {Vision as {Bayesian} inference: analysis by synthesis?},
  journal = {Trends in Cognitive Sciences},
  volume  = {10},
  number  = {7},
  pages   = {301--308},
  year    = {2006},
  doi     = {10.1016/j.tics.2006.05.002}
}

@inproceedings{kulkarni.neurips.2015,
  author    = {Kulkarni, Tejas D. and Whitney, William F. and Kohli, Pushmeet and Tenenbaum, Josh},
  title     = {Deep convolutional inverse graphics network},
  booktitle = {Advances in Neural Information Processing Systems},
  volume    = {28},
  year      = {2015},
  url       = {https://arxiv.org/abs/1503.03167},
  eprint    = {1503.03167},
  archivePrefix= {arXiv}
}

@inproceedings{loper.eccv.2014,
  author    = {Loper, Matthew M. and Black, Michael J.},
  title     = {{OpenDR}: An approximate differentiable renderer},
  booktitle = {European Conference on Computer Vision (ECCV)},
  series    = {Lecture Notes in Computer Science},
  volume    = {8695},
  pages     = {154--169},
  publisher = {Springer},
  year      = {2014},
  doi       = {10.1007/978-3-319-10584-0_11}
}

@inproceedings{li.deepim.eccv.2018,
  author    = {Li, Yi and Wang, Gu and Ji, Xiangyang and Xiang, Yu and Fox, Dieter},
  title     = {{DeepIM}: Deep iterative matching for {6D} pose estimation},
  booktitle = {European Conference on Computer Vision (ECCV)},
  year      = {2018},
  url       = {https://arxiv.org/abs/1804.00175},
  eprint    = {1804.00175},
  archivePrefix= {arXiv}
}

@article{besl.tpami.1992,
  author  = {Besl, Paul J. and McKay, Neil D.},
  title   = {A method for registration of {3-D} shapes},
  journal = {IEEE Transactions on Pattern Analysis and Machine Intelligence},
  volume  = {14},
  number  = {2},
  pages   = {239--256},
  year    = {1992},
  doi     = {10.1109/34.121791}
}

@article{murphy.arxiv.2021,
  author  = {Murphy, Kieran A. and Jampani, Varun and Ramalingam, Srikumar and Makadia, Ameesh},
  title   = {Learning {ABCs}: Approximate bijective correspondence for isolating factors of variation with weak supervision},
  journal = {arXiv preprint arXiv:2103.03240},
  year    = {2021},
  doi     = {10.48550/arXiv.2103.03240},
  eprint  = {2103.03240},
  archivePrefix= {arXiv}
}

@book{nocedal.springer.2006,
  author    = {Nocedal, Jorge and Wright, Stephen J.},
  title     = {Numerical Optimization},
  edition   = {2nd},
  publisher = {Springer},
  address   = {New York},
  year      = {2006},
  doi       = {10.1007/978-0-387-40065-5}
}

@article{levenberg.qam.1944,
  author  = {Levenberg, Kenneth},
  title   = {A method for the solution of certain non-linear problems in least squares},
  journal = {Quarterly of Applied Mathematics},
  volume  = {2},
  number  = {2},
  pages   = {164--168},
  year    = {1944},
  doi     = {10.1090/qam/10666}
}

@article{marquardt.siam.1963,
  author  = {Marquardt, Donald W.},
  title   = {An algorithm for least-squares estimation of nonlinear parameters},
  journal = {Journal of the Society for Industrial and Applied Mathematics},
  volume  = {11},
  number  = {2},
  pages   = {431--441},
  year    = {1963},
  doi     = {10.1137/0111030}
}

@book{deuflhard.springer.2011,
  author    = {Deuflhard, Peter},
  title     = {Newton Methods for Nonlinear Problems: Affine Invariance and Adaptive Algorithms},
  series    = {Springer Series in Computational Mathematics},
  volume    = {35},
  publisher = {Springer},
  address   = {Berlin},
  year      = {2011},
  doi       = {10.1007/978-3-642-23899-4}
}

@article{cai.sisc.2002,
  author  = {Cai, Xiao-Chuan and Keyes, David E.},
  title   = {Nonlinearly preconditioned inexact {Newton} algorithms},
  journal = {SIAM Journal on Scientific Computing},
  volume  = {24},
  number  = {1},
  pages   = {183--200},
  year    = {2002},
  doi     = {10.1137/S106482750037620X}
}

@article{lee.arxiv.2025,
  author  = {Lee, Youngkyu and Liu, Shanqing and Darbon, Jerome and Karniadakis, George Em},
  title   = {A neural-operator preconditioned {Newton} method for accelerated nonlinear solvers},
  journal = {arXiv preprint arXiv:2511.08811},
  year    = {2025},
  doi     = {10.48550/arXiv.2511.08811},
  eprint  = {2511.08811},
  archivePrefix= {arXiv}
}

@book{allgower.springer.1990,
  author    = {Allgower, Eugene L. and Georg, Kurt},
  title     = {Numerical Continuation Methods: An Introduction},
  publisher = {Springer},
  address   = {Berlin},
  year      = {1990},
  doi       = {10.1007/978-3-642-61257-2}
}

@inproceedings{bengio.icml.2009,
  author    = {Bengio, Yoshua and Louradour, J{\'e}r{\^o}me and Collobert, Ronan and Weston, Jason},
  title     = {Curriculum learning},
  booktitle = {Proceedings of the 26th International Conference on Machine Learning (ICML)},
  pages     = {41--48},
  year      = {2009},
  doi       = {10.1145/1553374.1553380}
}

@article{virieux.geophysics.2009,
  author  = {Virieux, Jean and Operto, St{\'e}phane},
  title   = {An overview of full-waveform inversion in exploration geophysics},
  journal = {Geophysics},
  volume  = {74},
  number  = {6},
  pages   = {WCC1--WCC26},
  year    = {2009},
  doi     = {10.1190/1.3238367}
}

@article{engquist.cms.2014,
  author  = {Engquist, Bj{\"o}rn and Froese, Brittany D.},
  title   = {Application of the {Wasserstein} metric to seismic signals},
  journal = {Communications in Mathematical Sciences},
  volume  = {12},
  number  = {5},
  pages   = {979--988},
  year    = {2014},
  url     = {https://arxiv.org/abs/1311.4581},
  eprint  = {1311.4581},
  archivePrefix= {arXiv}
}

@article{sun.frontiers.2022,
  author  = {Sun, Bingbing and Alkhalifah, Tariq},
  title   = {{ML-misfit}: A neural network formulation of the misfit function for full-waveform inversion},
  journal = {Frontiers in Earth Science},
  volume  = {10},
  pages   = {1011825},
  year    = {2022},
  doi     = {10.3389/feart.2022.1011825}
}

@article{meisters.duke.1963,
  author  = {Meisters, Gary H. and Olech, Czes{\l}aw},
  title   = {Locally one-to-one mappings and a classical theorem on schlicht functions},
  journal = {Duke Mathematical Journal},
  volume  = {30},
  number  = {1},
  pages   = {63--80},
  year    = {1963}
}

@article{gale.mathann.1965,
  author  = {Gale, David and Nikaid{\^o}, Hukukane},
  title   = {The {Jacobian} matrix and global univalence of mappings},
  journal = {Mathematische Annalen},
  volume  = {159},
  number  = {2},
  pages   = {81--93},
  year    = {1965},
  doi     = {10.1007/BF01360282}
}

@book{parthasarathy.lnm.1983,
  author    = {Parthasarathy, T.},
  title     = {On Global Univalence Theorems},
  series    = {Lecture Notes in Mathematics},
  volume    = {977},
  publisher = {Springer},
  address   = {Berlin},
  year      = {1983},
  doi       = {10.1007/BFb0065571}
}

@article{federer.tams.1959,
  author  = {Federer, Herbert},
  title   = {Curvature measures},
  journal = {Transactions of the American Mathematical Society},
  volume  = {93},
  number  = {3},
  pages   = {418--491},
  year    = {1959},
  doi     = {10.1090/S0002-9947-1959-0110078-1}
}

@book{golub.jhu.2013,
  author    = {Golub, Gene H. and Van Loan, Charles F.},
  title     = {Matrix Computations},
  edition   = {4th},
  publisher = {Johns Hopkins University Press},
  address   = {Baltimore},
  year      = {2013}
}

@article{brent.sisc.1985,
  author  = {Brent, Richard P. and Luk, Franklin T.},
  title   = {The solution of singular-value and symmetric eigenvalue problems on multiprocessor arrays},
  journal = {SIAM Journal on Scientific and Statistical Computing},
  volume  = {6},
  number  = {1},
  pages   = {69--84},
  year    = {1985},
  doi     = {10.1137/0906007}
}

@article{papamakarios.jmlr.2021,
  author  = {Papamakarios, George and Nalisnick, Eric and Rezende, Danilo Jimenez and Mohamed, Shakir and Lakshminarayanan, Balaji},
  title   = {Normalizing flows for probabilistic modeling and inference},
  journal = {Journal of Machine Learning Research},
  volume  = {22},
  number  = {57},
  pages   = {1--64},
  year    = {2021},
  url     = {https://jmlr.org/papers/v22/19-1028.html},
  eprint  = {1912.02762},
  archivePrefix= {arXiv}
}

@article{kobyzev.tpami.2021,
  author  = {Kobyzev, Ivan and Prince, Simon J. D. and Brubaker, Marcus A.},
  title   = {Normalizing flows: An introduction and review of current methods},
  journal = {IEEE Transactions on Pattern Analysis and Machine Intelligence},
  volume  = {43},
  number  = {11},
  pages   = {3964--3979},
  year    = {2021},
  doi     = {10.1109/TPAMI.2020.2992934},
  eprint  = {1908.09257},
  archivePrefix= {arXiv}
}

@article{dinh.arxiv.2014,
  author  = {Dinh, Laurent and Krueger, David and Bengio, Yoshua},
  title   = {{NICE}: Non-linear independent components estimation},
  journal = {arXiv preprint arXiv:1410.8516},
  year    = {2014},
  doi     = {10.48550/arXiv.1410.8516},
  eprint  = {1410.8516},
  archivePrefix= {arXiv}
}

@inproceedings{dinh.arxiv.2016,
  author    = {Dinh, Laurent and Sohl-Dickstein, Jascha and Bengio, Samy},
  title     = {Density estimation using {Real NVP}},
  booktitle = {International Conference on Learning Representations (ICLR)},
  year      = {2017},
  url       = {https://arxiv.org/abs/1605.08803},
  eprint    = {1605.08803},
  archivePrefix= {arXiv}
}

@inproceedings{kingma.arxiv.2018,
  author    = {Kingma, Diederik P. and Dhariwal, Prafulla},
  title     = {Glow: Generative flow with invertible $1\times1$ convolutions},
  booktitle = {Advances in Neural Information Processing Systems},
  volume    = {31},
  year      = {2018},
  url       = {https://arxiv.org/abs/1807.03039},
  eprint    = {1807.03039},
  archivePrefix= {arXiv}
}

@inproceedings{rezende.icml.2015,
  author    = {Rezende, Danilo Jimenez and Mohamed, Shakir},
  title     = {Variational inference with normalizing flows},
  booktitle = {Proceedings of the 32nd International Conference on Machine Learning (ICML)},
  series    = {PMLR},
  volume    = {37},
  pages     = {1530--1538},
  year      = {2015},
  url       = {https://arxiv.org/abs/1505.05770},
  eprint    = {1505.05770},
  archivePrefix= {arXiv}
}

@inproceedings{behrmann.icml.2019,
  author    = {Behrmann, Jens and Grathwohl, Will and Chen, Ricky T. Q. and Duvenaud, David and Jacobsen, J{\"o}rn-Henrik},
  title     = {Invertible residual networks},
  booktitle = {Proceedings of the 36th International Conference on Machine Learning (ICML)},
  series    = {PMLR},
  volume    = {97},
  pages     = {573--582},
  year      = {2019},
  url       = {https://arxiv.org/abs/1811.00995},
  eprint    = {1811.00995},
  archivePrefix= {arXiv}
}

@inproceedings{behrmann.aistats.2021,
  author    = {Behrmann, Jens and Vicol, Paul and Wang, Kuan-Chieh and Grosse, Roger and Jacobsen, J{\"o}rn-Henrik},
  title     = {Understanding and mitigating exploding inverses in invertible neural networks},
  booktitle = {Proceedings of the 24th International Conference on Artificial Intelligence and Statistics (AISTATS)},
  series    = {PMLR},
  volume    = {130},
  pages     = {1792--1800},
  year      = {2021},
  url       = {https://arxiv.org/abs/2006.09347},
  eprint    = {2006.09347},
  archivePrefix= {arXiv}
}

@inproceedings{gomez.neurips.2017,
  author    = {Gomez, Aidan N. and Ren, Mengye and Urtasun, Raquel and Grosse, Roger B.},
  title     = {The reversible residual network: Backpropagation without storing activations},
  booktitle = {Advances in Neural Information Processing Systems},
  volume    = {30},
  year      = {2017},
  url       = {https://arxiv.org/abs/1707.04585},
  eprint    = {1707.04585},
  archivePrefix= {arXiv}
}

@inproceedings{jacobsen.iclr.2018,
  author    = {Jacobsen, J{\"o}rn-Henrik and Smeulders, Arnold and Oyallon, Edouard},
  title     = {i-{RevNet}: Deep invertible networks},
  booktitle = {International Conference on Learning Representations (ICLR)},
  year      = {2018},
  url       = {https://arxiv.org/abs/1802.07088},
  eprint    = {1802.07088},
  archivePrefix= {arXiv}
}

@inproceedings{teshima.neurips.2020,
  author    = {Teshima, Takeshi and Ishikawa, Isao and Tojo, Koichi and Oono, Kenta and Ikeda, Masahiro and Sugiyama, Masashi},
  title     = {Coupling-based invertible neural networks are universal diffeomorphism approximators},
  booktitle = {Advances in Neural Information Processing Systems},
  volume    = {33},
  year      = {2020},
  url       = {https://arxiv.org/abs/2006.11469},
  eprint    = {2006.11469},
  archivePrefix= {arXiv}
}

@inproceedings{dupont.neurips.2019,
  author    = {Dupont, Emilien and Doucet, Arnaud and Teh, Yee Whye},
  title     = {Augmented neural {ODEs}},
  booktitle = {Advances in Neural Information Processing Systems},
  volume    = {32},
  year      = {2019},
  url       = {https://arxiv.org/abs/1904.01681},
  eprint    = {1904.01681},
  archivePrefix= {arXiv}
}

@inproceedings{jing.icml.2017,
  author    = {Jing, Li and Shen, Yichen and Dub{\v{c}}ek, Tena and Peurifoy, John and Skirlo, Scott and LeCun, Yann and Tegmark, Max and Solja{\v{c}}i{\'c}, Marin},
  title     = {Tunable efficient unitary neural networks ({EUNN}) and their application to {RNNs}},
  booktitle = {Proceedings of the 34th International Conference on Machine Learning (ICML)},
  series    = {PMLR},
  volume    = {70},
  pages     = {1733--1741},
  year      = {2017},
  url       = {https://arxiv.org/abs/1612.05231},
  eprint    = {1612.05231},
  archivePrefix= {arXiv}
}

@inproceedings{freifeld.iccv.2015,
  author    = {Freifeld, Oren and Hauberg, S{\o}ren and Batmanghelich, Kayhan and Fisher, III, John W.},
  title     = {Highly-expressive spaces of well-behaved transformations: Keeping it simple},
  booktitle = {Proceedings of the IEEE International Conference on Computer Vision (ICCV)},
  year      = {2015},
  doi       = {10.1109/ICCV.2015.333}
}

@article{freifeld.tpami.2017,
  author  = {Freifeld, Oren and Hauberg, S{\o}ren and Batmanghelich, Kayhan and Fisher, III, John W.},
  title   = {Transformations based on continuous piecewise-affine velocity fields},
  journal = {IEEE Transactions on Pattern Analysis and Machine Intelligence},
  volume  = {39},
  number  = {12},
  pages   = {2496--2509},
  year    = {2017},
  doi     = {10.1109/TPAMI.2016.2646685}
}

@inproceedings{shapiraweber.neurips.2019,
  author    = {Shapira Weber, Ron A. and Eyal, Matan and Skafte Detlefsen, Nicki and Shriki, Oren and Freifeld, Oren},
  title     = {Diffeomorphic temporal alignment nets},
  booktitle = {Advances in Neural Information Processing Systems},
  volume    = {32},
  year      = {2019},
  url       = {https://proceedings.neurips.cc/paper/2019/hash/db98dc0dbafde48e8f74c0de001d35e4-Abstract.html}
}

@article{beg.ijcv.2005,
  author  = {Beg, M. Faisal and Miller, Michael I. and Trouv{\'e}, Alain and Younes, Laurent},
  title   = {Computing large deformation metric mappings via geodesic flows of diffeomorphisms},
  journal = {International Journal of Computer Vision},
  volume  = {61},
  number  = {2},
  pages   = {139--157},
  year    = {2005},
  doi     = {10.1023/B:VISI.0000043755.93987.aa}
}

@inproceedings{rifai.icml.2011,
  author    = {Rifai, Salah and Vincent, Pascal and Muller, Xavier and Glorot, Xavier and Bengio, Yoshua},
  title     = {Contractive auto-encoders: Explicit invariance during feature extraction},
  booktitle = {Proceedings of the 28th International Conference on Machine Learning (ICML)},
  pages     = {833--840},
  year      = {2011},
  url       = {https://mlanthology.org/icml/2011/rifai2011icml-contractive/}
}

@inproceedings{simard.neurips.1992,
  author    = {Simard, Patrice and Victorri, Bernard and LeCun, Yann and Denker, John},
  title     = {Tangent prop -- A formalism for specifying selected invariances in an adaptive network},
  booktitle = {Advances in Neural Information Processing Systems},
  volume    = {4},
  pages     = {895--903},
  year      = {1992},
  url       = {https://proceedings.neurips.cc/paper/1991/hash/65658fde58ab3c2b6e5132a39fae7cb9-Abstract.html}
}

@article{hoffman.arxiv.2019,
  author  = {Hoffman, Judy and Roberts, Daniel A. and Yaida, Sho},
  title   = {Robust learning with {Jacobian} regularization},
  journal = {arXiv preprint arXiv:1908.02729},
  year    = {2019},
  doi     = {10.48550/arXiv.1908.02729},
  eprint  = {1908.02729},
  archivePrefix= {arXiv}
}

@article{hutchinson.1989,
  author  = {Hutchinson, Michael F.},
  title   = {A stochastic estimator of the trace of the influence matrix for {Laplacian} smoothing splines},
  journal = {Communications in Statistics -- Simulation and Computation},
  volume  = {18},
  number  = {3},
  pages   = {1059--1076},
  year    = {1989},
  doi     = {10.1080/03610918908812806}
}

@article{drucker.tnn.1992,
  author  = {Drucker, Harris and LeCun, Yann},
  title   = {Improving generalization performance using double backpropagation},
  journal = {IEEE Transactions on Neural Networks},
  volume  = {3},
  number  = {6},
  pages   = {991--997},
  year    = {1992},
  doi     = {10.1109/72.165600}
}

@inproceedings{czarnecki.neurips.2017,
  author    = {Czarnecki, Wojciech M. and Osindero, Simon and Jaderberg, Max and {\'S}wirszcz, Grzegorz and Pascanu, Razvan},
  title     = {Sobolev training for neural networks},
  booktitle = {Advances in Neural Information Processing Systems},
  volume    = {30},
  year      = {2017},
  url       = {https://arxiv.org/abs/1706.04859},
  eprint    = {1706.04859},
  archivePrefix= {arXiv}
}

@book{griewank.siam.2008,
  author    = {Griewank, Andreas and Walther, Andrea},
  title     = {Evaluating Derivatives: Principles and Techniques of Algorithmic Differentiation},
  edition   = {2nd},
  publisher = {SIAM},
  address   = {Philadelphia},
  year      = {2008},
  doi       = {10.1137/1.9780898717761}
}

@article{baydin.jmlr.2018,
  author  = {Baydin, At{\i}l{\i}m G{\"u}ne{\c{s}} and Pearlmutter, Barak A. and Radul, Alexey Andreyevich and Siskind, Jeffrey Mark},
  title   = {Automatic differentiation in machine learning: A survey},
  journal = {Journal of Machine Learning Research},
  volume  = {18},
  number  = {153},
  pages   = {1--43},
  year    = {2018},
  url     = {https://jmlr.org/papers/v18/17-468.html},
  eprint  = {1502.05767},
  archivePrefix= {arXiv}
}

@inproceedings{hadsell.cvpr.2006,
  author    = {Hadsell, Raia and Chopra, Sumit and LeCun, Yann},
  title     = {Dimensionality reduction by learning an invariant mapping},
  booktitle = {IEEE Computer Society Conference on Computer Vision and Pattern Recognition (CVPR)},
  pages     = {1735--1742},
  year      = {2006},
  doi       = {10.1109/CVPR.2006.100}
}

@inproceedings{higgins.arxiv.2017,
  author    = {Higgins, Irina and Matthey, Loic and Pal, Arka and Burgess, Christopher and Glorot, Xavier and Botvinick, Matthew and Mohamed, Shakir and Lerchner, Alexander},
  title     = {beta-{VAE}: Learning basic visual concepts with a constrained variational framework},
  booktitle = {International Conference on Learning Representations (ICLR)},
  year      = {2017},
  url       = {https://openreview.net/forum?id=Sy2fzU9gl}
}

@inproceedings{chen.arxiv.2018,
  author    = {Chen, Ricky T. Q. and Li, Xuechen and Grosse, Roger and Duvenaud, David},
  title     = {Isolating sources of disentanglement in variational autoencoders},
  booktitle = {Advances in Neural Information Processing Systems},
  volume    = {31},
  year      = {2018},
  url       = {https://arxiv.org/abs/1802.04942},
  eprint    = {1802.04942},
  archivePrefix= {arXiv}
}

@inproceedings{kim.arxiv.2018,
  author    = {Kim, Hyunjik and Mnih, Andriy},
  title     = {Disentangling by factorising},
  booktitle = {Proceedings of the 35th International Conference on Machine Learning (ICML)},
  series    = {PMLR},
  volume    = {80},
  year      = {2018},
  url       = {https://arxiv.org/abs/1802.05983},
  eprint    = {1802.05983},
  archivePrefix= {arXiv}
}

@article{hadjeres.arxiv.2017,
  author  = {Hadjeres, Ga{\"e}tan and Nielsen, Frank and Pachet, Fran{\c{c}}ois},
  title   = {{GLSR-VAE}: Geodesic latent space regularization for variational autoencoder architectures},
  journal = {arXiv preprint arXiv:1707.04588},
  year    = {2017},
  doi     = {10.48550/arXiv.1707.04588},
  eprint  = {1707.04588},
  archivePrefix= {arXiv}
}

@inproceedings{arvanitidis.arxiv.2017,
  author    = {Arvanitidis, Georgios and Hansen, Lars Kai and Hauberg, S{\o}ren},
  title     = {Latent space oddity: On the curvature of deep generative models},
  booktitle = {International Conference on Learning Representations (ICLR)},
  year      = {2018},
  url       = {https://arxiv.org/abs/1710.11379},
  eprint    = {1710.11379},
  archivePrefix= {arXiv}
}

@inproceedings{miani.neurlips.2022,
  author    = {Miani, Marco and Warburg, Frederik and Moreno-Mu{\~n}oz, Pablo and Skafte Detlefsen, Nicki and Hauberg, S{\o}ren},
  title     = {Laplacian autoencoders for learning stochastic representations},
  booktitle = {Advances in Neural Information Processing Systems},
  volume    = {35},
  pages     = {21059--21072},
  year      = {2022},
  url       = {https://proceedings.neurips.cc/paper_files/paper/2022/hash/84880289c9fcba0d4bdb198cdb8f5080-Abstract.html},
  eprint    = {2206.15078},
  archivePrefix= {arXiv}
}

@inproceedings{gallego-posada.2021,
  author    = {Gallego-Posada, Jose and Forr{\'e}, Patrick},
  title     = {Simplicial regularization},
  booktitle = {ICLR 2021 Workshop on Geometrical and Topological Representation Learning},
  year      = {2021},
  url       = {https://mlanthology.org/iclrw/2021/gallegoposada2021iclrw-simplicial/}
}

@article{rotman.arxiv.2022,
  author  = {Rotman, Michael and Dekel, Amit and Gur, Shir and Oz, Yaron and Wolf, Lior},
  title   = {Unsupervised disentanglement with tensor product representations on the torus},
  journal = {arXiv preprint arXiv:2202.06201},
  year    = {2022},
  doi     = {10.48550/arXiv.2202.06201},
  eprint  = {2202.06201},
  archivePrefix= {arXiv}
}

@article{higgins.arxiv.2022,
  author  = {Higgins, Irina and Racani{\`e}re, S{\'e}bastien and Rezende, Danilo},
  title   = {Symmetry-based representations for artificial and biological general intelligence},
  journal = {arXiv preprint arXiv:2203.09250},
  year    = {2022},
  doi     = {10.48550/arXiv.2203.09250},
  eprint  = {2203.09250},
  archivePrefix= {arXiv}
}

@inproceedings{rombach.arxiv.2021,
  author    = {Rombach, Robin and Blattmann, Andreas and Lorenz, Dominik and Esser, Patrick and Ommer, Bj{\"o}rn},
  title     = {High-resolution image synthesis with latent diffusion models},
  booktitle = {Proceedings of the IEEE/CVF Conference on Computer Vision and Pattern Recognition (CVPR)},
  year      = {2022},
  url       = {https://arxiv.org/abs/2112.10752},
  eprint    = {2112.10752},
  archivePrefix= {arXiv}
}

@inproceedings{alemi.arxiv.2017,
  author    = {Alemi, Alexander A. and Fischer, Ian and Dillon, Joshua V. and Murphy, Kevin},
  title     = {Deep variational information bottleneck},
  booktitle = {International Conference on Learning Representations (ICLR)},
  year      = {2017},
  url       = {https://arxiv.org/abs/1612.00410},
  eprint    = {1612.00410},
  archivePrefix= {arXiv}
}

@inproceedings{kouzelis.2025,
  author    = {Kouzelis, Theodoros and Kakogeorgiou, Ioannis and Gidaris, Spyros and Komodakis, Nikos},
  title     = {{EQ-VAE}: Equivariance regularized latent space for improved generative image modeling},
  booktitle = {Proceedings of the 42nd International Conference on Machine Learning (ICML)},
  series    = {PMLR},
  volume    = {267},
  pages     = {31648--31666},
  year      = {2025},
  url       = {https://proceedings.mlr.press/v267/kouzelis25a.html},
  eprint    = {2502.09509},
  archivePrefix= {arXiv}
}

@misc{polianskii.thesis.2018,
  author = {Polianskii, Vladislav},
  title  = {An investigation of neural network structure with topological data analysis},
  note   = {Master's thesis},
  year   = {2018}
}

@inproceedings{gabrielsson.2020,
  author    = {Gabrielsson, Rickard Br{\"u}el and Nelson, Bradley J. and Dwaraknath, Anjan and Skraba, Primoz},
  title     = {A topology layer for machine learning},
  booktitle = {Proceedings of the 23rd International Conference on Artificial Intelligence and Statistics (AISTATS)},
  series    = {PMLR},
  volume    = {108},
  pages     = {1553--1563},
  year      = {2020},
  url       = {https://arxiv.org/abs/1905.12200},
  eprint    = {1905.12200},
  archivePrefix= {arXiv}
}

@inproceedings{hofer.2019,
  author    = {Hofer, Christoph and Kwitt, Roland and Dixit, Mandar and Niethammer, Marc},
  title     = {Connectivity-optimized representation learning via persistent homology},
  booktitle = {Proceedings of the 36th International Conference on Machine Learning (ICML)},
  series    = {PMLR},
  volume    = {97},
  year      = {2019},
  url       = {https://proceedings.mlr.press/v97/hofer19a.html},
  eprint    = {1906.09003},
  archivePrefix= {arXiv}
}

@article{medbouhi.mdpi.2023,
  author  = {Medbouhi, Aniss Aiman and Polianskii, Vladislav and Varava, Anastasia and Kragic, Danica},
  title   = {{InvMap} and witness simplicial variational auto-encoders},
  journal = {Machine Learning and Knowledge Extraction},
  volume  = {5},
  number  = {1},
  pages   = {199--236},
  year    = {2023},
  doi     = {10.3390/make5010014}
}

@article{hauschultz.arxiv.2022,
  author  = {Hauschultz, Helene and Palm, Rasmus Berg and Moreno-Mu{\~n}oz, Pablo and Skafte Detlefsen, Nicki and du Plessis, Andrew Allan and Hauberg, S{\o}ren},
  title   = {Is an encoder within reach?},
  journal = {arXiv preprint arXiv:2206.01552},
  year    = {2022},
  doi     = {10.48550/arXiv.2206.01552},
  eprint  = {2206.01552},
  archivePrefix= {arXiv}
}

@article{gropp.arxiv.2020,
  author  = {Gropp, Amos and Atzmon, Matan and Lipman, Yaron},
  title   = {Isometric autoencoders},
  journal = {arXiv preprint arXiv:2006.09289},
  year    = {2020},
  doi     = {10.48550/arXiv.2006.09289},
  eprint  = {2006.09289},
  archivePrefix= {arXiv}
}

@article{hyvarinen.nn.1999,
  author  = {Hyv{\"a}rinen, Aapo and Pajunen, Petteri},
  title   = {Nonlinear independent component analysis: Existence and uniqueness results},
  journal = {Neural Networks},
  volume  = {12},
  number  = {3},
  pages   = {429--439},
  year    = {1999},
  doi     = {10.1016/S0893-6080(98)00140-3}
}

@inproceedings{khemakhem.aistats.2020,
  author    = {Khemakhem, Ilyes and Kingma, Diederik and Monti, Ricardo and Hyv{\"a}rinen, Aapo},
  title     = {Variational autoencoders and nonlinear {ICA}: A unifying framework},
  booktitle = {Proceedings of the 23rd International Conference on Artificial Intelligence and Statistics (AISTATS)},
  series    = {PMLR},
  volume    = {108},
  pages     = {2207--2217},
  year      = {2020},
  url       = {https://arxiv.org/abs/1907.04809},
  eprint    = {1907.04809},
  archivePrefix= {arXiv}
}

@inproceedings{goodfellow.arxiv.2014,
  author    = {Goodfellow, Ian J. and Shlens, Jonathon and Szegedy, Christian},
  title     = {Explaining and harnessing adversarial examples},
  booktitle = {International Conference on Learning Representations (ICLR)},
  year      = {2015},
  url       = {https://arxiv.org/abs/1412.6572},
  eprint    = {1412.6572},
  archivePrefix= {arXiv}
}

@inproceedings{madry.arxiv.2017,
  author    = {Madry, Aleksander and Makelov, Aleksandar and Schmidt, Ludwig and Tsipras, Dimitris and Vladu, Adrian},
  title     = {Towards deep learning models resistant to adversarial attacks},
  booktitle = {International Conference on Learning Representations (ICLR)},
  year      = {2018},
  url       = {https://arxiv.org/abs/1706.06083},
  eprint    = {1706.06083},
  archivePrefix= {arXiv}
}

@article{rahimian.arxiv.2019,
  author  = {Rahimian, Hamed and Mehrotra, Sanjay},
  title   = {Frameworks and results in distributionally robust optimization},
  journal = {Open Journal of Mathematical Optimization},
  volume  = {3},
  pages   = {1--85},
  year    = {2022},
  url     = {https://arxiv.org/abs/1908.05659},
  eprint  = {1908.05659},
  archivePrefix= {arXiv}
}

@techreport{settles.tr.2009,
  author      = {Settles, Burr},
  title       = {Active Learning Literature Survey},
  institution = {University of Wisconsin--Madison},
  type        = {Computer Sciences Technical Report},
  number      = {1648},
  year        = {2009},
  url         = {https://minds.wisconsin.edu/handle/1793/60660}
}

@misc{twincher-itodd,
  author       = {Gonoskov, Arkady},
  title        = {{6D} pose estimation from depth images with twinchers: {ITODD} benchmark},
  year         = {2026},
  howpublished = {Research log with scripts and configurations, \url{https://github.com/arkady-gonoskov/twincher/applications/itodd/}}
}

@inproceedings{drost.iccvw.2017,
  author    = {Drost, Bertram and Ulrich, Markus and Bergmann, Paul and H{\"a}rtinger, Philipp and Steger, Carsten},
  title     = {Introducing {MVTec ITODD} -- A dataset for {3D} object recognition in industry},
  booktitle = {IEEE International Conference on Computer Vision Workshops (ICCVW)},
  pages     = {2200--2208},
  year      = {2017},
  doi       = {10.1109/ICCVW.2017.257}
}

@inproceedings{hodan.eccvw.2016,
  author    = {Hoda{\v{n}}, Tom{\'a}{\v{s}} and Matas, Ji{\v{r}}{\'\i} and Obdr{\v{z}}{\'a}lek, {\v{S}}t{\v{e}}p{\'a}n},
  title     = {On evaluation of {6D} object pose estimation},
  booktitle = {European Conference on Computer Vision Workshops (ECCVW)},
  pages     = {606--619},
  year      = {2016},
  doi       = {10.1007/978-3-319-49409-8_52}
}

@inproceedings{hodan.eccv.2018,
  author    = {Hoda{\v{n}}, Tom{\'a}{\v{s}} and Michel, Frank and Brachmann, Eric and Kehl, Wadim and Glent Buch, Anders and Kraft, Dirk and Drost, Bertram and Vidal, Joel and Ihrke, Stephan and Zabulis, Xenophon and Sahin, Caner and Manhardt, Fabian and Tombari, Federico and Kim, Tae-Kyun and Matas, Ji{\v{r}}{\'\i} and Rother, Carsten},
  title     = {{BOP}: Benchmark for {6D} object pose estimation},
  booktitle = {European Conference on Computer Vision (ECCV)},
  year      = {2018},
  url       = {https://arxiv.org/abs/1808.08319},
  eprint    = {1808.08319},
  archivePrefix= {arXiv}
}

@inproceedings{hodan.eccvw.2020,
  author    = {Hoda{\v{n}}, Tom{\'a}{\v{s}} and Sundermeyer, Martin and Drost, Bertram and Labb{\'e}, Yann and Brachmann, Eric and Michel, Frank and Rother, Carsten and Matas, Ji{\v{r}}{\'\i}},
  title     = {{BOP} Challenge 2020 on {6D} object localization},
  booktitle = {European Conference on Computer Vision Workshops (ECCVW)},
  year      = {2020},
  url       = {https://arxiv.org/abs/2009.07378},
  eprint    = {2009.07378},
  archivePrefix= {arXiv}
}

@inproceedings{sundermeyer.cvprw.2023,
  author    = {Sundermeyer, Martin and Hoda{\v{n}}, Tom{\'a}{\v{s}} and Labb{\'e}, Yann and Wang, Gu and Brachmann, Eric and Drost, Bertram and Rother, Carsten and Matas, Ji{\v{r}}{\'\i}},
  title     = {{BOP} Challenge 2022 on detection, segmentation and pose estimation of specific rigid objects},
  booktitle = {IEEE/CVF Conference on Computer Vision and Pattern Recognition Workshops (CVPRW)},
  pages     = {2785--2794},
  year      = {2023},
  url       = {https://arxiv.org/abs/2302.13075},
  eprint    = {2302.13075},
  archivePrefix= {arXiv}
}

@inproceedings{hodan.cvprw.2024,
  author    = {Hoda{\v{n}}, Tom{\'a}{\v{s}} and Sundermeyer, Martin and Labb{\'e}, Yann and Nguyen, Van Nguyen and Wang, Gu and Brachmann, Eric and Drost, Bertram and Lepetit, Vincent and Rother, Carsten and Matas, Ji{\v{r}}{\'\i}},
  title     = {{BOP} Challenge 2023 on detection, segmentation and pose estimation of seen and unseen rigid objects},
  booktitle = {IEEE/CVF Conference on Computer Vision and Pattern Recognition Workshops (CVPRW)},
  year      = {2024},
  url       = {https://arxiv.org/abs/2403.09799},
  eprint    = {2403.09799},
  archivePrefix= {arXiv}
}

@inproceedings{drost.cvpr.2010,
  author    = {Drost, Bertram and Ulrich, Markus and Navab, Nassir and Ilic, Slobodan},
  title     = {Model globally, match locally: Efficient and robust {3D} object recognition},
  booktitle = {IEEE Conference on Computer Vision and Pattern Recognition (CVPR)},
  pages     = {998--1005},
  year      = {2010},
  doi       = {10.1109/CVPR.2010.5540108}
}

@article{vidal.sensors.2018,
  author  = {Vidal, Joel and Lin, Chyi-Yeu and Llad{\'o}, Xavier and Mart{\'\i}, Robert},
  title   = {A method for {6D} pose estimation of free-form rigid objects using point pair features on range data},
  journal = {Sensors},
  volume  = {18},
  number  = {8},
  pages   = {2678},
  year    = {2018},
  doi     = {10.3390/s18082678}
}

@inproceedings{hinterstoisser.accv.2012,
  author    = {Hinterstoisser, Stefan and Lepetit, Vincent and Ilic, Slobodan and Holzer, Stefan and Bradski, Gary and Konolige, Kurt and Navab, Nassir},
  title     = {Model based training, detection and pose estimation of texture-less {3D} objects in heavily cluttered scenes},
  booktitle = {Asian Conference on Computer Vision (ACCV)},
  pages     = {548--562},
  year      = {2012},
  doi       = {10.1007/978-3-642-37331-2_42}
}

@inproceedings{xiang.rss.2018,
  author    = {Xiang, Yu and Schmidt, Tanner and Narayanan, Venkatraman and Fox, Dieter},
  title     = {{PoseCNN}: A convolutional neural network for {6D} object pose estimation in cluttered scenes},
  booktitle = {Robotics: Science and Systems (RSS)},
  year      = {2018},
  url       = {https://arxiv.org/abs/1711.00199},
  eprint    = {1711.00199},
  archivePrefix= {arXiv}
}

@inproceedings{peng.cvpr.2019,
  author    = {Peng, Sida and Liu, Yuan and Huang, Qixing and Zhou, Xiaowei and Bao, Hujun},
  title     = {{PVNet}: Pixel-wise voting network for {6DoF} pose estimation},
  booktitle = {IEEE/CVF Conference on Computer Vision and Pattern Recognition (CVPR)},
  year      = {2019},
  url       = {https://arxiv.org/abs/1812.11788},
  eprint    = {1812.11788},
  archivePrefix= {arXiv}
}

@inproceedings{hodan.cvpr.2020,
  author    = {Hoda{\v{n}}, Tom{\'a}{\v{s}} and Bar{\'a}th, D{\'a}niel and Matas, Ji{\v{r}}{\'\i}},
  title     = {{EPOS}: Estimating {6D} pose of objects with symmetries},
  booktitle = {IEEE/CVF Conference on Computer Vision and Pattern Recognition (CVPR)},
  year      = {2020},
  url       = {https://arxiv.org/abs/2004.00605},
  eprint    = {2004.00605},
  archivePrefix= {arXiv}
}

@inproceedings{wang.cvpr.2021,
  author    = {Wang, Gu and Manhardt, Fabian and Tombari, Federico and Ji, Xiangyang},
  title     = {{GDR-Net}: Geometry-guided direct regression network for monocular {6D} object pose estimation},
  booktitle = {IEEE/CVF Conference on Computer Vision and Pattern Recognition (CVPR)},
  year      = {2021},
  url       = {https://arxiv.org/abs/2102.12145},
  eprint    = {2102.12145},
  archivePrefix= {arXiv}
}

@inproceedings{haugaard.cvpr.2022,
  author    = {Haugaard, Rasmus Laurvig and Buch, Anders Glent},
  title     = {{SurfEmb}: Dense and continuous correspondence distributions for object pose estimation with learnt surface embeddings},
  booktitle = {IEEE/CVF Conference on Computer Vision and Pattern Recognition (CVPR)},
  year      = {2022},
  url       = {https://arxiv.org/abs/2111.13489},
  eprint    = {2111.13489},
  archivePrefix= {arXiv}
}

@inproceedings{su.cvpr.2022,
  author    = {Su, Yongzhi and Saleh, Mahdi and Fetzer, Torben and Rambach, Jason and Navab, Nassir and Busam, Benjamin and Stricker, Didier and Tombari, Federico},
  title     = {{ZebraPose}: Coarse to fine surface encoding for {6DoF} object pose estimation},
  booktitle = {IEEE/CVF Conference on Computer Vision and Pattern Recognition (CVPR)},
  year      = {2022},
  url       = {https://arxiv.org/abs/2203.09418},
  eprint    = {2203.09418},
  archivePrefix= {arXiv}
}

@inproceedings{labbe.eccv.2020,
  author    = {Labb{\'e}, Yann and Carpentier, Justin and Aubry, Mathieu and Sivic, Josef},
  title     = {{CosyPose}: Consistent multi-view multi-object {6D} pose estimation},
  booktitle = {European Conference on Computer Vision (ECCV)},
  year      = {2020},
  url       = {https://arxiv.org/abs/2008.08465},
  eprint    = {2008.08465},
  archivePrefix= {arXiv}
}

@inproceedings{labbe.corl.2022,
  author    = {Labb{\'e}, Yann and Manuelli, Lucas and Mousavian, Arsalan and Tyree, Stephen and Birchfield, Stan and Tremblay, Jonathan and Carpentier, Justin and Aubry, Mathieu and Fox, Dieter and Sivic, Josef},
  title     = {{MegaPose}: {6D} pose estimation of novel objects via render \& compare},
  booktitle = {Conference on Robot Learning (CoRL)},
  year      = {2022},
  url       = {https://arxiv.org/abs/2212.06870},
  eprint    = {2212.06870},
  archivePrefix= {arXiv}
}

@inproceedings{lipson.cvpr.2022,
  author    = {Lipson, Lahav and Teed, Zachary and Goyal, Ankit and Deng, Jia},
  title     = {Coupled iterative refinement for {6D} multi-object pose estimation},
  booktitle = {IEEE/CVF Conference on Computer Vision and Pattern Recognition (CVPR)},
  year      = {2022},
  url       = {https://arxiv.org/abs/2204.12516},
  eprint    = {2204.12516},
  archivePrefix= {arXiv}
}

@inproceedings{wen.cvpr.2024,
  author    = {Wen, Bowen and Yang, Wei and Kautz, Jan and Birchfield, Stan},
  title     = {{FoundationPose}: Unified {6D} pose estimation and tracking of novel objects},
  booktitle = {IEEE/CVF Conference on Computer Vision and Pattern Recognition (CVPR)},
  year      = {2024},
  url       = {https://arxiv.org/abs/2312.08344},
  eprint    = {2312.08344},
  archivePrefix= {arXiv}
}

@inproceedings{kato.cvpr.2018,
  author    = {Kato, Hiroharu and Ushiku, Yoshitaka and Harada, Tatsuya},
  title     = {Neural {3D} mesh renderer},
  booktitle = {IEEE/CVF Conference on Computer Vision and Pattern Recognition (CVPR)},
  year      = {2018},
  url       = {https://arxiv.org/abs/1711.07566},
  eprint    = {1711.07566},
  archivePrefix= {arXiv}
}

@inproceedings{liu.iccv.2019,
  author    = {Liu, Shichen and Li, Tianye and Chen, Weikai and Li, Hao},
  title     = {Soft rasterizer: A differentiable renderer for image-based {3D} reasoning},
  booktitle = {IEEE/CVF International Conference on Computer Vision (ICCV)},
  year      = {2019},
  url       = {https://arxiv.org/abs/1904.01786},
  eprint    = {1904.01786},
  archivePrefix= {arXiv}
}

@inproceedings{yen-chen.iros.2021,
  author    = {Yen-Chen, Lin and Florence, Pete and Barron, Jonathan T. and Rodriguez, Alberto and Isola, Phillip and Lin, Tsung-Yi},
  title     = {{iNeRF}: Inverting neural radiance fields for pose estimation},
  booktitle = {IEEE/RSJ International Conference on Intelligent Robots and Systems (IROS)},
  year      = {2021},
  url       = {https://arxiv.org/abs/2012.05877},
  eprint    = {2012.05877},
  archivePrefix= {arXiv}
}

@inproceedings{wohlhart.cvpr.2015,
  author    = {Wohlhart, Paul and Lepetit, Vincent},
  title     = {Learning descriptors for object recognition and {3D} pose estimation},
  booktitle = {IEEE Conference on Computer Vision and Pattern Recognition (CVPR)},
  year      = {2015},
  url       = {https://arxiv.org/abs/1502.05908},
  eprint    = {1502.05908},
  archivePrefix= {arXiv}
}

@inproceedings{sundermeyer.eccv.2018,
  author    = {Sundermeyer, Martin and Marton, Zoltan-Csaba and Durner, Maximilian and Brucker, Manuel and Triebel, Rudolph},
  title     = {Implicit {3D} orientation learning for {6D} object detection from {RGB} images},
  booktitle = {European Conference on Computer Vision (ECCV)},
  year      = {2018},
  url       = {https://arxiv.org/abs/1902.01275},
  eprint    = {1902.01275},
  archivePrefix= {arXiv}
}

@inproceedings{zhou.cvpr.2019,
  author    = {Zhou, Yi and Barnes, Connelly and Lu, Jingwan and Yang, Jimei and Li, Hao},
  title     = {On the continuity of rotation representations in neural networks},
  booktitle = {IEEE/CVF Conference on Computer Vision and Pattern Recognition (CVPR)},
  year      = {2019},
  url       = {https://arxiv.org/abs/1812.07035},
  eprint    = {1812.07035},
  archivePrefix= {arXiv}
}

@inproceedings{tobin.iros.2017,
  author    = {Tobin, Josh and Fong, Rachel and Ray, Alex and Schneider, Jonas and Zaremba, Wojciech and Abbeel, Pieter},
  title     = {Domain randomization for transferring deep neural networks from simulation to the real world},
  booktitle = {IEEE/RSJ International Conference on Intelligent Robots and Systems (IROS)},
  year      = {2017},
  url       = {https://arxiv.org/abs/1703.06907},
  eprint    = {1703.06907},
  archivePrefix= {arXiv}
}

@inproceedings{kirillov.iccv.2023,
  author    = {Kirillov, Alexander and Mintun, Eric and Ravi, Nikhila and Mao, Hanzi and Rolland, Chloe and Gustafson, Laura and Xiao, Tete and Whitehead, Spencer and Berg, Alexander C. and Lo, Wan-Yen and Doll{\'a}r, Piotr and Girshick, Ross},
  title     = {Segment anything},
  booktitle = {IEEE/CVF International Conference on Computer Vision (ICCV)},
  year      = {2023},
  url       = {https://arxiv.org/abs/2304.02643},
  eprint    = {2304.02643},
  archivePrefix= {arXiv}
}

@inproceedings{nguyen.iccvw.2023,
  author    = {Nguyen, Van Nguyen and Groueix, Thibault and Ponimatkin, Georgy and Lepetit, Vincent and Hoda{\v{n}}, Tom{\'a}{\v{s}}},
  title     = {{CNOS}: A strong baseline for {CAD}-based novel object segmentation},
  booktitle = {IEEE/CVF International Conference on Computer Vision Workshops (ICCVW)},
  year      = {2023},
  url       = {https://arxiv.org/abs/2307.11067},
  eprint    = {2307.11067},
  archivePrefix= {arXiv}
}

@inproceedings{li.eccv.2016,
  author    = {Li, Ke and Malik, Jitendra},
  title     = {Amodal instance segmentation},
  booktitle = {European Conference on Computer Vision (ECCV)},
  year      = {2016},
  url       = {https://arxiv.org/abs/1604.08202},
  eprint    = {1604.08202},
  archivePrefix= {arXiv}
}

@inproceedings{zhu.cvpr.2017,
  author    = {Zhu, Yan and Tian, Yuandong and Metaxas, Dimitris and Doll{\'a}r, Piotr},
  title     = {Semantic amodal segmentation},
  booktitle = {IEEE Conference on Computer Vision and Pattern Recognition (CVPR)},
  year      = {2017},
  url       = {https://arxiv.org/abs/1509.01329},
  eprint    = {1509.01329},
  archivePrefix= {arXiv}
}

@article{xiao.pr.2010,
  author  = {Xiao, Bin and Ma, Jian-Feng and Wang, Xuan},
  title   = {Image analysis by {Bessel--Fourier} moments},
  journal = {Pattern Recognition},
  volume  = {43},
  number  = {8},
  pages   = {2620--2629},
  year    = {2010},
  doi     = {10.1016/j.patcog.2010.03.013}
}

@article{dempster.jrssb.1977,
  author  = {Dempster, Arthur P. and Laird, Nan M. and Rubin, Donald B.},
  title   = {Maximum likelihood from incomplete data via the {EM} algorithm},
  journal = {Journal of the Royal Statistical Society: Series B (Methodological)},
  volume  = {39},
  number  = {1},
  pages   = {1--38},
  year    = {1977},
  doi     = {10.1111/j.2517-6161.1977.tb01600.x}
}

@inproceedings{curless.siggraph.1996,
  author    = {Curless, Brian and Levoy, Marc},
  title     = {A volumetric method for building complex models from range images},
  booktitle = {Proceedings of SIGGRAPH},
  pages     = {303--312},
  year      = {1996},
  doi       = {10.1145/237170.237269}
}

@article{koenderink.bc.1979,
  author  = {Koenderink, Jan J. and van Doorn, Andrea J.},
  title   = {The internal representation of solid shape with respect to vision},
  journal = {Biological Cybernetics},
  volume  = {32},
  number  = {4},
  pages   = {211--216},
  year    = {1979},
  doi     = {10.1007/BF00337644}
}

@article{plantinga.ijcv.1990,
  author  = {Plantinga, Harry and Dyer, Charles R.},
  title   = {Visibility, occlusion, and the aspect graph},
  journal = {International Journal of Computer Vision},
  volume  = {5},
  number  = {2},
  pages   = {137--160},
  year    = {1990},
  doi     = {10.1007/BF00054919}
}

@book{milnor.1965,
  author    = {Milnor, John W.},
  title     = {Topology from the Differentiable Viewpoint},
  publisher = {University Press of Virginia},
  address   = {Charlottesville},
  year      = {1965}
}

@inproceedings{chen.iclr.2016,
  author    = {Chen, Tianqi and Goodfellow, Ian and Shlens, Jonathon},
  title     = {{Net2Net}: Accelerating learning via knowledge transfer},
  booktitle = {International Conference on Learning Representations (ICLR)},
  year      = {2016},
  url       = {https://arxiv.org/abs/1511.05641},
  eprint    = {1511.05641},
  archivePrefix= {arXiv}
}

@inproceedings{liu.iclr.2020,
  author    = {Liu, Liyuan and Jiang, Haoming and He, Pengcheng and Chen, Weizhu and Liu, Xiaodong and Gao, Jianfeng and Han, Jiawei},
  title     = {On the variance of the adaptive learning rate and beyond},
  booktitle = {International Conference on Learning Representations (ICLR)},
  year      = {2020},
  url       = {https://arxiv.org/abs/1908.03265},
  eprint    = {1908.03265},
  archivePrefix= {arXiv}
}

@inproceedings{shazeer.icml.2018,
  author    = {Shazeer, Noam and Stern, Mitchell},
  title     = {Adafactor: Adaptive learning rates with sublinear memory cost},
  booktitle = {International Conference on Machine Learning (ICML)},
  year      = {2018},
  url       = {https://arxiv.org/abs/1804.04235},
  eprint    = {1804.04235},
  archivePrefix= {arXiv}
}

@article{molybog.arxiv.2023,
  author  = {Molybog, Igor and others},
  title   = {A theory on {Adam} instability in large-scale machine learning},
  journal = {arXiv preprint arXiv:2304.09871},
  year    = {2023},
  url     = {https://arxiv.org/abs/2304.09871},
  eprint  = {2304.09871},
  archivePrefix= {arXiv}
}

@inproceedings{huang.iclr.2025,
  author    = {Huang, Tianjin and Zhu, Ziquan and Jin, Gaojie and Liu, Lu and Wang, Zhangyang and Liu, Shiwei},
  title     = {{SPAM}: Spike-aware {Adam} with momentum reset for stable {LLM} training},
  booktitle = {International Conference on Learning Representations (ICLR)},
  year      = {2025},
  url       = {https://arxiv.org/abs/2501.06842},
  eprint    = {2501.06842},
  archivePrefix= {arXiv}
}

@inproceedings{bai.icml.2026,
  author    = {Bai, Zhiwei and Zhou, Zhangchen and Zhao, Jiajie and Li, Xiaolong and Li, Zhiyu and Xiong, Feiyu and Yang, Hongkang and Zhang, Yaoyu and Xu, Zhi-Qin John},
  title     = {Adaptive preconditioners trigger loss spikes in {Adam}},
  booktitle = {International Conference on Machine Learning (ICML)},
  series    = {Proceedings of Machine Learning Research},
  volume    = {306},
  pages     = {5566--5601},
  year      = {2026},
  url       = {https://arxiv.org/abs/2506.04805},
  eprint    = {2506.04805},
  archivePrefix= {arXiv}
}

@misc{bop.leaderboard.itodd,
  author       = {{BOP}},
  title        = {{BOP} leaderboard: Model-based {6D} localization of seen objects, {ITODD}},
  howpublished = {\url{https://bop.felk.cvut.cz/leaderboards/pose-estimation-bop19/itodd/}},
  note         = {Accessed 2026-10-26},
  year         = {2026}
}

\appendix
\section{Propagation of derivative tensors}
\label{app:derivatives}

We describe the three passes for a single case; in practice, they are executed for a batch of $n_c$ cases in parallel on CPU or GPU. The parameters of the twinches that map $s^j$ to $s^{j+1}$ are denoted by $a^j$, and $M^j = \partial s^{j+1}/\partial s^j$ denotes the layer Jacobian. Because each layer consists of disjoint $2\times 2$ blocks, $M^j$ is block diagonal: multiplying a vector by $M^j$, by its transpose, or by its inverse costs $O(n_s)$, and the matrix need not be formed. In contrast, forming the $n_s \times n_s$ matrix $\partial s^{\mathrm{out}}/\partial s^{\mathrm{in}}$, let alone $\partial^2 s^{\mathrm{out}}/\partial s^{\mathrm{in}}\partial a$, for each case would be wasteful.

\paragraph{Forward-query.}
Given $n_p$ input tangent vectors $Q^0 = \partial s^0/\partial p$ (e.g., $Q^0 = \partial f/\partial p$), for each layer $j = 0, \ldots, n_l - 1$:
\begin{enumerate}
  \item compute $s^{j+1}$ and the blocks of $M^j$;
  \item update $Q^{j+1} = M^j Q^j$.
\end{enumerate}
The leading $n_p$ rows of $Q^{n_l}$ form $J = \partial r/\partial p$. The pass costs $O(n_c n_s n_l n_p)$ operations and requires $O(n_c n_s n_p)$ memory (storing the parameters requires $O(n_s n_l)$). The plain evaluation of $s^{n_l}$, referred to as \emph{forward-inference}, costs $O(n_c n_s n_l)$.

\paragraph{Backward-variance.}
To obtain $V^0 = \partial r/\partial s^0$, e.g., $\partial r/\partial y$ for inference, we perform a forward-inference pass, initialize $V^{n_l} = \partial r/\partial s^{n_l} = I_{n_p, n_s}$, where $(I_{n_p,n_s})_{i,k} = \delta_{i,k}$ is an $n_p \times n_s$ matrix, and for $j = n_l - 1, \ldots, 0$:
\begin{enumerate}
  \item reconstruct $s^j$ from $s^{j+1}$ with the closed-form inverse and compute the blocks of $M^j$;
  \item update $V^j = V^{j+1} M^j$.
\end{enumerate}
The cost and memory are as for forward-query.

\paragraph{Backward-query.}
We consider a loss $L$ that depends on $J = \partial r/\partial p$ (more generally, on an $n_r \times n_p$ Jacobian with $n_r, n_p \le n_s$) and possibly on the states $s^j$ of all layers. The latter dependence is used to keep all components of $s$ within a window in which the twinches remain effective, consistent with the range $c_0, c_1 \in (-2, 2)$ of their centers. We define auxiliary tensors, with $k$ indexing the components of $r$, $m$ the components of $p$, and $\alpha, \beta$ the components of $s^j$:
\begin{equation}
A^j_\alpha = \sum_{k,m,\beta} \frac{\partial L}{\partial J_{km}}\, \frac{\partial^2 r_k}{\partial s^j_\alpha\, \partial s^j_\beta}\, C^j_{\beta m}, \qquad
B^j_{\alpha m} = \sum_k \frac{\partial L}{\partial J_{km}}\, \frac{\partial r_k}{\partial s^j_\alpha}, \qquad
C^j = \frac{\partial s^j}{\partial p}, \qquad
b^j = \frac{\partial L}{\partial s^j},
\label{eq:aux}
\end{equation}
so that $A^j, b^j \in \R^{n_s}$ and $B^j, C^j \in \R^{n_s \times n_p}$; here $b^j$ refers to the explicit dependence of $L$ on the states. After a forward-query pass, we initialize
\begin{equation}
A^{n_l} = 0, \qquad B^{n_l} = \frac{\partial L}{\partial J}\ \text{(zero-padded to $n_s \times n_p$)}, \qquad C^{n_l} = Q^{n_l}, \qquad b^{n_l} = \frac{\partial L}{\partial s^{n_l}},
\label{eq:aux-init}
\end{equation}
where $B^{n_l}$ and $b^{n_l}$ specify the loss for the case under consideration. For $j = n_l - 1, \ldots, 0$, after reconstructing $s^j$ from $s^{j+1}$, we compute
\begin{equation}
\begin{aligned}
&\text{1. } C^j = \big(M^j\big)^{-1} C^{j+1};\\
&\text{2. } \left.\frac{\partial L}{\partial a^j}\right|_{B} = \sum_{\gamma,\beta,m} B^{j+1}_{\gamma m}\, \frac{\partial^2 s^{j+1}_\gamma}{\partial s^j_\beta\, \partial a^j}\, C^j_{\beta m};\\
&\text{3. } \left.\frac{\partial L}{\partial a^j}\right|_{A} = \sum_\gamma A^{j+1}_\gamma\, \frac{\partial s^{j+1}_\gamma}{\partial a^j};\\
&\text{4. } A^j_\alpha = \sum_\gamma A^{j+1}_\gamma M^j_{\gamma\alpha} + \sum_{\gamma,\beta,m} B^{j+1}_{\gamma m}\, \frac{\partial^2 s^{j+1}_\gamma}{\partial s^j_\alpha\, \partial s^j_\beta}\, C^j_{\beta m};\\
&\text{5. } B^j = \big(M^j\big)^\top B^{j+1};\\
&\text{6. } \left.\frac{\partial L}{\partial a^j}\right|_{b} = \sum_\gamma b^{j+1}_\gamma\, \frac{\partial s^{j+1}_\gamma}{\partial a^j};\\
&\text{7. } b^j = \big(M^j\big)^\top b^{j+1} + \frac{\partial L_{\mathrm{window}}}{\partial s^j};\\
&\text{8. } \frac{\partial L^{\mathrm{batch}}}{\partial a^j} = \sum_{\mathrm{batch}} \left( \left.\frac{\partial L}{\partial a^j}\right|_{A} + \left.\frac{\partial L}{\partial a^j}\right|_{B} + \left.\frac{\partial L}{\partial a^j}\right|_{b} \right),
\end{aligned}
\label{eq:backward-query}
\end{equation}
where $L_{\mathrm{window}}$ denotes the penalty on components of $s^j$ leaving the window. The B-part accounts for the change of the Jacobian of layer $j$ itself, the A-part for the change of the Jacobians of all subsequent layers caused by the change of $s^{j+1}$, and the b-part for the explicit dependence of the loss on the states. Step~1 recovers $C^j$ backward through the inverse layer Jacobian, so that no intermediate states or tangent vectors need to be stored. Despite the third-order tensors involved, all operations are block sparse, and the pass costs $O(n_c n_s n_l n_p)$ operations while requiring $O(n_c n_s n_p)$ memory.

\FloatBarrier
\section{Details of the ITODD application}
\label{app:itodd}

This appendix summarizes the charts and encoders of Sec.~\ref{sec:itodd} and, in more detail, what made the training reliable for the connector, because these findings concern twinchers in general. The research log~\cite{twincher-itodd} contains the scripts, configurations, and further experiments.

\paragraph{Charts.}
The distance $D$ of the virtual camera ranges over 638.4--776.0\,mm, the camera--object distances of the ITODD test images. For the cylinder, $p = (D, e, \psi)$, where $e \in \pm 72.8^\circ$ is the elevation of the viewing direction above the plane perpendicular to the cylinder axis and $\psi$ the roll, with $\psi \in \pm 72.8^\circ$ (chart A) or $\psi \in [17.2^\circ, 162.8^\circ]$ (chart B). The ranges exclude the view along the axis, where the roll is undefined, and they are too narrow to contain two views related by the flip symmetry, $(e, \psi) \sim (-e, \psi + \pi)$. For the connector, $p = (D, e, \varphi, \psi)$, with the elevation $e \in \pm 60^\circ$ above a plane of the model, the azimuth $\varphi \in \pm 150^\circ$ within this plane, and the roll $\psi \in c \pm 75^\circ$ about a center $c \in \{0, \pm 120^\circ\}$. In charts Z0--Z2, the plane is perpendicular to the $z$ axis of the model, and in charts Y0--Y2, to its $y$ axis. A roll range of $150^\circ$ overlaps with its neighbors but cannot contain two rolls $180^\circ$ apart, which look alike because of the near half-turn symmetry. The views at the centers of the Y charts look into the bore of the branch of the connector, and those at $\varphi = \pm 90^\circ$ along its main pipe.

\paragraph{Encoders.}
For the connector, the heights $h(\rho, \theta)$ of the depth map above the background plane are sampled on a disk about the center of mass of the silhouette, whose radius is half the side of the rendered square, and expanded in the orthonormal Bessel modes of the disk~\cite{xiao.pr.2010},
\begin{equation}
c_{nm} = \int_0^{2\pi}\!\!\int_0^1 h(\rho, \theta)\, N_{nm} J_m(j_{mn} \rho)\, e^{-im\theta}\, \rho\, \mathrm{d}\rho\, \mathrm{d}\theta,
\label{eq:fourier-bessel}
\end{equation}
where $J_m$ is the Bessel function of the first kind, $j_{mn}$ its $n$-th zero, and $N_{nm}$ a normalization. The real and imaginary parts of the modes of lowest order $|m| + 2(n-1)$ give 63 components $c_k$, each mapped to $y_k = \xi/\sqrt{1 + \xi^2}$ with $\xi = C e^{-(n/3)^2} c_k$ and $C = 12$, which suppresses fine radial structure. The background plane lies 20\,mm behind the mean depth of the visible object, and heights below it are clipped to zero, so that deep structures, such as the inner walls of bores, do not enter $y$; one further component encodes the mean depth. A roll of the camera by $\alpha$ multiplies $c_{nm}$ by $e^{-im\alpha}$, which makes the dependence on the roll, and its derivative, exact.

\paragraph{Resolution of the training data.}
A twincher sees $f$ only through its training data. With too coarse a grid, it satisfied $\det J > 0$ everywhere on the interpolated data, whereas the exact data showed $\det J \le 0$ in patches where the interpolated tangent vectors $\partial f/\partial p$ deviated strongly from the exact ones (by up to 80\%). A practical criterion, which can be checked before training, is that the 90\% quantile of the relative error of the interpolated tangent vectors should stay below about 0.2. For chart Y0, it requires about 32 nodes in elevation, 96 in azimuth, 24 in roll, and 8 in distance over the full ranges, i.e., $5.9 \times 10^5$ nodes in four dimensions. For the connector, the exact roll dependence of the Fourier--Bessel components reduced the grid to $8 \times 32 \times 96$ nodes with the roll at the center of the chart, 24 times fewer, with smaller interpolation errors. All runs used batches of 5120 samples and the Adam optimizer with a learning rate of $10^{-3}$.

\begin{figure}[t]
\centering
\includegraphics[width=\linewidth]{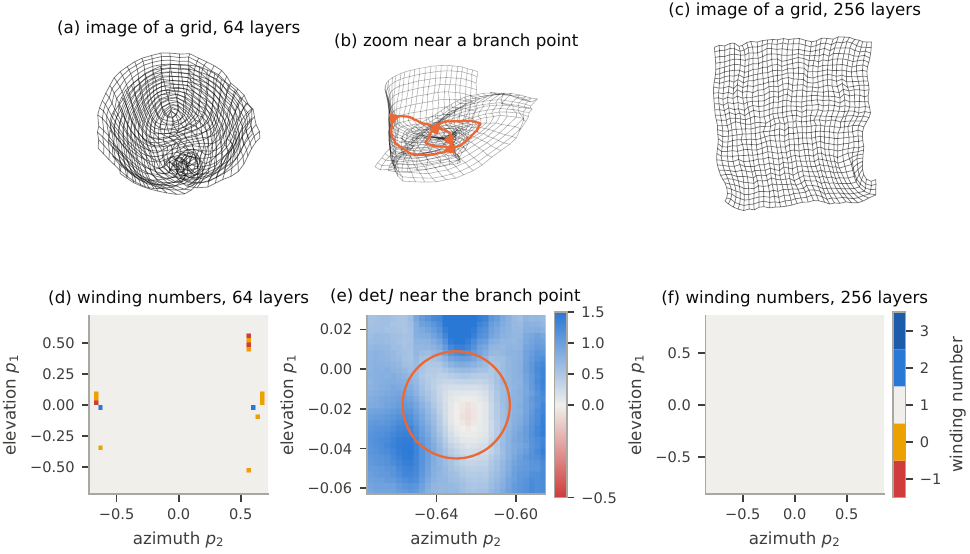}
\caption{Branch points in a two-parameter version of chart Y0 of the connector (elevation and azimuth; an $8 \times 8$ grid of local depth averages as encoder). (a)~Image under $g = r \circ f$ of a grid on $\|p\|_\infty \le 0.72$ for a twincher with 64 layers, whose curriculum stalled at $R = 0.735$; (c)~the same on $\|p\|_\infty \le 0.86$ for 256 layers. (d,~f)~Winding numbers of the images of small circles tiling the same domains: an embedding has winding number 1 everywhere, as for 256 layers; for 64 layers, 0.9\% of the circles deviate, where 2 marks a branch point and 0 and $-1$ mark folds. (b)~Zoom on a branch point of the 64-layer twincher: the image of a small circle around it (orange) winds twice. (e)~$\det J$ around the same point: it is positive except in a small disk inside the circle (orange), so that the determinant loss barely registers the defect.}
\label{fig:itodd-vortices}
\end{figure}

\paragraph{Branch points and winding numbers.}
With adequate data, the most difficult chart (Y0) still failed: the curriculum stalled, and the iteration converged to wrong solutions with the same representation, although $\det J > 0$ almost everywhere. Figure~\ref{fig:itodd-vortices} shows the cause in a two-parameter version of the chart (elevation and azimuth): branch points at which $g$ is locally two-to-one, as $z \mapsto z^2$ at the origin. Around a branch point, $\det J$ is nonnegative and falls below the margin only in a small disk, so that $L_{\det}$ barely registers the defect, whereas the promotion rule of the curriculum registers it enough to stall. This is a concrete instance of the gap between local and global injectivity noted in Remark~\ref{rem:dataspace}. Such defects can be detected in any dimension by winding numbers, a discrete form of the topological degree~\cite{milnor.1965}: the image under $g$ of a small circle in a coordinate plane of $\mathcal{P}$ is projected onto the plane of its two leading principal axes, oriented by the images of the two coordinate directions, and for an embedding it winds exactly once around the image of the center of the circle. We therefore evaluate winding numbers on tilings of coordinate planes in every exact check. In the two-parameter chart, more layers ($n_l = 256$ instead of 64) removed the branch points; with three parameters, this was not sufficient.

\paragraph{The encoder matters.}
The branch points were located at views along bores, where inner surfaces appear and disappear as the view changes slightly, i.e., at visibility events in the sense of aspect graphs~\cite{koenderink.bc.1979, plantinga.ijcv.1990}. There, the curve $f(p)$ nearly turns back on itself: along short paths in the azimuth, the distance between the end points of $f(p)$ was only 0.26 of its arc length. Measuring heights from a nearby background plane, which removes such deep structures from $y$, raised this ratio to 0.45, raised the 1\% quantile of $\sigma_{\min}/\sigma_{\max}$ of $\partial f/\partial p$ from 0.05 to 0.07, and halved the interpolation error of the tangent vectors; Fourier--Bessel components of heights measured from the far background did not improve the ratio (0.25). With the new encoder, training passed the configurations at which all previous runs had failed. Such properties of $f$ can be measured before any training and predicted which encoder would train. More generally, a twincher has to undo whatever the encoder makes hard, so that the choice of encoder is part of the conditioning of the problem.

\paragraph{Constraints beyond the curriculum domain.}
Nothing in $L_{\mathrm{hs}}$ constrains $r$ outside the current hypercube $[-R, R]^{n_p}$. In a twincher saved shortly before a collapse, the column norms of $J$ were at most 1.7 inside the hypercube but reached 45--200 just beyond it, with $\det J \le 0$ for 30\% of the samples there. When the hypercube grew into such a region, single samples contributed trimming losses of order $10^3$, whereas the total loss was of order $10^{-3}$, and the training collapsed. Evaluating the trimming term additionally on samples drawn from the whole domain $[-1,1]^{n_p}$, at the cost of one extra forward-query pass per iteration, prevented these collapses and gave the first functional three-parameter chart Y0.

\paragraph{Function-preserving growth of depth.}
A twincher that stalls for lack of capacity can be extended from its best state: a layer of twinches with zero amplitude ($t = 0$) is the identity, so that inserting such a layer after every second layer (256 to 384 layers) left the map unchanged up to $4 \times 10^{-14}$, a function-preserving transformation in the spirit of Net2Net~\cite{chen.iclr.2016}. Restarting a trained twincher with a fresh Adam optimizer, however, destroyed the map within 50 iterations, even at a learning rate of $10^{-5}$: in its first steps, Adam moves every parameter by about the learning rate, irrespective of the size of its gradient, whereas a trained twincher requires small and very uneven steps. Updating only the moments of Adam for 1000 iterations before resuming, followed by a warm-up of the learning rate~\cite{liu.iclr.2020}, solved this problem. With 384 layers, the four-parameter chart Y0 then reached $R = 0.86$.

\begin{figure}[t]
\centering
\includegraphics[width=\linewidth]{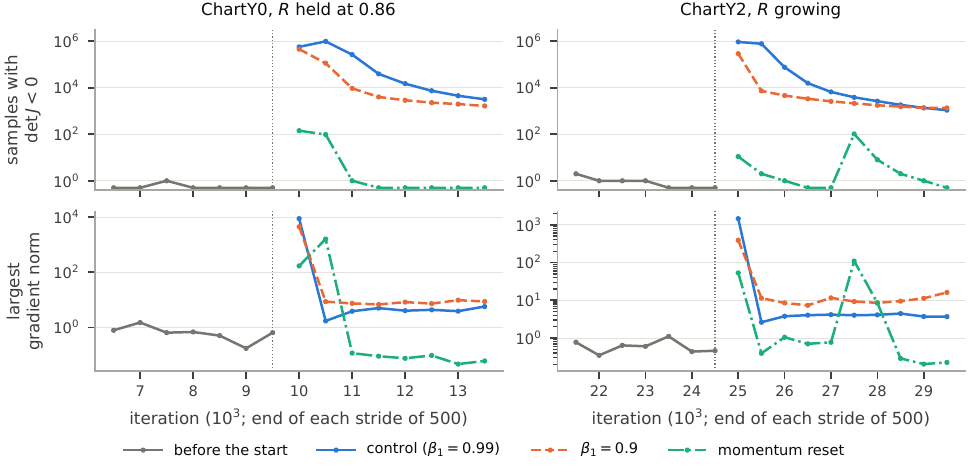}
\caption{Exact replays of two explosions, starting from the state saved before each (dotted line; gray: the original run before it): chart Y0 with the curriculum held at $R = 0.86$ (left) and chart Y2 during the growth of $R$ (right). Top: samples with $\det J < 0$ per stride of 500 iterations ($2.56 \times 10^6$ samples; zero is drawn at the bottom of the axis). Bottom: largest gradient norm in the stride. The control and the run with $\beta_1 = 0.9$ explode; with the momentum reset, gradient spikes still occur but are caught, and chart Y2 continued its curriculum to $R = 0.86$, where the others stalled at $R = 0.757$.}
\label{fig:itodd-momentum}
\end{figure}

\paragraph{Loss spikes and a momentum reset.}
Rare explosions remained: within a few hundred iterations, $\det J$ became negative on a large part of the domain, coinciding with spikes of the gradient norm by three to four orders of magnitude, caused by a small region with large derivatives. Adam converts such a spike into a coherent displacement of all affected parameters. If a gradient component jumps at one step to a value much larger than usual, the first moment carries it for about $1/(1-\beta_1)$ steps, while the second moment keeps the corresponding normalization for about $1/(1-\beta_2)$ steps, so that the total displacement of the parameter is
\begin{equation}
\Delta\theta \approx \eta\, \frac{1 - \beta_1}{\sqrt{1 - \beta_2}} \sum_{k \ge 0} \left(\frac{\beta_1}{\sqrt{\beta_2}}\right)^{k} \approx \frac{\eta}{\sqrt{1 - \beta_2}} \approx 32\,\eta
\label{eq:spike-displacement}
\end{equation}
for a learning rate $\eta$ and $\beta_2 = 0.999$, nearly independently of $\beta_1$ (33$\eta$ for $\beta_1 = 0.99$ and 32$\eta$ for $\beta_1 = 0.9$); $\beta_1$ only sets how fast the displacement occurs. Loss spikes associated with the momentum and the adaptive normalization of Adam have been analyzed in large-scale training~\cite{molybog.arxiv.2023, huang.iclr.2025, bai.icml.2026}. In exact replays of two explosions from the states preceding them (Fig.~\ref{fig:itodd-momentum}), lowering $\beta_1$ to 0.9 did not prevent them, in line with Eq.~\eqref{eq:spike-displacement}. A momentum reset prevented both: when the number of samples with $\det J < 0$ in a batch jumps, the last update is reverted, the first moment of Adam is set to zero while the second moment is kept, and the learning rate is temporarily reduced. This is similar to the momentum reset of SPAM~\cite{huang.iclr.2025}; here, the inflated second moment additionally damps the subsequent steps of exactly the affected parameters. In later runs, which used the reset, no explosion occurred. Bounding the update of each parameter, as in Adafactor~\cite{shazeer.icml.2018}, is an alternative that remains to be tested.

\begin{table}[t]
\centering
\caption{Twinchers of the connector charts ($n_s = 128$, $n_l = 384$). Iterations until $R = 0.86$ and until training stopped (two consecutive clean exact checks), training time on a laptop GPU (partly shared with other runs), and rollbacks to an earlier checkpoint. Evaluation with rendered data on $\|p\|_\infty \le 0.86$: smallest $\det J$ at 8192 random points, and fraction of 512 (Y0: 1024) random synthetic views solved from $p^{(0)} = 0$ in 40 steps. Last two columns: sensitivity $\eta$, the median of $\|\Delta p\| / \|\delta y\|$ for perturbations $\delta y$ normal to $\mathcal{M}$ of norm up to 0.5, before and after the robustness post-training. For the two cylinder charts, $\eta$ (perturbations of norm 0.1--1) decreased from 0.18 and 0.20 to 0.033 and 0.035.}
\label{tab:itodd-charts}
\small
\begin{tabular}{lcccccccc}
\toprule
 & \multicolumn{2}{c}{Iterations ($10^3$)} & Time & & & Solved & \multicolumn{2}{c}{$\eta$} \\
\cmidrule(lr){2-3} \cmidrule(lr){8-9}
Chart & $R = 0.86$ & stop & (min) & Rollbacks & $\min \det J$ & $<10^{-2}$ / $<10^{-3}$ & before & after \\
\midrule
Z0 & 10.5 & 19.5 & 144 & 0 & 0.56 & 100\% / 99.0\% & 0.73 & 0.25 \\
Z1 & 12.0 & 15.0 & 56 & 0 & 0.67 & 100\% / 99.2\% & 0.92 & 0.30 \\
Z2 & 12.5 & 15.0 & 56 & 0 & 0.62 & 100\% / 98.8\% & 0.86 & 0.28 \\
Y0 & 31.5 & 32.5 & 160 & --$^{a}$ & 0.26 & 100\% / 98.8\% & 0.86 & 0.54 \\
Y1 & 27.0 & 30.0 & 137 & 2 & 0.45 & 100\% / 97.9\% & 0.92 & 0.49 \\
Y2 & 28.5 & 32.5 & 120 & 1 & 0.53 & 100\% / 98.0\% & 1.00 & 0.55 \\
\bottomrule
\end{tabular}

\smallskip
\footnotesize{$^{a}$ 25{,}000 iterations with 256 layers, then extended to 384 layers and restarted from the best state.}
\end{table}

\end{document}